\documentclass{article}
\usepackage{iclr2027_conference,times}
\iclrfinalcopy
\usepackage{booktabs}
\usepackage{multirow}
\usepackage{graphicx}
\usepackage{float}
\usepackage{lipsum}

\usepackage{amsmath,amsfonts,bm}

\def\eqref#1{equation~\ref{#1}}
\def\1{\bm{1}}

\DeclareMathAlphabet{\mathsfit}{\encodingdefault}{\sfdefault}{m}{sl}
\SetMathAlphabet{\mathsfit}{bold}{\encodingdefault}{\sfdefault}{bx}{n}

\usepackage{hyperref}
\usepackage{url}
\usepackage{CJKutf8}
\usepackage{xcolor}

\definecolor{darkgreen}{rgb}{0.0, 0.6, 0.0}

\definecolor{purple}{rgb}{0.6, 0.0, 0.6}

\title{Head Dimension Matters in Language Models for Long-Context Retrieval}

\author{Seiya Tokui, Shoichiro Yamaguchi \& Sosuke Kobayashi\\
Preferred Networks, Inc.\\
Tokyo, Japan \\
\texttt{\{tokui,guguchi,sosk\}@preferred.jp}
}

\begin{document}

\maketitle
\lhead{Preprint}

\begin{abstract}
Multi-head attention retrieves and combines context representations through query--key interactions and value aggregation, with head dimension constraining the per-head representation space.
In long-context processing, it remains unclear whether allocating a fixed parameter budget to fewer, wider heads improves information retrieval.
In this paper, we investigate how head dimension affects the performance of long-context retrieval tasks.
Our empirical studies reveal that pretraining loss and short-context non-retrieval task performance are not affected much by head dimension, while long-context retrieval performance improves with wider heads at fixed model sizes.
We find that models with wider heads have a higher proportion of retrieval heads, associating more parameters with retrieval.
Our ablation studies also show that the retrieval performance improves with wider heads even when the projection weights in attention modules are frozen during long-context continual pretraining, suggesting that wider heads not only benefit updates in attention modules but lead to more effective adaptation of components such as embeddings and MLPs.
\end{abstract}

\section{Introduction}

Large language models (LLMs) are increasingly expected to process long texts including documents, multi-turn conversations, and code repositories, as well as longer reasoning traces used for test-time scaling \citep{s1}.
Although modern LLM architectures support processing long inputs, a large supported context window does not necessarily imply effective use of information throughout the context \citep{ruler,litm}.
It is therefore increasingly important to explore improvements in architectures and training recipes for enhanced long-context capabilities.

The ability to retrieve task-relevant information from a large amount of context is a fundamental prerequisite for effective long-context processing \citep{ruler,helmet,icr2}.
Recent work suggests that successful retrieval is not uniformly implemented across a Transformer: a small subset of attention heads has been shown to play a central role in extracting relevant information from long inputs \citep{retrievalhead,qrhead}.

These findings motivate a specific architectural question: \emph{under a fixed attention parameter budget, how does allocating capacity to fewer, wider heads affect long-context retrieval?}
In this work, we study the relationship between long-context retrieval capability and the head dimension of self attention modules.
We train 0.6B and 3B parameter models from scratch through a simplified training pipeline consisting of pretraining (PT) and long-context continual pretraining (CPT) with web and code-centric data mixes, and additionally apply supervised fine-tuning (SFT) to 3B models to examine whether head dimension affects performance on short-context, non-retrieval tasks.
Parameter-matched comparisons of models with identical architectures except for the balance between the number of heads and the head dimension reveal that wider heads lead to better performance on retrieval tasks.

To investigate the performance gap between models with different head dimensions, we first measured the retrieval scores of attention heads following \citet{retrievalhead}.
We found that wider-head models have a higher proportion of retrieval heads, with comparable absolute retrieval-head counts despite fewer heads overall.
This indicates that wider-head models associate more parameters with retrieval heads.
Comparisons between pretrained checkpoints and their corresponding CPT checkpoints further suggest that much of this association is already established during pretraining.

We then repeated CPT with frozen projection weights in attention modules to examine how head dimension affects long-context adaptation.
The improvement in retrieval performance after CPT is larger in models with wider heads even with frozen projections in attention, whereas updating all the projections provides similar additional improvements across head dimensions.
Further analyses show that the sharpness of attention and changes in token-level attention rankings do not fully explain the performance gap.
These results suggest that wider heads improve long-context retrieval by enabling more effective adaptation of the remaining trainable components, including embeddings and MLPs, rather than through greater plasticity of the attention modules themselves.

\section{Experimental setup}

We trained decoder-only Transformers from scratch through a three-stage pipeline consisting 
of PT, CPT, and SFT.
We use models with 0.6B parameters for ablation experiments, and use models with 3B parameters to verify that the comparison is also observed at a larger model scale.
Both types of models consist of sliding window attention (SWA) layers with a window size of 512 tokens and full attention layers that attend to all past key-values, interleaved in 3:1 ratio.
We apply standard RoPE to SWA heads and 64 dimensional partial RoPE to full attention heads.
For partial RoPE, we use a small base $\theta=100$, with which we can use the same $\theta$ without additional scaling during context extension.

For an input sequence $X \in \mathbb{R}^{L \times D}$ with sequence length $L$ and model dimension $D$, multi-head attention (MHA) with $H$ heads is written as
\begin{align}
    \operatorname{MHA}(X)
    &=
    \operatorname{Concat}(\operatorname{head}_1,\ldots,\operatorname{head}_H) W_O, \notag\\
    \operatorname{head}_h
    &=
    \operatorname{softmax}
    \left(
        \frac{XW_Q^{(h)\top} W_K^{(h)} X^\top}{\sqrt{d}} + M
    \right)XW_V^{(h)\top},
    \notag
\end{align}
where $d$ is the head dimension, $W_Q^{(h)}, W_K^{(h)}, W_V^{(h)} \in \mathbb{R}^{d \times D}$, $W_O \in \mathbb{R}^{Hd \times D}$, and $M$ represents the causal or sliding window mask.
The number of heads $H$ and head dimension $d$ are standard architectural hyperparameters in MHA.
For a fixed model dimension $D$, the number of attention parameters scales as $\mathcal{O}(DHd)$ and dense attention computation scales as $\mathcal{O}(L^2Hd)$.
Our models use grouped-query attention, where $W_K^{(h)}$ and $W_V^{(h)}$ are shared among heads in the same group.
We use a fixed query-to-KV head ratio, so the KV-cache size also scales with $Hd$.
Thus, when $Hd$ is fixed, changing the number and size of heads leaves the attention parameter count, computation, and KV-cache size essentially unchanged, while changing how the same parameter and compute budgets are allocated across heads.

\begin{table}[b]
    \caption{
      Architecture names and their configurations.
      Dimensions and head counts refer to full-attention layers.
      The query and key dimensions are always identical, whereas the
      value dimension may differ.
      Full configurations are provided in Appendix~\ref{sec:details}.
    }
    \begin{center}
    \begin{tabular}{c|c|cc|cc}
        \toprule
        Model Name & Params & Q/K Dim. & V Dim. & \#Q Heads & \#KV Heads \\
        \midrule
         \texttt{0.6B-d128}       & 0.6B & 128 & 128 & 16 & 4 \\
         \texttt{0.6B-d256}       & 0.6B & 256 & 256 & 8  & 2 \\
         \texttt{0.6B-d512}       & 0.6B & 512 & 512 & 4  & 1 \\
         \texttt{0.6B-k512-v128}  & 0.6B & 512 & 128 & 8  & 2 \\
         \texttt{0.6B-k128-v512}  & 0.6B & 128 & 512 & 8 & 2 \\
        \midrule
         \texttt{3B-d128}         & 3B   & 128 & 128 & 32 & 8 \\
         \texttt{3B-d256}         & 3B   & 256 & 256 & 16 & 4 \\
         \texttt{3B-d512}         & 3B   & 512 & 512 & 8  & 2 \\
        \bottomrule
    \end{tabular}
    \end{center}
    \label{tbl:head-size}
\end{table}

For each model scale, we prepare three architectures that are identical except for the number of heads and the dimension of each head in full attention layers as shown in Table \ref{tbl:head-size}.
For the symmetric configurations at each model scale, the parameter allocation and key-value cache size are identical.
For the 0.6B model family, we conduct additional experiments on models with asymmetric head size with distinct QK and V dimensions, whose parameter count is slightly larger.

In the PT and CPT stages, we use data mixes consisting of web (60\%), code (30\%), and SFT-like (10\%) datasets.
The sequence length is set to 4,096 in PT and 262,144 in CPT.
We use 100B tokens for PT and 8B tokens for CPT.
We apply SFT to the 3B models using short and long SFT trajectories with the same token ratio, totaling 1B tokens.

We use the AdamW optimizer \citep{adamw} for all the training stages.
In the PT and SFT stages, the learning rate is warmed up from zero to the maximum learning rate and held constant for the remaining steps.
In the CPT stage, we skip the learning rate warm-up and use a constant learning rate.
Additional details of architectures, data mixes, and training setup are provided in Appendix~\ref{sec:details}.

\paragraph{Evaluation:}
We measure the long-context retrieval capability with the RULER benchmark \citep{ruler}.
We evaluate 13 RULER tasks, comprising eight needle-retrieval variants, variable tracking, common-word extraction, frequent-word extraction, and two question-answering tasks.
We evaluate all of these tasks with context lengths ranging from 4,096 to 262,144.
We only generate 100 examples for each task to keep the evaluation cost low.
We call this variant \emph{RULER-100}, which we confirmed does not affect the stability of the evaluation.

\section{Model performance comparisons}
\label{sec:long_context_cpt}

We compare the head allocations in Table~\ref{tbl:head-size} to assess whether wider heads improve long-context retrieval after CPT. 
We examine RULER performance alongside next-token prediction (NTP) training loss to determine whether the two metrics favor the same head allocation.

\begin{table}[t]
    \centering
    \caption{RULER-100 overall scores averaged over 13 tasks and seven lengths from 4K to 256K. Scores are on a 0--100 scale. The 0.6B results are means $\pm$ sample standard deviations across three runs; each 3B allocation has one run.}
    \label{tab:cpt_ruler_overall}
    \begin{tabular}{rclllll}
        \toprule
        Model size & Stage & \texttt{d128} & \texttt{d256} & \texttt{d512} & \texttt{k512-v128} & \texttt{k128-v512}\\ 
        \midrule
        0.6B & CPT
        & 31.25 {\scriptsize $\pm$ 7.75}
        & 35.46 {\scriptsize $\pm$ 3.60}
        & \textbf{41.21} {\scriptsize $\pm$ 5.86} & 37.78 {\scriptsize $\pm$ 4.15} & 31.28 {\scriptsize $\pm$ 2.61}\\
        3B & CPT & 35.93 & 36.72 & \textbf{50.93} & - & - \\
        3B & SFT & 48.28 & 51.81 & \textbf{58.49} & - & - \\
        \bottomrule
    \end{tabular}
\end{table}

\begin{figure}[t]
    \centering
    \begin{minipage}[t]{0.325\linewidth}
        \centering
        \includegraphics[width=\linewidth]{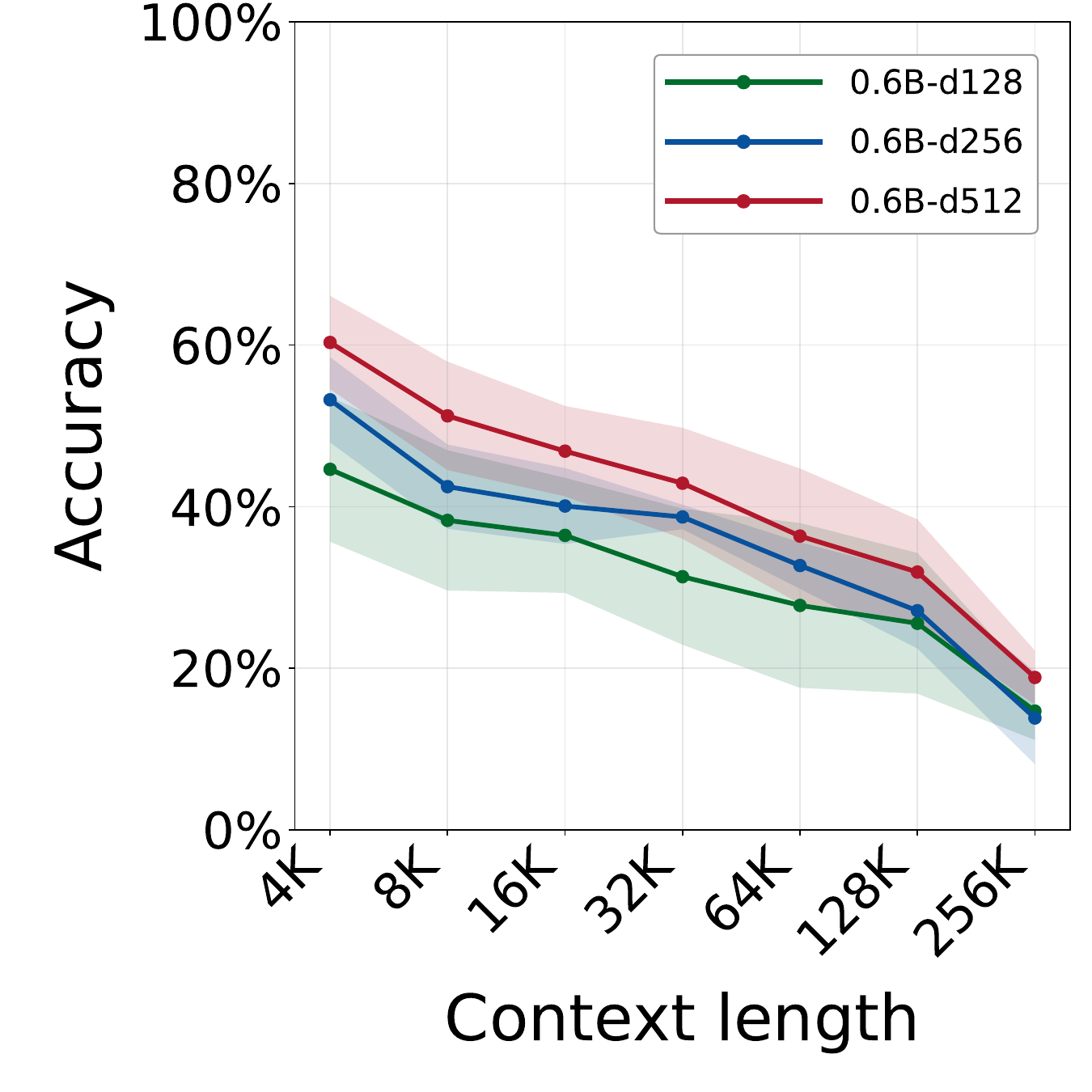}
        \vspace{-1.5em}
        \par\small (a) 0.6B (CPT)
    \end{minipage}
    \begin{minipage}[t]{0.325\linewidth}
        \centering
        \includegraphics[width=\linewidth]{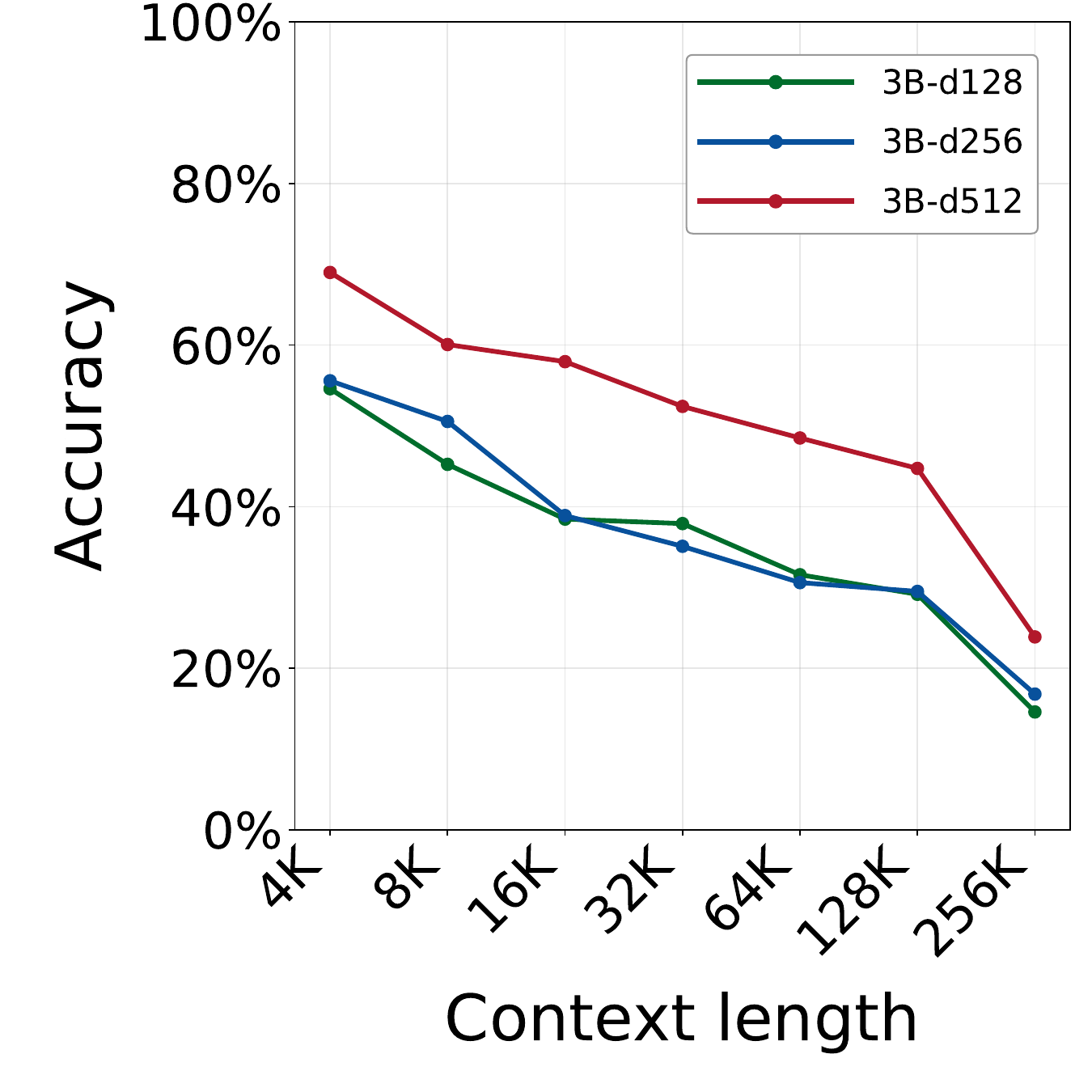}
        \vspace{-1.5em}
        \par\small (b) 3B (CPT)
    \end{minipage}
    \begin{minipage}[t]{0.325\linewidth}
        \centering
        \includegraphics[width=\linewidth]{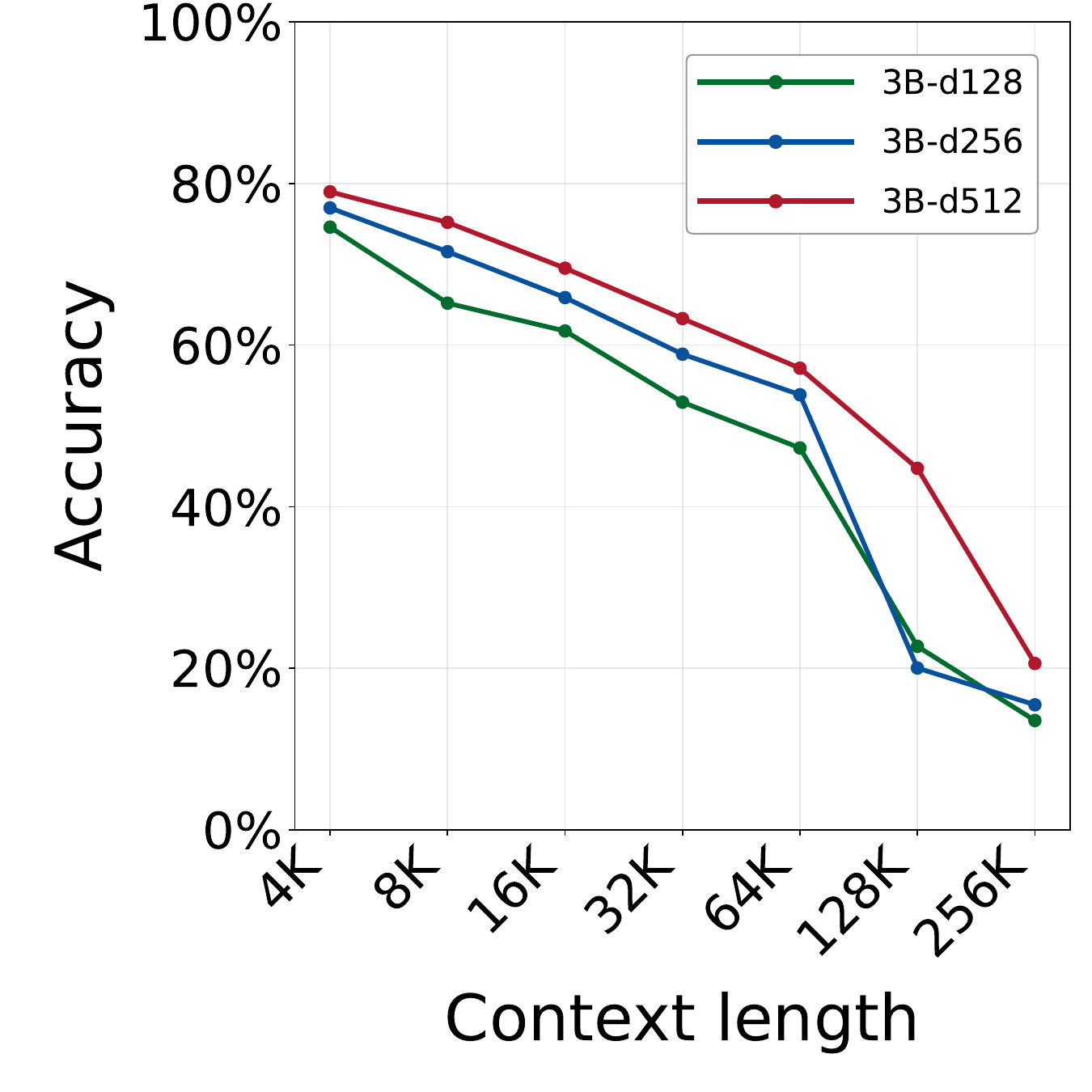}
        \vspace{-1.5em}
        \par\small (c) 3B (SFT)
    \end{minipage}
    \caption{RULER-100 scores by context length. Each point averages 13 tasks at the indicated context length on a 0--100 scale. In (a), lines and bands show the mean and $\pm$ sample standard deviation across three 0.6B runs. In all the settings, \texttt{d512} achieves the highest (average) score at every   evaluated length.}
    \label{fig:cpt_ruler_by_length}
\end{figure}

\paragraph{Wider heads improve retrieval at a fixed attention budget.}
Table~\ref{tab:cpt_ruler_overall} and Figure~\ref{fig:cpt_ruler_by_length} show the average and length-wise RULER-100 scores, respectively.
The \texttt{d512} models achieve the highest average scores in all the settings and context lengths.
On the average scores at 0.6B parameter scale, \texttt{d512} exceeds \texttt{d128} and \texttt{d256} by 9.96 and 5.75 points, respectively; at 3B parameter scale, the gains are 15.00 and 14.21 points.

These results favor allocating the fixed attention parameter budget to fewer, wider heads for long-context retrieval.
The advantage is not uniform across individual tasks, and all allocations still degrade at the longest lengths; Appendix~\ref{app:head_allocation_tasks} provides the breakdown.

\begin{figure}[t]
    \centering
    \begin{minipage}[t]{0.4\linewidth}
        \centering
        \includegraphics[width=\linewidth]{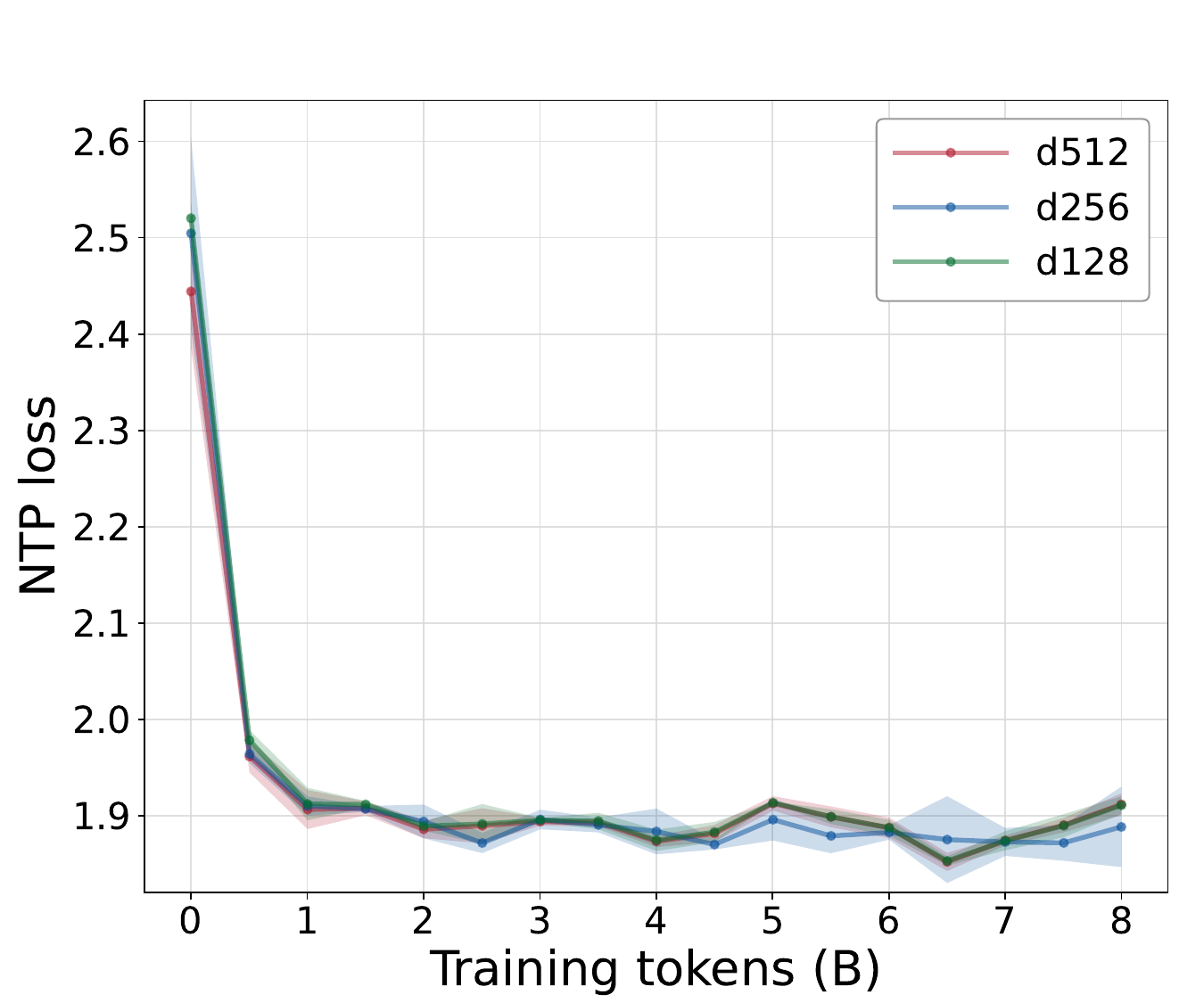}
        \vspace{-1.5em}
        \par\small (a) 0.6B
    \end{minipage}
    \begin{minipage}[t]{0.4\linewidth}
        \centering
        \includegraphics[width=\linewidth]{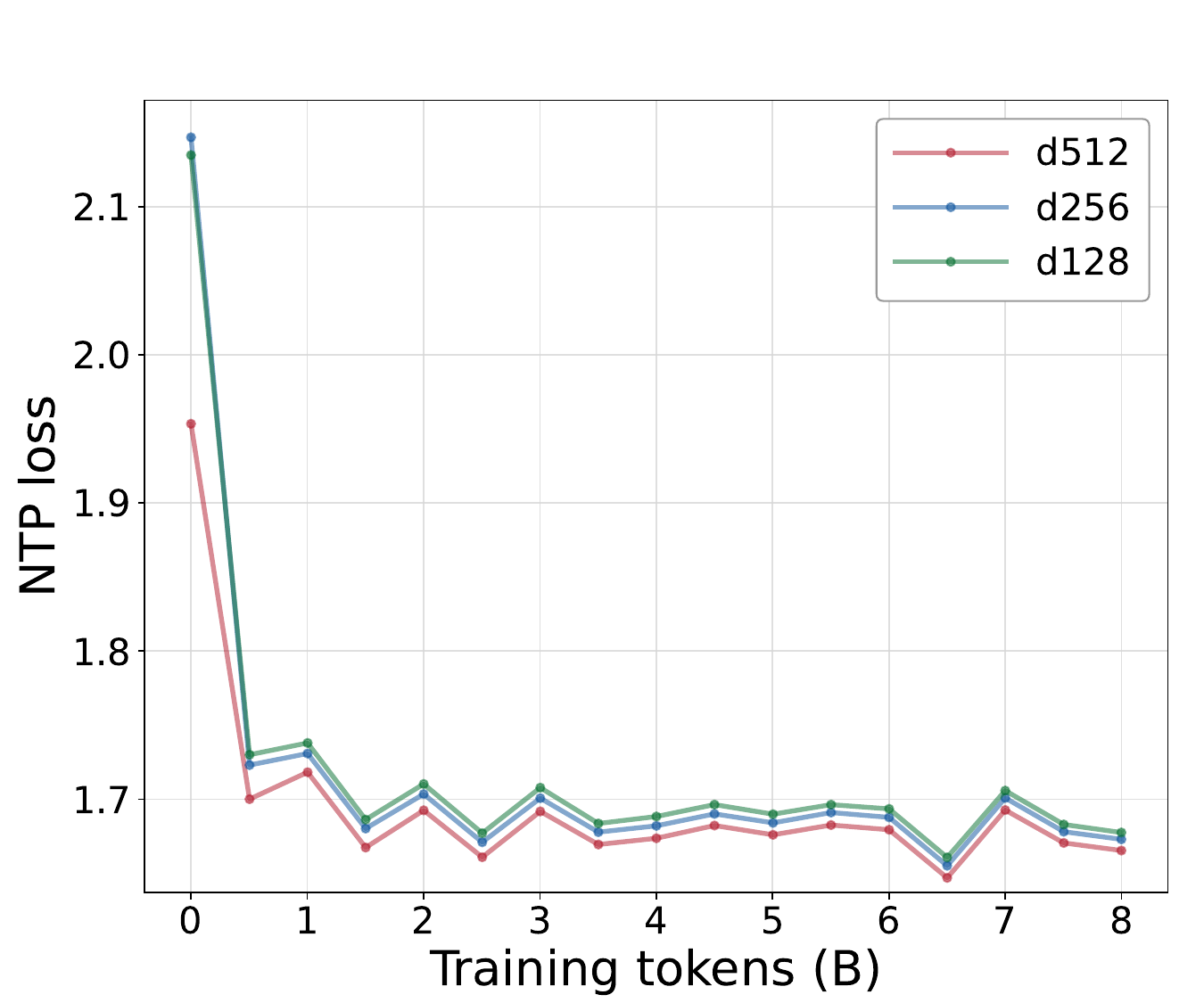}
        \vspace{-1.5em}
        \par\small (b) 3B
    \end{minipage}
    \caption{training NTP loss during 8B-token CPT at 256K. The horizontal axis counts CPT tokens; the vertical axis is training cross-entropy. In (a), lines and bands are the mean and $\pm1$ sample standard deviation across three 0.6B runs. Lower training loss does not always correspond to higher RULER performance.}
    \label{fig:cpt_training_loss}
\end{figure}

\paragraph{Training loss does not fully capture the head-dimension effect.}
Figure~\ref{fig:cpt_training_loss} shows the CPT training NTP loss. 
PT losses also vary little across head dimensions, as shown in Table~\ref{tab:pt_cpt_ntp_loss}.
At 0.6B parameter scale, \texttt{d256} has the lowest mean final loss, yet \texttt{d512} has the highest overall RULER-100 score. 
At 3B parameter scale, \texttt{d512} is best on both metrics. 
Thus, training loss alone does not consistently identify the head allocation with the best long-context retrieval performance. 
This is consistent with prior findings that standard token-averaged perplexity is not a reliable proxy for long-context downstream performance \citep{hu2024canperplexity,fang2025wrongperplexity}.
Final losses and PT trajectories are reported in Appendix~\ref{app:pt_cpt_loss}.

\paragraph{Larger QK dimensions benefit retrieval more than larger V dimensions.}
Among the 0.6B models, \texttt{d512} achieves the highest mean
RULER-100 score, followed by \texttt{k512-v128} and \texttt{d256},
while \texttt{d128} and \texttt{k128-v512} have nearly identical
mean scores (Table~\ref{tab:cpt_ruler_overall}).
The two asymmetric configurations have the same head counts,
and their performance gap favors larger QK over larger V dimensions
for long-context retrieval.
At a QK dimension of 512, the higher mean score of \texttt{d512}
relative to \texttt{k512-v128} suggests an additional benefit
from a larger V dimension.
At a QK dimension of 128, however, \texttt{k128-v512} shows
essentially no gain over \texttt{d128}.
Although the latter two comparisons also change head count,
this pattern is consistent with a QK bottleneck: narrow QK
representations may constrain the selection of relevant positions,
limiting the benefit of additional V capacity for conveying
information from those positions.

\begin{table}[t]
\caption{Short-context task performance for the 3B models after SFT. Abbreviations: ARC-C (ARC-Challenge), ARC-E (ARC-Easy), HS (HellaSwag), WG (WinoGrande), OBQA (OpenBookQA), CSQA (CommonsenseQA), Avg (Average).}
\begin{center}
\begin{tabular}{lcccccccc|c}
\toprule
Head Size & ARC-C & ARC-E & HS & WG & RACE & PIQA & OBQA & CSQA & Avg \\
\midrule
\texttt{d128} & 35.6 & 62.3 & 58.2 & 57.4 & 36.2 & 73.0 & 36.6 & 38.1 & 49.7 \\
\texttt{d256} & \textbf{35.8} & 61.0 & 58.1 & \textbf{59.4} & 35.0 & 73.3 & 37.4 & 28.9 & 48.6 \\
\texttt{d512} & 33.9 & \textbf{64.0} & \textbf{58.7} & 57.9 & \textbf{37.4} & \textbf{73.7} & \textbf{39.0} & \textbf{39.6} & \textbf{50.5} \\
\bottomrule
\end{tabular}
\end{center}
\label{tbl:shorteval}
\end{table}

\paragraph{The retrieval performance is not traded-off with short-context task performance.}
To assess whether a trade off exists between long-context retrieval capabilities and short-context task performance, we evaluate 3B models after SFT with short-context downstream tasks.
The task set consists of question answering (ARC-Easy and ARC-Challenge \citep{arc}), general commonsense reasoning (CommonsenseQA \citep{commonsenseqa}, WinoGrande \citep{WinoGrande}), commonsense NLI (Hellaswag \citep{hellaswag}), physical commonsense reasoning (PIQA \citep{piqa}),  reading comprehension (RACE \citep{race}), and scientific QA (OpenBookQA \citep{openbookqa}).
The results are shown in Table~\ref{tbl:shorteval}.
The model with a head dimension of 512 performs the best on average, while we observe no clear monotonic trend with the head dimensions.
We can at least conclude that head dimension does not affect the short-context tasks negatively, and the improvement in long-context retrieval capability is not traded-off with the short-context performance.

\section{Analysis of retrieval heads}

\begin{figure}[t]
    \centering
    \begin{minipage}[t]{0.7\linewidth}
        \centering
        \includegraphics[width=\linewidth]{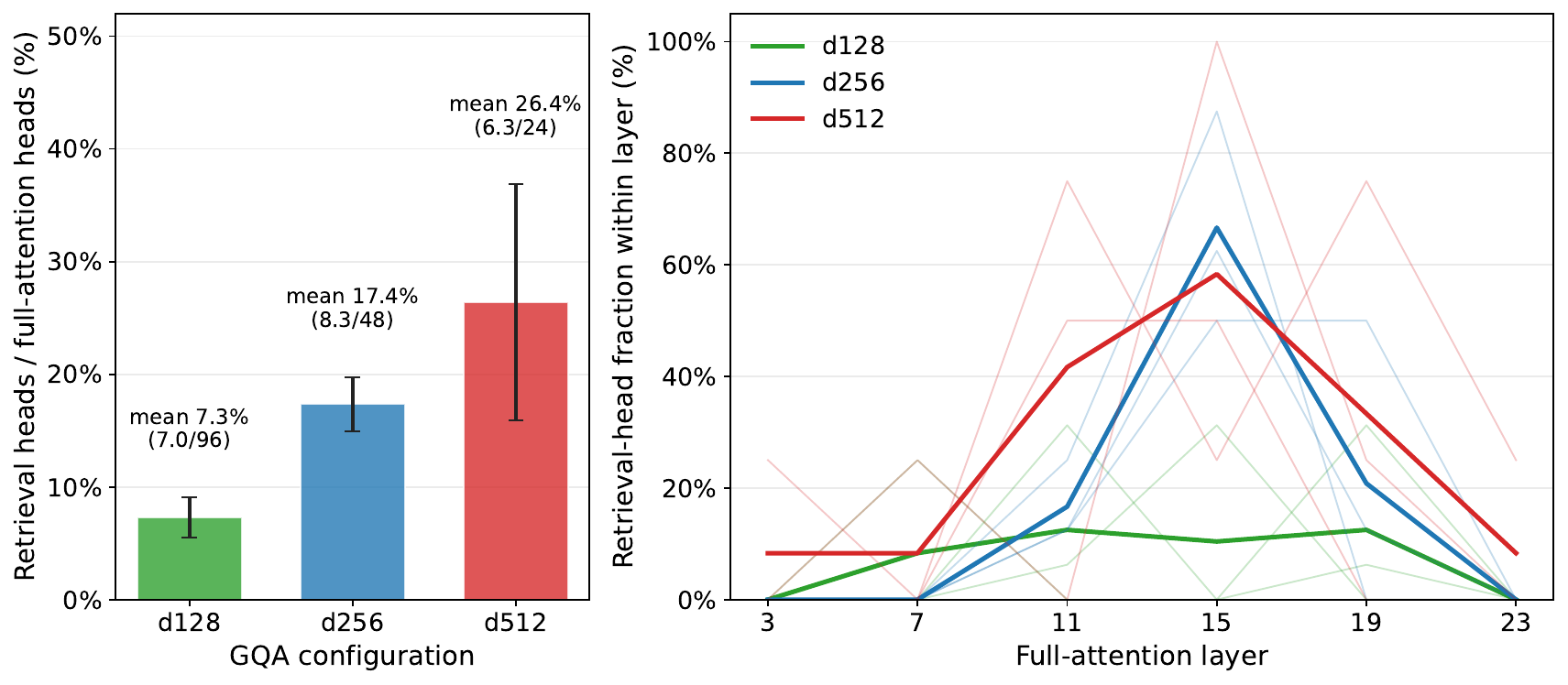}
        \vspace{-1em}
        \par\small (a) 0.6B
    \end{minipage}
    \begin{minipage}[t]{0.7\linewidth}
        \centering
        \includegraphics[width=\linewidth]{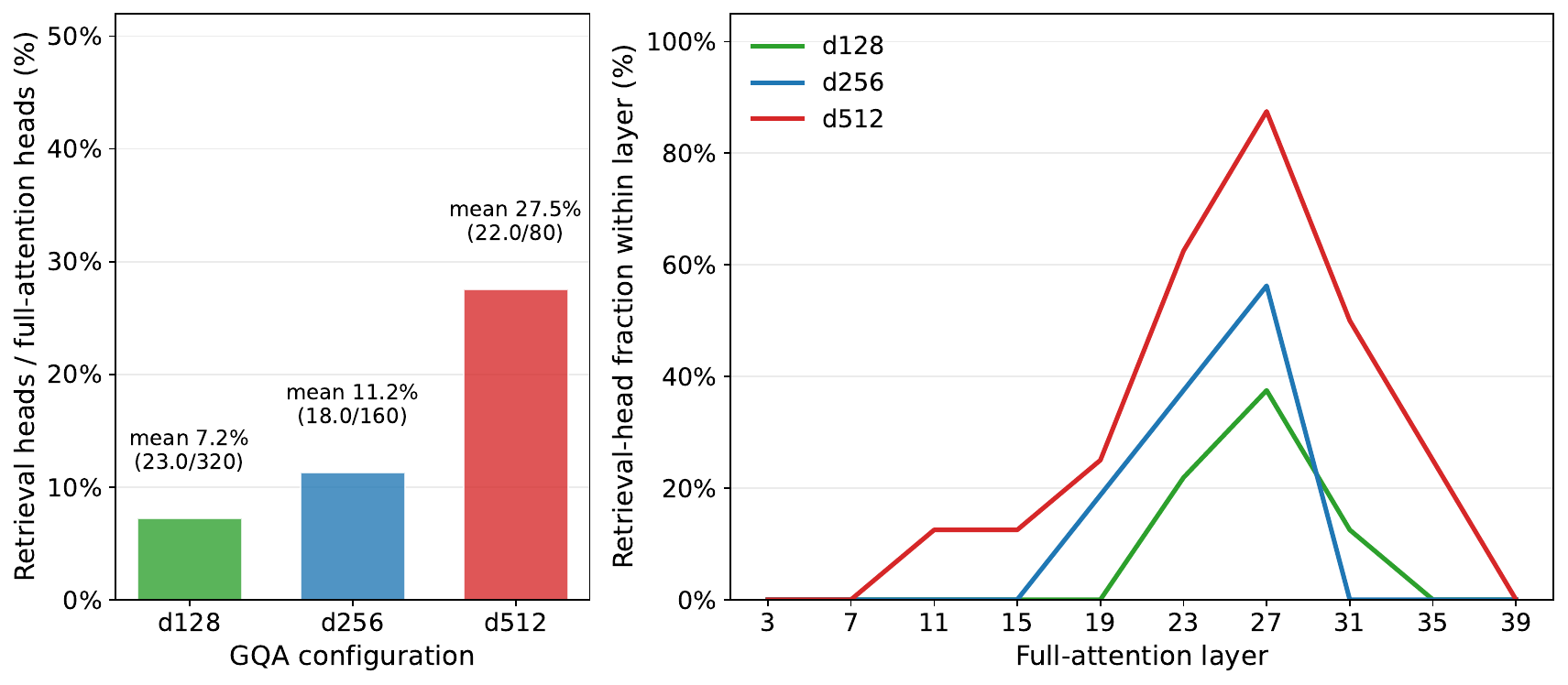}
        \vspace{-1em}
        \par\small (b) 3B (after SFT)
    \end{minipage}
    \caption{Retrieval-head fractions in the (a) 0.6B and (b) 3B models.
    Left: the fraction of full-attention query heads classified as retrieval heads; annotations give the retrieval-head count relative to the total number of eligible heads.
    Right: retrieval-head fractions within individual full-attention layers, using zero-based layer indices. 
    The 0.6B results are means across three runs; faint lines show individual runs.
    Each 3B model has one run.
    Wider-head models exhibit higher retrieval-head fractions at both scales, while absolute retrieval-head counts remain comparable across allocations.
    }
    \label{fig:long-context-head-dimension}
\end{figure}

To assess the mechanistic differences caused by changing head dimension, we measured the number of retrieval heads in each full attention layer.
We use the single-needle variants of needle-in-a-haystack tasks in RULER and measured the retrieval score of each full-attention head.
Following \citet{retrievalhead}, we count a head with the retrieval score $\ge 0.1$ as a retrieval head.
The results in Figure \ref{fig:long-context-head-dimension} reveal that models with wider heads allocate a higher ratio of full-attention heads for retrieval.
Note that the number of retrieval heads does not change significantly despite of fewer total number of heads in models with wider heads.
This indicates that the models with wider heads associate more parameters with retrieval heads in total, which may lead to higher accuracy in attending to needles due to the larger capacity of retrieving queries.

\subsection{Retrieval-head emergence before and after CPT}
\label{subsec:retrieval_head_pt_cpt}

\begin{table}[t]
    \centering
    \caption{
    Retrieval-head counts before and after CPT for the 0.6B models.
    PT denotes the 100B-token PT checkpoint evaluated with
    inference-time attention scaling
    (Appendix~\ref{app:pt_temperature_scaling});
    CPT denotes the corresponding checkpoint after 8B tokens of
    full-parameter CPT, evaluated without additional scaling.
    The rightmost column reports the number of retrieval heads common to both stages, matched by layer and query-head index within each run.
    All the counts are computed within each run and
    then averaged across three runs per configuration.
    }
    \label{tab:retrieval_head_pt_cpt_overlap}
    \begin{tabular}{lccc}
        \toprule
        Model & PT (attention scaling) & CPT & Shared \\
        \midrule
        \texttt{0.6B-d128} & $4.00$ & $7.00$ & $4.00$ \\
        \texttt{0.6B-d256} & $8.33$ & $8.33$ & $8.00$ \\
        \texttt{0.6B-d512} & $5.00$ & $6.33$ & $5.00$ \\
        \bottomrule
    \end{tabular}
\end{table}

To examine how retrieval heads change through CPT,
we compare the 100B-token PT checkpoints with their corresponding
8B-token full-parameter CPT checkpoints for the three symmetric
0.6B-parameter configurations.
The PT checkpoints are evaluated with the inference-time attention
scaling described in Appendix~\ref{app:pt_temperature_scaling},
without updating any model parameters, whereas CPT checkpoints
use no additional scaling.
We identify retrieval heads using the same single-needle RULER
protocol and retrieval-score threshold ($\ge 0.1$) as above.
Table~\ref{tab:retrieval_head_pt_cpt_overlap} reports the numbers of detected and shared full-attention query heads
averaged over three runs.

For \texttt{d128}, the mean retrieval-head count increases from
4.00 in scaled PT to 7.00 after CPT, with all the retrieval heads detected in PT checkpoints retained.
The retrieval-head fraction accordingly rises from 4.2\% to 7.3\%,
indicating that the capacity allocated to detected retrieval heads
is even smaller at PT than after CPT.

The changes in detected counts are smaller for the wider-head models.
For \texttt{d256}, the mean count is 8.33 at both stages, with 8.00 shared heads.
For \texttt{d512}, the mean count increases modestly from 5.00 to 6.33, with all the retrieval heads detected in PT checkpoints retained.
Thus, much of the retrieval heads in these models are already present at the end of pretraining.
Run-level counts and shared membership are reported in Appendix ~\ref{app:retrieval_head_freeze}.

\section{Head dimension and long-context adaptation}
\label{sec:attention_adaptation}

The analyses provided above characterize how head dimension affects retrieval capabilities; here, we examine how models with different head dimensions adapt during CPT.
Because updates on query, key, value, and output projections can directly alter the behavior of attention modules,
we first isolate their contribution by freezing these projections during CPT
in Section~\ref{sec:attention_projection_freeze}.
We then examine two complementary properties of the resulting attention
maps in Section~\ref{sec:token_rank_reorganization}: the concentration of attention probabilities and the ranking of attended token positions.
Together, these analyses test whether adaptation within attention modules accounts for the differences in retrieval performance across head dimensions.

\subsection{Retrieval performance gains of CPT with frozen attention weight matrices}
\label{sec:attention_projection_freeze}

For each symmetric 0.6B parameter configuration---\texttt{d128}, \texttt{d256},
and \texttt{d512}---we conduct \emph{full-parameter CPT} (Full CPT) and \emph{attention-frozen CPT} (Frozen CPT) from PT checkpoints of three independent runs.
Within each run, both conditions start from the same 100B-token PT checkpoint and undergo 8B tokens of CPT at a sequence length of 256K.
In Frozen CPT, we fix $W_Q,W_K,W_V,W_O$ in both full-attention and
sliding-window attention layers and only update other parameters; see Table~\ref{tab:attention_freeze_policy} for the detailed parameter-update policies.
Note that fixed projections do not imply fixed attention maps: their inputs and normalization gains can still adapt.

We evaluate RULER-100, averaging over 13 tasks and seven
context lengths from 4K to 256K.
As a performance baseline, we evaluate PT checkpoints with inference-time attention logit scaling (\emph{scaled PT}; see Appendix~\ref{app:pt_temperature_scaling}), which requires no parameter updates.
Neither CPT condition uses this additional scaling.
Scores are averaged across three runs.
The result for each task and length is provided in Figure~\ref{fig:attention_freeze_ruler_by_length} and
Figures~\ref{fig:freeze_tasks_d128}--\ref{fig:freeze_tasks_d512}.

\begin{figure}[t]
    \hspace{-5mm}
    \begin{minipage}[t]{0.345\linewidth}
        \centering
        \includegraphics[width=\linewidth]{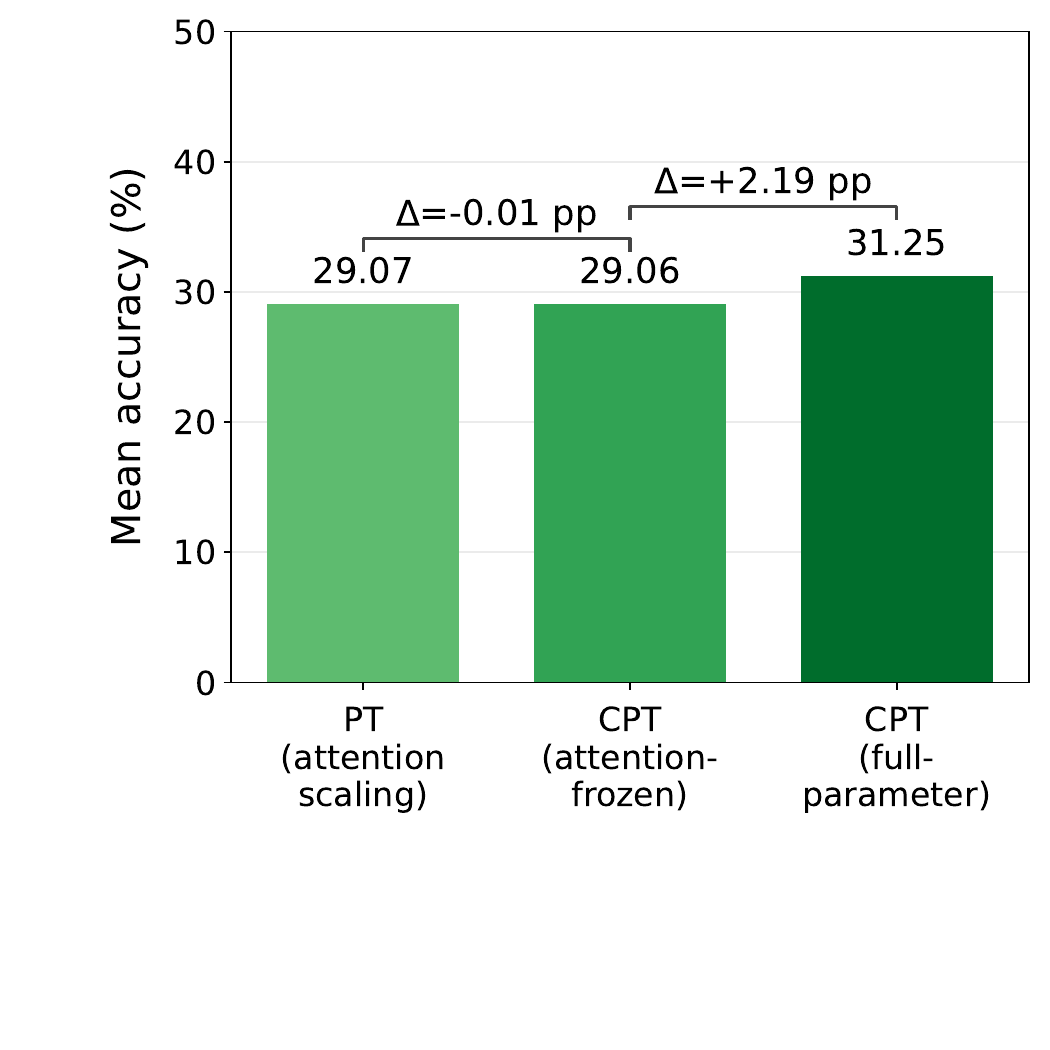}
        \vspace{-13mm}
        \par\small \hspace{5mm}(a) \texttt{0.6B-d128}
    \end{minipage}
    \hspace{-2.5mm}
    \begin{minipage}[t]{0.345\linewidth}
        \centering
        \includegraphics[width=\linewidth]{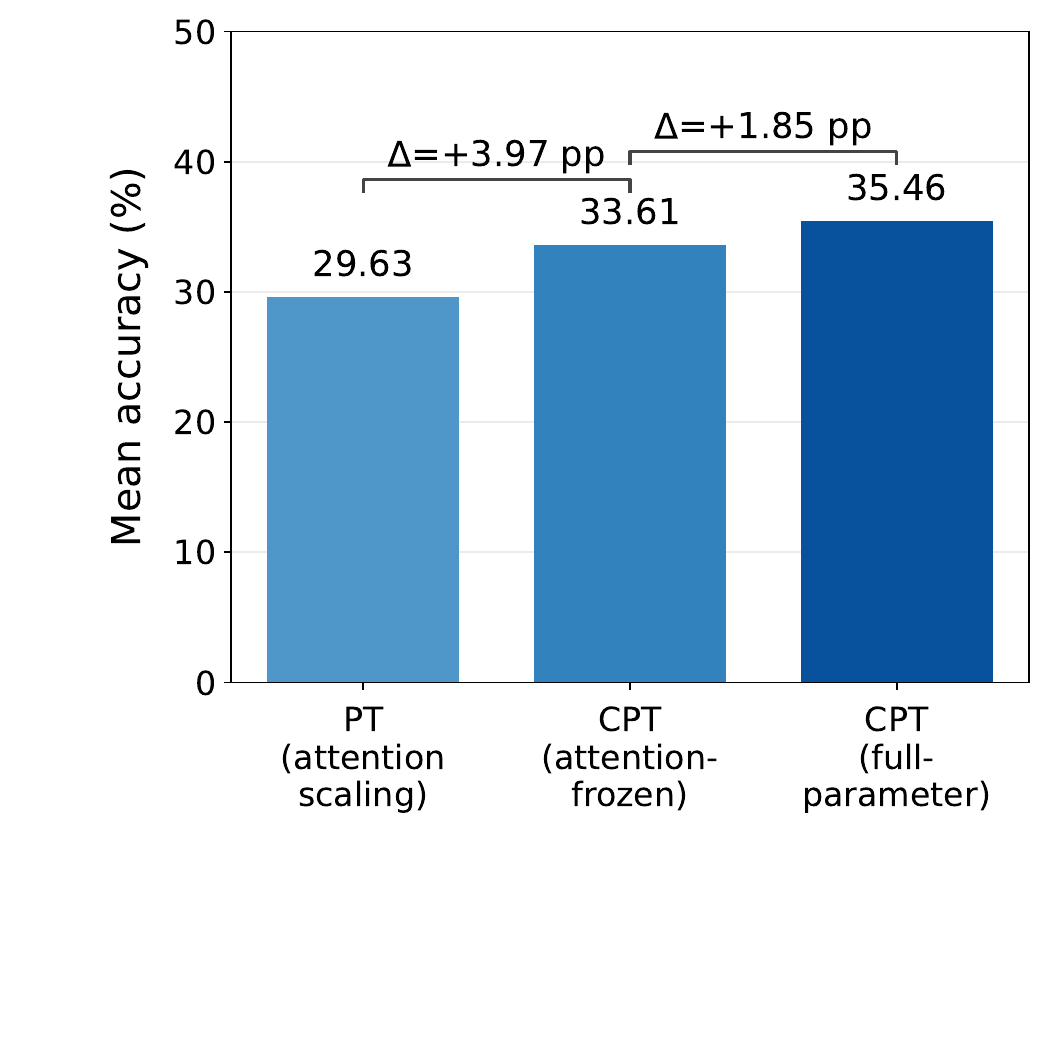}
        \vspace{-13mm}
        \par\small \hspace{5mm}(b) \texttt{0.6B-d256}
    \end{minipage}
    \hspace{-2.5mm}
    \begin{minipage}[t]{0.345\linewidth}
        \centering
        \includegraphics[width=\linewidth]{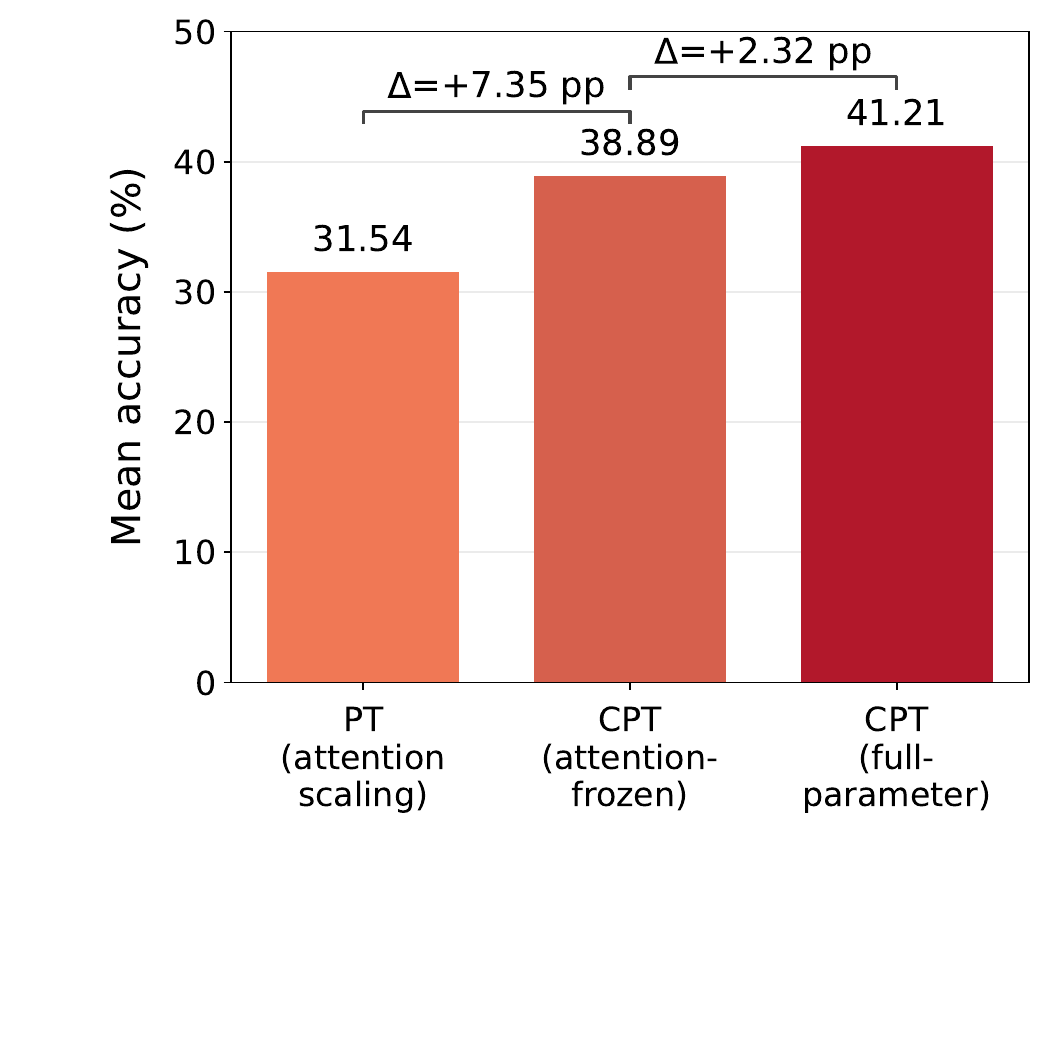}
        \vspace{-13mm}
        \par\small \hspace{5mm}(c) \texttt{0.6B-d512}
    \end{minipage}
    \vspace{-5mm}
    \caption{
    Retrieval performance with and without attention weight matrices updates for the three 0.6B-parameter configurations.
    Bars show means across three training runs, each evaluated and averaged over 13 RULER-100 tasks and seven context lengths from 4K to 256K.
    The PT baseline uses inference-time attention scaling; neither CPT condition uses additional scaling.
    Brackets show mean differences in score points, computed before rounding.}
    \label{fig:attention_freeze_ruler}
\end{figure}

\paragraph{The head-dimension dependence persists without projection updates.}
Figure~\ref{fig:attention_freeze_ruler} shows that the mean improvement
of Frozen CPT over scaled PT increases with head dimension:
$-0.01$, $+3.97$, and $+7.35$ points for \texttt{d128},
\texttt{d256}, and \texttt{d512}, respectively.
In contrast, the additional mean benefit of Full CPT over Frozen CPT
is similar across configurations: $2.19$, $1.85$, and $2.32$ points.
The strong head-dimension dependence therefore appears in performance gains
achievable while projections in attention modules remain fixed, rather than
in the incremental benefit of updating them.

Full CPT also increases the trainable parameter fraction from
$83.4\%$ to $100\%$, without changing model size.
Its roughly two-point advantage is compatible with a general benefit
from optimizing additional parameters, rather than a benefit that
grows with head dimension.
The comparison does not isolate trainable parameter count from
parameter identity, so it cannot establish that count alone causes
the improvement.
Nevertheless, these results do not support that the performance gap in retrieval tasks among models with different head dimensions is explained by the benefit of updates in attention weight matrices.

\subsection{How does head dimension affect attention maps and model representations?}
\label{sec:token_rank_reorganization}

\begin{figure}[t]
    \centering
    \begin{minipage}[t]{0.45\linewidth}
        \centering
        \includegraphics[width=\linewidth]{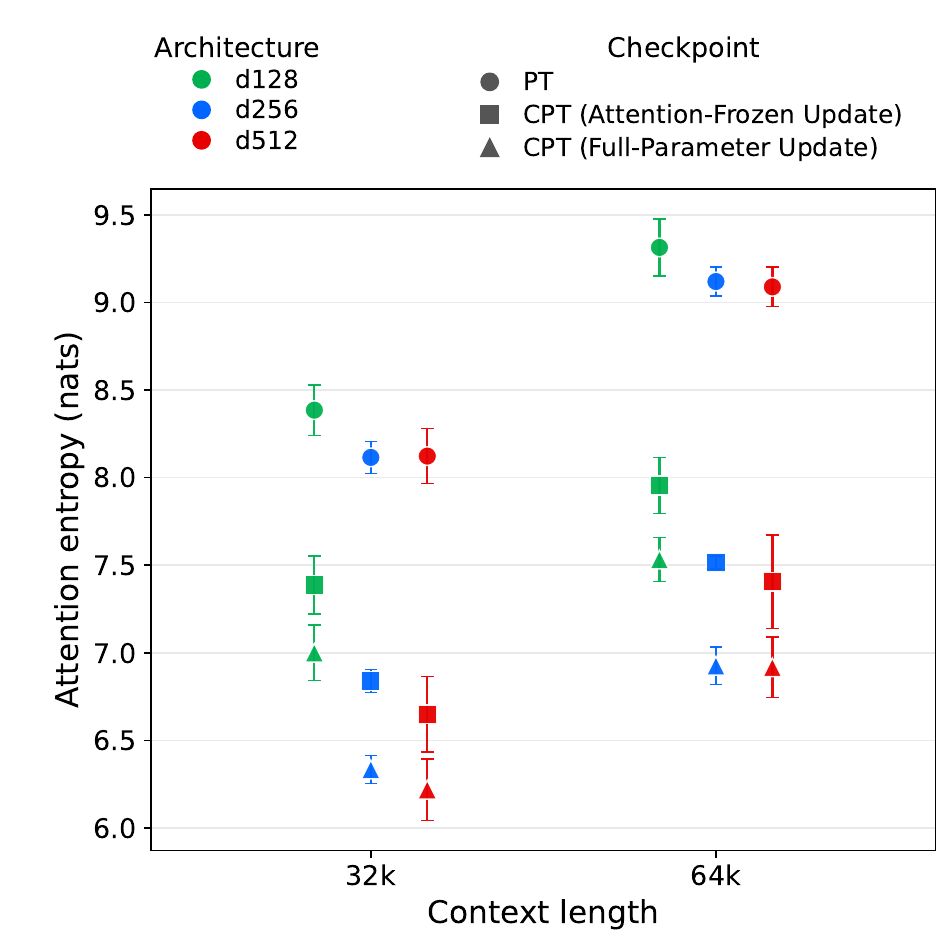}
        \par\small (a) attention entropy
    \end{minipage}
    \begin{minipage}[t]{0.45\linewidth}
        \centering
        \includegraphics[width=\linewidth]{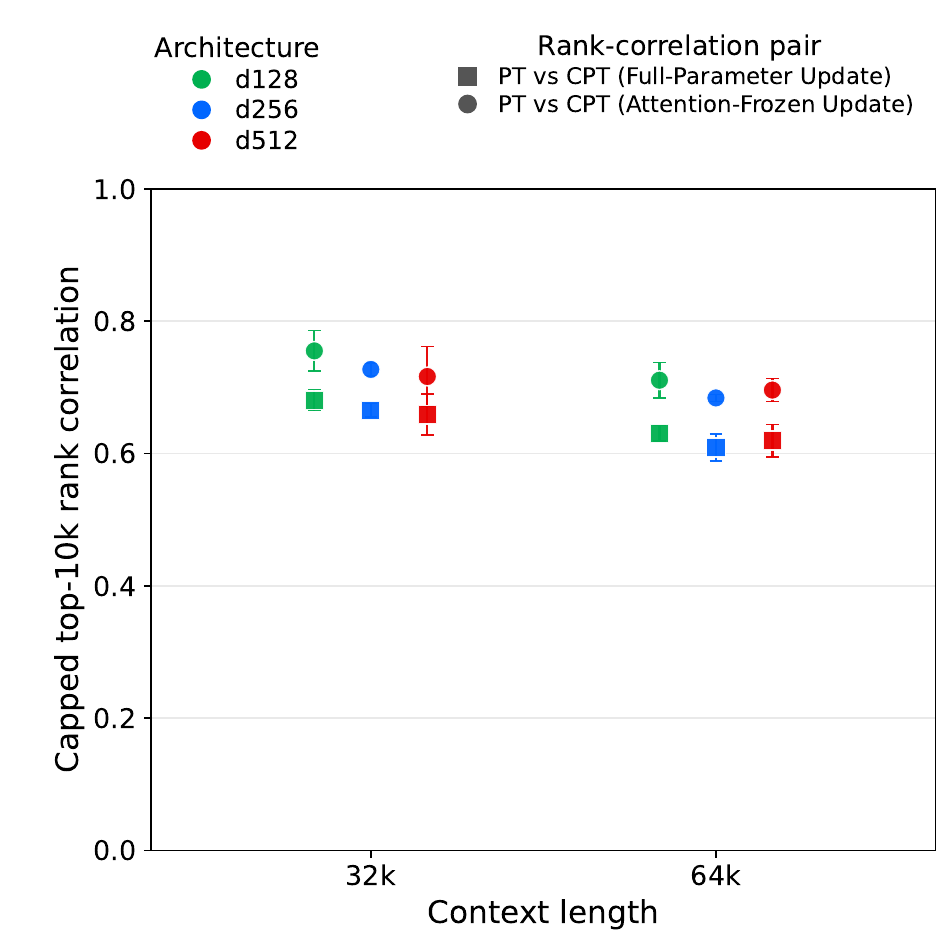}
        \par\small (b) rank correlation
    \end{minipage}
    \caption{
    Attention-map diagnostics on shared RULER examples at context lengths of 32K and 64K, using queries predicting the first three reference-answer tokens.
    (a) Mean attention entropy; lower values indicate more concentrated attention.
    (b) Capped top-10k rank correlation between each CPT checkpoint and the corresponding PT checkpoint; higher values indicate more similar token prioritization.
    PT checkpoints are evaluated without attention scaling.
    Points and error bars show means and $\pm1$ sample standard deviation across three training runs, after averaging within each run.
    }
    \label{fig:attention_map}
\end{figure}

\begin{figure}[t]
    \centering
    \includegraphics[width=0.7\linewidth]{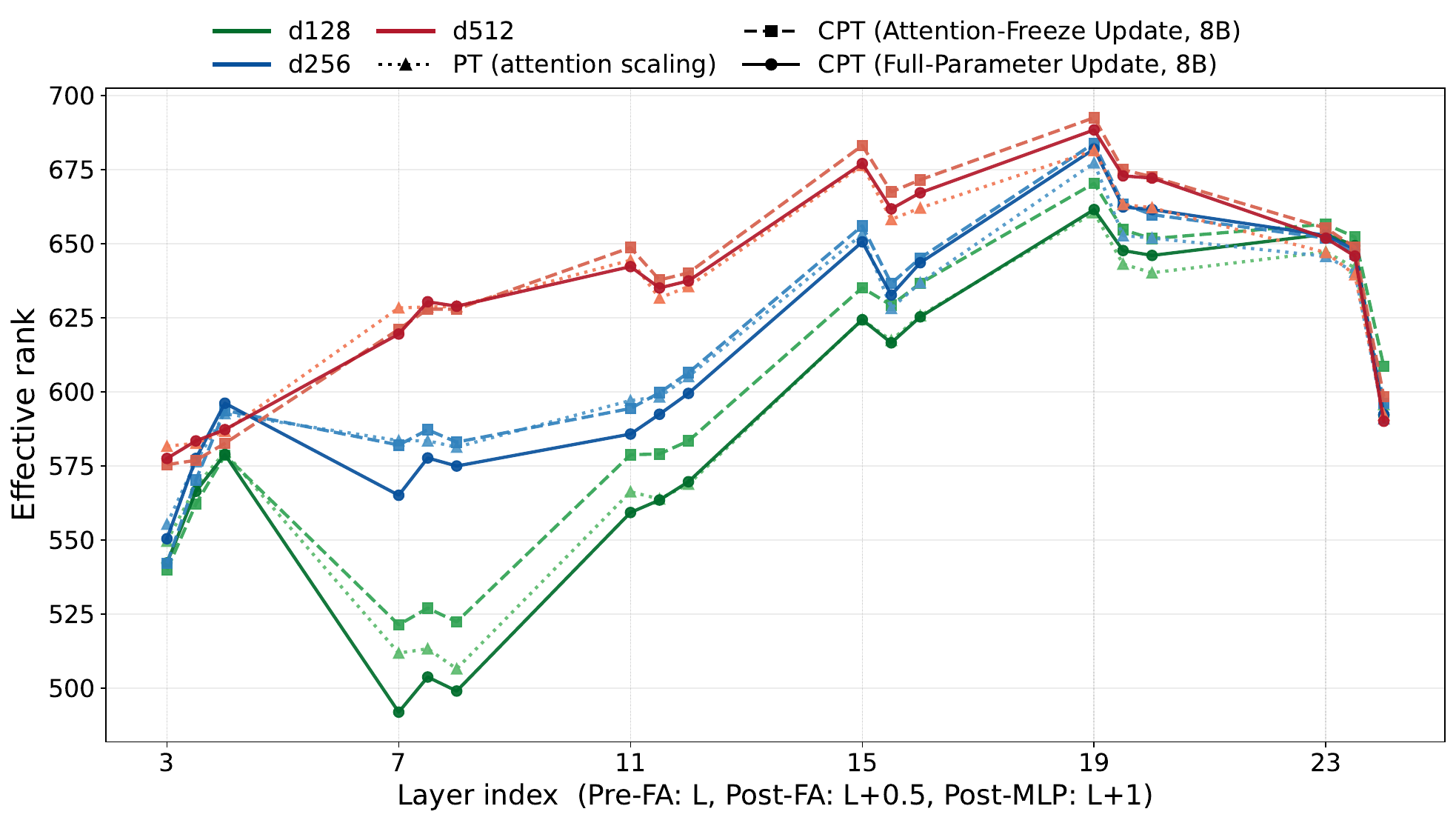}
    \caption{
    Residual-stream effective rank at 32K context length for the three symmetric 0.6B configurations under scaled PT, Frozen CPT, and Full CPT.
    Each point is computed from the mean singular-value spectrum over 300 shared examples and three training seeds.
    For a full-attention layer with zero-based index $L$, the effective ranks of representations at Pre-FA, Post-FA, and Post-MLP are plotted at the layer index $L$, $L+0.5$, and $L+1$, respectively.
    These offsets distinguish measurement sites within the same block, not additional layers.
    }
    \label{fig:residual_effective_rank}
\end{figure}

We next examine whether changes in attention maps explain the performance gains of wider heads under frozen attention weight matrices.
For Figure~\ref{fig:attention_map}, we compare PT without attention scaling with the corresponding 8B-token Full and Frozen CPT checkpoints on six RULER tasks, using 300 shared examples at each of 32K and 64K.
We analyze full-attention queries predicting the first three reference-answer tokens under teacher forcing.
We report means and sample standard deviations across three runs after within-run aggregation.
See Appendix~\ref{app:attention_map_setup} for details.

\paragraph{Lower entropy does not fully explain the wider-head advantage.}
For attention probabilities $a_j$ over causally visible token positions,
we measure entropy in nats as $H=-\sum_j a_j\log a_j$.
Lower entropy indicates more concentrated attention, but does not
establish whether the attended positions are relevant.
Figure~\ref{fig:attention_map}(a) shows lower mean entropy after CPT
for every head dimension, with Full CPT below Frozen CPT and
Frozen CPT below PT at both context lengths.
The higher post-CPT entropy of \texttt{d128} is consistent with the lower retrieval performance as it indicates less accurate information retrieval from the context.
However, \texttt{d256} and \texttt{d512} have similar mean entropies
under each CPT checkpoint despite their performance gap.
Taken together, these results suggest that sharper attention may contribute to retrieval improvements, but average probability concentration alone does not explain the head-dimension effect.

\paragraph{Greater token reordering does not consistently favor wider heads.}
We compare token prioritization between PT and each CPT checkpoint using \emph{capped top-10k rank correlation}.
For each matched layer, query head, and answer query on the same example, we rank all causally visible token positions by decreasing attention probability in each checkpoint.
We replace ranks beyond $K=10{,}000$ with $K+1$ and compute Pearson correlation between the resulting rank vectors, aligned by token position.
Thus, lower-ranked positions remain in the comparison, but differences in their ordering are ignored.
Lower correlation indicates greater changes in which tokens receive high attention and how they are ordered, rather than changes in probability magnitudes.

Figure~\ref{fig:attention_map}(b) shows greater reordering under Full CPT than under Frozen CPT, but differences across head dimensions are small and do not consistently track retrieval performance.
At 64K, \texttt{d256} exhibits more reordering than \texttt{d512}
under both CPT conditions despite its lower overall RULER-100 score.
Thus, this measure of token reordering does not fully explain the wider-head advantage.

\paragraph{Head dimension shapes the model's internal representations.}
We examine the effective rank \citep{roy2007effectiverank} of the residual stream at 32K context length for the three 0.6B configurations, using prompt-token representations before and after attention and after the MLP in each full-attention block.
Figure~\ref{fig:residual_effective_rank} shows that wider-head models generally have higher effective ranks in intermediate blocks, particularly at layers 7--19.
This separation is already visible in scaled PT and persists under Frozen CPT, so it is not specific to updates on projection weights in attention modules during CPT.
These results associate head allocation with substantial differences in residual-stream structure, showing that the influence of head dimension extends beyond attention modules to the internal representations propagated through the model.
See Appendix~\ref{app:effective_rank_setup} for the details.

\paragraph{Implications.}
Together, Figures~\ref{fig:attention_freeze_ruler} and~\ref{fig:attention_map} show that wider heads achieve larger mean retrieval performance gains even with fixed projection weights in attention modules, whereas the measured concentration and reordering of attention maps do not fully account for those gains.
Figure~\ref{fig:residual_effective_rank} further shows that head allocation is associated with differences in residual-stream spectral structure even under Frozen CPT.
These results point to an interaction between head allocation and adaptation of the remaining trainable components, including embeddings and MLPs, as a plausible explanation, rather than a larger direct benefit from updating attention weight matrices.

\section{Related work}

\paragraph{Head dimension and the allocation of attention capacity.}
Multi-head attention aggregates information through distinct attention distributions and representation subspaces \citep{vaswani2017attention}.
At fixed total projection width, increasing the number of heads reduces each head's dimension.
\citet{bhojanapalli2020lowrank} show that low-dimensional query--key parameterizations can restrict the attainable attention distributions, motivating head dimension as a capacity constraint rather than merely an implementation choice.
Conversely, theoretical advantages of multiple heads have been established for memorization and in-context linear regression \citep{mahdavi2024memorization,cui2024superiority}.
Head-count requirements also arise in restricted retrieval and multi-task computation settings \citep{yu2026headcount,hsu2026multitask}. These results establish task-dependent trade-offs, not a universally optimal head allocation.
Our experiments examine this trade-off for long-context language models trained from scratch, keeping total query and key/value projection widths and the per-token KV-cache size fixed. This differs from changing the degree of key/value sharing, as in grouped-query attention \citep{ainslie2023gqa}.
Note that, in most open-weight LLM families published with multiple parameter scales, the number of heads in attention layers is increased in larger models, while the dimension of each head is kept fixed \citep{llama3,qwen3,gemma4}.

\paragraph{Long-context training and architectural adaptation.}
Prior work improves long-context adaptation through data selection and domain balancing \citep{fu2024dataengineering,chen2024longdependency}, as well as CPT and instruction-tuning recipes \citep{gao2025prolong,xu2025ultralong}.
\citet{lu2024controlled} compare context-extension methods under standardized model and data conditions, highlighting the importance of controlled evaluation.
Examining learning dynamics, \citet{liang2026learningdynamics} observe increasing retrieval-head scores during CPT alongside largely stable identities of highly ranked heads in later training.
\citet{bertsch2026olmpool} further shows that architectural choices, including normalization and KV sharing, can jointly produce substantial differences in long-context performance that standard short-context metrics poorly predict.
Our study addresses how the allocation of attention capacity across heads affects retrieval and adaptation under a common training recipe.
We vary head dimension and head count while keeping total projection widths, the query-to-KV head ratio, and per-token KV-cache size fixed.

\section{Conclusion}
In this paper, we systematically compare language models with identical architectures except for the balance between the number of heads and the dimension of each head, demonstrating that wider heads improve long-context retrieval at fixed model sizes.
Our analyses suggest that this improvement is associated with (1) increased parameter association with retrieval heads, as wider-head models have comparable numbers of retrieval heads despite having fewer heads overall, and (2) more effective adaptation of components such as embeddings and MLPs during long-context continual pretraining, rather than attention-map changes alone.
The experiments conducted in this paper are limited to the head dimensions of at most 512; further increasing the head dimension requires improvements in efficient attention computation, which is left for future work.

\newpage

\subsection*{Acknowledgements}

We thank the National Institute of Information and Communications Technology (NICT) for the opportunity to conduct this research under our joint research framework.

\bibliography{main}

@inproceedings{
adamw,
title={Decoupled Weight Decay Regularization},
author={Ilya Loshchilov and Frank Hutter},
booktitle={International Conference on Learning Representations},
year={2019},
url={https://openreview.net/forum?id=Bkg6RiCqY7},
}

@article{arc,
      author    = {Peter Clark  and Isaac Cowhey and Oren Etzioni and Tushar Khot and
                    Ashish Sabharwal and Carissa Schoenick and Oyvind Tafjord},
      title     = {Think you have Solved Question Answering? Try ARC, the AI2 Reasoning Challenge},
      journal   = {arXiv:1803.05457v1},
      year      = {2018},
}

@article{chatqa2,
  title={ChatQA 2: Bridging the Gap to Proprietary LLMs in Long Context and RAG Capabilities},
  author={Xu, Peng and Ping, Wei and Wu, Xianchao and Liu, Zihan and Shoeybi, Mohammad and Catanzaro, Bryan},
  journal={arXiv preprint arXiv:2407.14482},
  year={2024}
}

@inproceedings{commonsenseqa,
    title = "{C}ommonsense{QA}: A Question Answering Challenge Targeting Commonsense Knowledge",
    author = "Talmor, Alon  and
      Herzig, Jonathan  and
      Lourie, Nicholas  and
      Berant, Jonathan",
    booktitle = "Proceedings of the 2019 Conference of the North {A}merican Chapter of the Association for Computational Linguistics: Human Language Technologies, Volume 1 (Long and Short Papers)",
    month = jun,
    year = "2019",
    address = "Minneapolis, Minnesota",
    publisher = "Association for Computational Linguistics",
    url = "https://aclanthology.org/N19-1421",
    doi = "10.18653/v1/N19-1421",
    pages = "4149--4158",
    archivePrefix = "arXiv",
    eprint        = "1811.00937",
    primaryClass  = "cs",
}

@inproceedings{helmet,
title={{HELMET}: How to Evaluate Long-context Models Effectively and Thoroughly},
author={Howard Yen and Tianyu Gao and Minmin Hou and Ke Ding and Daniel Fleischer and Peter Izsak and Moshe Wasserblat and Danqi Chen},
booktitle={The Thirteenth International Conference on Learning Representations},
year={2025},
url={https://openreview.net/forum?id=293V3bJbmE}
}

@misc{gemma4,
      title={Gemma 4 Technical Report}, 
      author={Gemma Team},
      year={2026},
      eprint={2607.02770},
      archivePrefix={arXiv},
      primaryClass={cs.CL},
      url={https://arxiv.org/abs/2607.02770},
}

@inproceedings{hellaswag,
    title={HellaSwag: Can a Machine Really Finish Your Sentence?},
    author={Zellers, Rowan and Holtzman, Ari and Bisk, Yonatan and Farhadi, Ali and Choi, Yejin},
    booktitle ={Proceedings of the 57th Annual Meeting of the Association for Computational Linguistics},
    year={2019}
}

@inproceedings{icr2,
    title = "Eliciting In-context Retrieval and Reasoning for Long-context Large Language Models",
    author = "Qiu, Yifu  and
      Embar, Varun R.  and
      Zhang, Yizhe  and
      Jaitly, Navdeep  and
      Cohen, Shay B.  and
      Han, Benjamin",
    editor = "Che, Wanxiang  and
      Nabende, Joyce  and
      Shutova, Ekaterina  and
      Pilehvar, Mohammad Taher",
    booktitle = "Findings of the Association for Computational Linguistics: ACL 2025",
    month = jul,
    year = "2025",
    address = "Vienna, Austria",
    publisher = "Association for Computational Linguistics",
    url = "https://aclanthology.org/2025.findings-acl.165/",
    doi = "10.18653/v1/2025.findings-acl.165",
    pages = "3176--3192",
    ISBN = "979-8-89176-256-5"
}

@misc{llama3,
      title={The Llama 3 Herd of Models}, 
      author={Aaron Grattafiori and Abhimanyu Dubey and Abhinav Jauhri and Abhinav Pandey and Abhishek Kadian and Ahmad Al-Dahle and Aiesha Letman and Akhil Mathur and Alan Schelten and Alex Vaughan and Amy Yang and Angela Fan and Anirudh Goyal and Anthony Hartshorn and Aobo Yang and Archi Mitra and Archie Sravankumar and Artem Korenev and Arthur Hinsvark and Arun Rao and Aston Zhang and Aurelien Rodriguez and Austen Gregerson and Ava Spataru and Baptiste Roziere and Bethany Biron and Binh Tang and Bobbie Chern and Charlotte Caucheteux and Chaya Nayak and Chloe Bi and Chris Marra and Chris McConnell and Christian Keller and Christophe Touret and Chunyang Wu and Corinne Wong and Cristian Canton Ferrer and Cyrus Nikolaidis and Damien Allonsius and Daniel Song and Danielle Pintz and Danny Livshits and Danny Wyatt and David Esiobu and Dhruv Choudhary and Dhruv Mahajan and Diego Garcia-Olano and Diego Perino and Dieuwke Hupkes and Egor Lakomkin and Ehab AlBadawy and Elina Lobanova and Emily Dinan and Eric Michael Smith and Filip Radenovic and Francisco Guzmán and Frank Zhang and Gabriel Synnaeve and Gabrielle Lee and Georgia Lewis Anderson and Govind Thattai and Graeme Nail and Gregoire Mialon and Guan Pang and Guillem Cucurell and Hailey Nguyen and Hannah Korevaar and Hu Xu and Hugo Touvron and Iliyan Zarov and Imanol Arrieta Ibarra and Isabel Kloumann and Ishan Misra and Ivan Evtimov and Jack Zhang and Jade Copet and Jaewon Lee and Jan Geffert and Jana Vranes and Jason Park and Jay Mahadeokar and Jeet Shah and Jelmer van der Linde and Jennifer Billock and Jenny Hong and Jenya Lee and Jeremy Fu and Jianfeng Chi and Jianyu Huang and Jiawen Liu and Jie Wang and Jiecao Yu and Joanna Bitton and Joe Spisak and Jongsoo Park and Joseph Rocca and Joshua Johnstun and Joshua Saxe and Junteng Jia and Kalyan Vasuden Alwala and Karthik Prasad and Kartikeya Upasani and Kate Plawiak and Ke Li and Kenneth Heafield and Kevin Stone and Khalid El-Arini and Krithika Iyer and Kshitiz Malik and Kuenley Chiu and Kunal Bhalla and Kushal Lakhotia and Lauren Rantala-Yeary and Laurens van der Maaten and Lawrence Chen and Liang Tan and Liz Jenkins and Louis Martin and Lovish Madaan and Lubo Malo and Lukas Blecher and Lukas Landzaat and Luke de Oliveira and Madeline Muzzi and Mahesh Pasupuleti and Mannat Singh and Manohar Paluri and Marcin Kardas and Maria Tsimpoukelli and Mathew Oldham and Mathieu Rita and Maya Pavlova and Melanie Kambadur and Mike Lewis and Min Si and Mitesh Kumar Singh and Mona Hassan and Naman Goyal and Narjes Torabi and Nikolay Bashlykov and Nikolay Bogoychev and Niladri Chatterji and Ning Zhang and Olivier Duchenne and Onur Çelebi and Patrick Alrassy and Pengchuan Zhang and Pengwei Li and Petar Vasic and Peter Weng and Prajjwal Bhargava and Pratik Dubal and Praveen Krishnan and Punit Singh Koura and Puxin Xu and Qing He and Qingxiao Dong and Ragavan Srinivasan and Raj Ganapathy and Ramon Calderer and Ricardo Silveira Cabral and Robert Stojnic and Roberta Raileanu and Rohan Maheswari and Rohit Girdhar and Rohit Patel and Romain Sauvestre and Ronnie Polidoro and Roshan Sumbaly and Ross Taylor and Ruan Silva and Rui Hou and Rui Wang and Saghar Hosseini and Sahana Chennabasappa and Sanjay Singh and Sean Bell and Seohyun Sonia Kim and Sergey Edunov and Shaoliang Nie and Sharan Narang and Sharath Raparthy and Sheng Shen and Shengye Wan and Shruti Bhosale and Shun Zhang and Simon Vandenhende and Soumya Batra and Spencer Whitman and Sten Sootla and Stephane Collot and Suchin Gururangan and Sydney Borodinsky and Tamar Herman and Tara Fowler and Tarek Sheasha and Thomas Georgiou and Thomas Scialom and Tobias Speckbacher and Todor Mihaylov and Tong Xiao and Ujjwal Karn and Vedanuj Goswami and Vibhor Gupta and Vignesh Ramanathan and Viktor Kerkez and Vincent Gonguet and Virginie Do and Vish Vogeti and Vítor Albiero and Vladan Petrovic and Weiwei Chu and Wenhan Xiong and Wenyin Fu and Whitney Meers and Xavier Martinet and Xiaodong Wang and Xiaofang Wang and Xiaoqing Ellen Tan and Xide Xia and Xinfeng Xie and Xuchao Jia and Xuewei Wang and Yaelle Goldschlag and Yashesh Gaur and Yasmine Babaei and Yi Wen and Yiwen Song and Yuchen Zhang and Yue Li and Yuning Mao and Zacharie Delpierre Coudert and Zheng Yan and Zhengxing Chen and Zoe Papakipos and Aaditya Singh and Aayushi Srivastava and Abha Jain and Adam Kelsey and Adam Shajnfeld and Adithya Gangidi and Adolfo Victoria and Ahuva Goldstand and Ajay Menon and Ajay Sharma and Alex Boesenberg and Alexei Baevski and Allie Feinstein and Amanda Kallet and Amit Sangani and Amos Teo and Anam Yunus and Andrei Lupu and Andres Alvarado and Andrew Caples and Andrew Gu and Andrew Ho and Andrew Poulton and Andrew Ryan and Ankit Ramchandani and Annie Dong and Annie Franco and Anuj Goyal and Aparajita Saraf and Arkabandhu Chowdhury and Ashley Gabriel and Ashwin Bharambe and Assaf Eisenman and Azadeh Yazdan and Beau James and Ben Maurer and Benjamin Leonhardi and Bernie Huang and Beth Loyd and Beto De Paola and Bhargavi Paranjape and Bing Liu and Bo Wu and Boyu Ni and Braden Hancock and Bram Wasti and Brandon Spence and Brani Stojkovic and Brian Gamido and Britt Montalvo and Carl Parker and Carly Burton and Catalina Mejia and Ce Liu and Changhan Wang and Changkyu Kim and Chao Zhou and Chester Hu and Ching-Hsiang Chu and Chris Cai and Chris Tindal and Christoph Feichtenhofer and Cynthia Gao and Damon Civin and Dana Beaty and Daniel Kreymer and Daniel Li and David Adkins and David Xu and Davide Testuggine and Delia David and Devi Parikh and Diana Liskovich and Didem Foss and Dingkang Wang and Duc Le and Dustin Holland and Edward Dowling and Eissa Jamil and Elaine Montgomery and Eleonora Presani and Emily Hahn and Emily Wood and Eric-Tuan Le and Erik Brinkman and Esteban Arcaute and Evan Dunbar and Evan Smothers and Fei Sun and Felix Kreuk and Feng Tian and Filippos Kokkinos and Firat Ozgenel and Francesco Caggioni and Frank Kanayet and Frank Seide and Gabriela Medina Florez and Gabriella Schwarz and Gada Badeer and Georgia Swee and Gil Halpern and Grant Herman and Grigory Sizov and Guangyi and Zhang and Guna Lakshminarayanan and Hakan Inan and Hamid Shojanazeri and Han Zou and Hannah Wang and Hanwen Zha and Haroun Habeeb and Harrison Rudolph and Helen Suk and Henry Aspegren and Hunter Goldman and Hongyuan Zhan and Ibrahim Damlaj and Igor Molybog and Igor Tufanov and Ilias Leontiadis and Irina-Elena Veliche and Itai Gat and Jake Weissman and James Geboski and James Kohli and Janice Lam and Japhet Asher and Jean-Baptiste Gaya and Jeff Marcus and Jeff Tang and Jennifer Chan and Jenny Zhen and Jeremy Reizenstein and Jeremy Teboul and Jessica Zhong and Jian Jin and Jingyi Yang and Joe Cummings and Jon Carvill and Jon Shepard and Jonathan McPhie and Jonathan Torres and Josh Ginsburg and Junjie Wang and Kai Wu and Kam Hou U and Karan Saxena and Kartikay Khandelwal and Katayoun Zand and Kathy Matosich and Kaushik Veeraraghavan and Kelly Michelena and Keqian Li and Kiran Jagadeesh and Kun Huang and Kunal Chawla and Kyle Huang and Lailin Chen and Lakshya Garg and Lavender A and Leandro Silva and Lee Bell and Lei Zhang and Liangpeng Guo and Licheng Yu and Liron Moshkovich and Luca Wehrstedt and Madian Khabsa and Manav Avalani and Manish Bhatt and Martynas Mankus and Matan Hasson and Matthew Lennie and Matthias Reso and Maxim Groshev and Maxim Naumov and Maya Lathi and Meghan Keneally and Miao Liu and Michael L. Seltzer and Michal Valko and Michelle Restrepo and Mihir Patel and Mik Vyatskov and Mikayel Samvelyan and Mike Clark and Mike Macey and Mike Wang and Miquel Jubert Hermoso and Mo Metanat and Mohammad Rastegari and Munish Bansal and Nandhini Santhanam and Natascha Parks and Natasha White and Navyata Bawa and Nayan Singhal and Nick Egebo and Nicolas Usunier and Nikhil Mehta and Nikolay Pavlovich Laptev and Ning Dong and Norman Cheng and Oleg Chernoguz and Olivia Hart and Omkar Salpekar and Ozlem Kalinli and Parkin Kent and Parth Parekh and Paul Saab and Pavan Balaji and Pedro Rittner and Philip Bontrager and Pierre Roux and Piotr Dollar and Polina Zvyagina and Prashant Ratanchandani and Pritish Yuvraj and Qian Liang and Rachad Alao and Rachel Rodriguez and Rafi Ayub and Raghotham Murthy and Raghu Nayani and Rahul Mitra and Rangaprabhu Parthasarathy and Raymond Li and Rebekkah Hogan and Robin Battey and Rocky Wang and Russ Howes and Ruty Rinott and Sachin Mehta and Sachin Siby and Sai Jayesh Bondu and Samyak Datta and Sara Chugh and Sara Hunt and Sargun Dhillon and Sasha Sidorov and Satadru Pan and Saurabh Mahajan and Saurabh Verma and Seiji Yamamoto and Sharadh Ramaswamy and Shaun Lindsay and Shaun Lindsay and Sheng Feng and Shenghao Lin and Shengxin Cindy Zha and Shishir Patil and Shiva Shankar and Shuqiang Zhang and Shuqiang Zhang and Sinong Wang and Sneha Agarwal and Soji Sajuyigbe and Soumith Chintala and Stephanie Max and Stephen Chen and Steve Kehoe and Steve Satterfield and Sudarshan Govindaprasad and Sumit Gupta and Summer Deng and Sungmin Cho and Sunny Virk and Suraj Subramanian and Sy Choudhury and Sydney Goldman and Tal Remez and Tamar Glaser and Tamara Best and Thilo Koehler and Thomas Robinson and Tianhe Li and Tianjun Zhang and Tim Matthews and Timothy Chou and Tzook Shaked and Varun Vontimitta and Victoria Ajayi and Victoria Montanez and Vijai Mohan and Vinay Satish Kumar and Vishal Mangla and Vlad Ionescu and Vlad Poenaru and Vlad Tiberiu Mihailescu and Vladimir Ivanov and Wei Li and Wenchen Wang and Wenwen Jiang and Wes Bouaziz and Will Constable and Xiaocheng Tang and Xiaojian Wu and Xiaolan Wang and Xilun Wu and Xinbo Gao and Yaniv Kleinman and Yanjun Chen and Ye Hu and Ye Jia and Ye Qi and Yenda Li and Yilin Zhang and Ying Zhang and Yossi Adi and Youngjin Nam and Yu and Wang and Yu Zhao and Yuchen Hao and Yundi Qian and Yunlu Li and Yuzi He and Zach Rait and Zachary DeVito and Zef Rosnbrick and Zhaoduo Wen and Zhenyu Yang and Zhiwei Zhao and Zhiyu Ma},
      year={2024},
      eprint={2407.21783},
      archivePrefix={arXiv},
      primaryClass={cs.AI},
      url={https://arxiv.org/abs/2407.21783}, 
}

@article{litm,
    title = "Lost in the Middle: How Language Models Use Long Contexts",
    author = "Liu, Nelson F.  and
      Lin, Kevin  and
      Hewitt, John  and
      Paranjape, Ashwin  and
      Bevilacqua, Michele  and
      Petroni, Fabio  and
      Liang, Percy",
    journal = "Transactions of the Association for Computational Linguistics",
    volume = "12",
    year = "2024",
    address = "Cambridge, MA",
    publisher = "MIT Press",
    url = "https://aclanthology.org/2024.tacl-1.9/",
    doi = "10.1162/tacl_a_00638",
    pages = "157--173"
}

@misc{nemotronnano2,
      title={NVIDIA Nemotron Nano 2: An Accurate and Efficient Hybrid Mamba-Transformer Reasoning Model}, 
      author={NVIDIA : Aarti Basant and Abhijit Khairnar and Abhijit Paithankar and Abhinav Khattar and Adithya Renduchintala and Aditya Malte and Akhiad Bercovich and Akshay Hazare and Alejandra Rico and Aleksander Ficek and Alex Kondratenko and Alex Shaposhnikov and Alexander Bukharin and Ali Taghibakhshi and Amelia Barton and Ameya Sunil Mahabaleshwarkar and Amy Shen and Andrew Tao and Ann Guan and Anna Shors and Anubhav Mandarwal and Arham Mehta and Arun Venkatesan and Ashton Sharabiani and Ashwath Aithal and Ashwin Poojary and Ayush Dattagupta and Balaram Buddharaju and Banghua Zhu and Barnaby Simkin and Bilal Kartal and Bita Darvish Rouhani and Bobby Chen and Boris Ginsburg and Brandon Norick and Brian Yu and Bryan Catanzaro and Charles Wang and Charlie Truong and Chetan Mungekar and Chintan Patel and Chris Alexiuk and Christian Munley and Christopher Parisien and Dan Su and Daniel Afrimi and Daniel Korzekwa and Daniel Rohrer and Daria Gitman and David Mosallanezhad and Deepak Narayanan and Dima Rekesh and Dina Yared and Dmytro Pykhtar and Dong Ahn and Duncan Riach and Eileen Long and Elliott Ning and Eric Chung and Erick Galinkin and Evelina Bakhturina and Gargi Prasad and Gerald Shen and Haifeng Qian and Haim Elisha and Harsh Sharma and Hayley Ross and Helen Ngo and Herman Sahota and Hexin Wang and Hoo Chang Shin and Hua Huang and Iain Cunningham and Igor Gitman and Ivan Moshkov and Jaehun Jung and Jan Kautz and Jane Polak Scowcroft and Jared Casper and Jian Zhang and Jiaqi Zeng and Jimmy Zhang and Jinze Xue and Jocelyn Huang and Joey Conway and John Kamalu and Jonathan Cohen and Joseph Jennings and Julien Veron Vialard and Junkeun Yi and Jupinder Parmar and Kari Briski and Katherine Cheung and Katherine Luna and Keith Wyss and Keshav Santhanam and Kezhi Kong and Krzysztof Pawelec and Kumar Anik and Kunlun Li and Kushan Ahmadian and Lawrence McAfee and Laya Sleiman and Leon Derczynski and Luis Vega and Maer Rodrigues de Melo and Makesh Narsimhan Sreedhar and Marcin Chochowski and Mark Cai and Markus Kliegl and Marta Stepniewska-Dziubinska and Matvei Novikov and Mehrzad Samadi and Meredith Price and Meriem Boubdir and Michael Boone and Michael Evans and Michal Bien and Michal Zawalski and Miguel Martinez and Mike Chrzanowski and Mohammad Shoeybi and Mostofa Patwary and Namit Dhameja and Nave Assaf and Negar Habibi and Nidhi Bhatia and Nikki Pope and Nima Tajbakhsh and Nirmal Kumar Juluru and Oleg Rybakov and Oleksii Hrinchuk and Oleksii Kuchaiev and Oluwatobi Olabiyi and Pablo Ribalta and Padmavathy Subramanian and Parth Chadha and Pavlo Molchanov and Peter Dykas and Peter Jin and Piotr Bialecki and Piotr Januszewski and Pradeep Thalasta and Prashant Gaikwad and Prasoon Varshney and Pritam Gundecha and Przemek Tredak and Rabeeh Karimi Mahabadi and Rajen Patel and Ran El-Yaniv and Ranjit Rajan and Ria Cheruvu and Rima Shahbazyan and Ritika Borkar and Ritu Gala and Roger Waleffe and Ruoxi Zhang and Russell J. Hewett and Ryan Prenger and Sahil Jain and Samuel Kriman and Sanjeev Satheesh and Saori Kaji and Sarah Yurick and Saurav Muralidharan and Sean Narenthiran and Seonmyeong Bak and Sepehr Sameni and Seungju Han and Shanmugam Ramasamy and Shaona Ghosh and Sharath Turuvekere Sreenivas and Shelby Thomas and Shizhe Diao and Shreya Gopal and Shrimai Prabhumoye and Shubham Toshniwal and Shuoyang Ding and Siddharth Singh and Siddhartha Jain and Somshubra Majumdar and Soumye Singhal and Stefania Alborghetti and Syeda Nahida Akter and Terry Kong and Tim Moon and Tomasz Hliwiak and Tomer Asida and Tony Wang and Tugrul Konuk and Twinkle Vashishth and Tyler Poon and Udi Karpas and Vahid Noroozi and Venkat Srinivasan and Vijay Korthikanti and Vikram Fugro and Vineeth Kalluru and Vitaly Kurin and Vitaly Lavrukhin and Wasi Uddin Ahmad and Wei Du and Wonmin Byeon and Ximing Lu and Xin Dong and Yashaswi Karnati and Yejin Choi and Yian Zhang and Ying Lin and Yonggan Fu and Yoshi Suhara and Zhen Dong and Zhiyu Li and Zhongbo Zhu and Zijia Chen},
      year={2025},
      eprint={2508.14444},
      archivePrefix={arXiv},
      primaryClass={cs.CL},
      url={https://arxiv.org/abs/2508.14444}, 
}

@inproceedings{openbookqa,
    title={Can a Suit of Armor Conduct Electricity? A New Dataset for Open Book Question Answering},
    author={Todor Mihaylov and Peter Clark and Tushar Khot and Ashish Sabharwal},
    booktitle={EMNLP},
    year={2018}
}

@inproceedings{piqa,
    author = {Yonatan Bisk and Rowan Zellers and
            Ronan Le Bras and Jianfeng Gao
            and Yejin Choi},
    title = {PIQA: Reasoning about Physical Commonsense in
           Natural Language},
    booktitle = {Thirty-Fourth AAAI Conference on
               Artificial Intelligence},
    year = {2020},
}

@inproceedings{qrhead,
    title = "Query-Focused Retrieval Heads Improve Long-Context Reasoning and Re-ranking",
    author = "Zhang, Wuwei  and
      Yin, Fangcong  and
      Yen, Howard  and
      Chen, Danqi  and
      Ye, Xi",
    editor = "Christodoulopoulos, Christos  and
      Chakraborty, Tanmoy  and
      Rose, Carolyn  and
      Peng, Violet",
    booktitle = "Proceedings of the 2025 Conference on Empirical Methods in Natural Language Processing",
    month = nov,
    year = "2025",
    address = "Suzhou, China",
    publisher = "Association for Computational Linguistics",
    url = "https://aclanthology.org/2025.emnlp-main.1214/",
    doi = "10.18653/v1/2025.emnlp-main.1214",
    pages = "23791--23805",
    ISBN = "979-8-89176-332-6"
}

@misc{qwen3,
      title={Qwen3 Technical Report}, 
      author={An Yang and Anfeng Li and Baosong Yang and Beichen Zhang and Binyuan Hui and Bo Zheng and Bowen Yu and Chang Gao and Chengen Huang and Chenxu Lv and Chujie Zheng and Dayiheng Liu and Fan Zhou and Fei Huang and Feng Hu and Hao Ge and Haoran Wei and Huan Lin and Jialong Tang and Jian Yang and Jianhong Tu and Jianwei Zhang and Jianxin Yang and Jiaxi Yang and Jing Zhou and Jingren Zhou and Junyang Lin and Kai Dang and Keqin Bao and Kexin Yang and Le Yu and Lianghao Deng and Mei Li and Mingfeng Xue and Mingze Li and Pei Zhang and Peng Wang and Qin Zhu and Rui Men and Ruize Gao and Shixuan Liu and Shuang Luo and Tianhao Li and Tianyi Tang and Wenbiao Yin and Xingzhang Ren and Xinyu Wang and Xinyu Zhang and Xuancheng Ren and Yang Fan and Yang Su and Yichang Zhang and Yinger Zhang and Yu Wan and Yuqiong Liu and Zekun Wang and Zeyu Cui and Zhenru Zhang and Zhipeng Zhou and Zihan Qiu},
      year={2025},
      eprint={2505.09388},
      archivePrefix={arXiv},
      primaryClass={cs.CL},
      url={https://arxiv.org/abs/2505.09388}, 
}

@inproceedings{race,
    title = "{RACE}: Large-scale {R}e{A}ding Comprehension Dataset From Examinations",
    author = "Lai, Guokun  and
      Xie, Qizhe  and
      Liu, Hanxiao  and
      Yang, Yiming  and
      Hovy, Eduard",
    editor = "Palmer, Martha  and
      Hwa, Rebecca  and
      Riedel, Sebastian",
    booktitle = "Proceedings of the 2017 Conference on Empirical Methods in Natural Language Processing",
    month = sep,
    year = "2017",
    address = "Copenhagen, Denmark",
    publisher = "Association for Computational Linguistics",
    url = "https://aclanthology.org/D17-1082",
    doi = "10.18653/v1/D17-1082",
    pages = "785--794"
}

@inproceedings{retrievalhead,
title={Retrieval Head Mechanistically Explains Long-Context Factuality},
author={Wenhao Wu and Yizhong Wang and Guangxuan Xiao and Hao Peng and Yao Fu},
booktitle={The Thirteenth International Conference on Learning Representations},
year={2025},
url={https://openreview.net/forum?id=EytBpUGB1Z}
}

@inproceedings{rmsnorm,
 author = {Zhang, Biao and Sennrich, Rico},
 booktitle = {Advances in Neural Information Processing Systems},
 editor = {H. Wallach and H. Larochelle and A. Beygelzimer and F. d\textquotesingle Alch\'{e}-Buc and E. Fox and R. Garnett},
 pages = {},
 publisher = {Curran Associates, Inc.},
 title = {Root Mean Square Layer Normalization},
 url = {https://proceedings.neurips.cc/paper_files/paper/2019/file/1e8a19426224ca89e83cef47f1e7f53b-Paper.pdf},
 volume = {32},
 year = {2019}
}

@inproceedings{ruler,
title={{RULER}: What{\textquoteright}s the Real Context Size of Your Long-Context Language Models?},
author={Cheng-Ping Hsieh and Simeng Sun and Samuel Kriman and Shantanu Acharya and Dima Rekesh and Fei Jia and Yang Zhang and Boris Ginsburg},
booktitle={First Conference on Language Modeling},
year={2024},
url={https://openreview.net/forum?id=kIoBbc76Sy}
}

@misc{s1,
      title={s1: Simple test-time scaling}, 
      author={Niklas Muennighoff and Zitong Yang and Weijia Shi and Xiang Lisa Li and Li Fei-Fei and Hannaneh Hajishirzi and Luke Zettlemoyer and Percy Liang and Emmanuel Candès and Tatsunori Hashimoto},
      year={2025},
      eprint={2501.19393},
      archivePrefix={arXiv},
      primaryClass={cs.CL},
      url={https://arxiv.org/abs/2501.19393}, 
}

@misc{swallowcode,
      title={Rewriting Pre-Training Data Boosts LLM Performance in Math and Code}, 
      author={Kazuki Fujii and Yukito Tajima and Sakae Mizuki and Hinari Shimada and Taihei Shiotani and Koshiro Saito and Masanari Ohi and Masaki Kawamura and Taishi Nakamura and Takumi Okamoto and Shigeki Ishida and Kakeru Hattori and Youmi Ma and Hiroya Takamura and Rio Yokota and Naoaki Okazaki},
      year={2025},
      eprint={2505.02881},
      archivePrefix={arXiv},
      primaryClass={cs.LG},
      url={https://arxiv.org/abs/2505.02881}, 
}

@article{winogrande,
    title={WinoGrande: An Adversarial Winograd Schema Challenge at Scale},
    author={Sakaguchi, Keisuke and Bras, Ronan Le and Bhagavatula, Chandra and Choi, Yejin},
    journal={arXiv preprint arXiv:1907.10641},
    year={2019}
}

@inproceedings{vaswani2017attention,
  title = {Attention Is All You Need},
  author = {Ashish Vaswani and Noam Shazeer and Niki Parmar and Jakob Uszkoreit and Llion Jones and Aidan N. Gomez and Lukasz Kaiser and Illia Polosukhin},
  year = {2017},
  booktitle = {Advances in Neural Information Processing Systems 30},
  eprint = {1706.03762},
  archivePrefix = {arXiv},
  url = {https://arxiv.org/abs/1706.03762}
}

@inproceedings{bhojanapalli2020lowrank,
  title = {Low-Rank Bottleneck in Multi-head Attention Models},
  author = {Srinadh Bhojanapalli and Chulhee Yun and Ankit Singh Rawat and Sashank J. Reddi and Sanjiv Kumar},
  year = {2020},
  booktitle = {Proceedings of the 37th International Conference on Machine Learning},
  eprint = {2002.07028},
  archivePrefix = {arXiv},
  url = {https://arxiv.org/abs/2002.07028}
}

@inproceedings{ainslie2023gqa,
  title = {{GQA}: Training Generalized Multi-Query Transformer Models from Multi-Head Checkpoints},
  author = {Joshua Ainslie and James Lee-Thorp and Michiel de Jong and Yury Zemlyanskiy and Federico Lebron and Sumit Sanghai},
  year = {2023},
  booktitle = {Proceedings of the 2023 Conference on Empirical Methods in Natural Language Processing},
  eprint = {2305.13245},
  archivePrefix = {arXiv},
  url = {https://arxiv.org/abs/2305.13245}
}

@inproceedings{yu2026headcount,
  title = {The Effect of Attention Head Count on Transformer Approximation},
  author = {Penghao Yu and Haotian Jiang and Zeyu Bao and Ruoxi Yu and Qianxiao Li},
  year = {2026},
  booktitle = {International Conference on Learning Representations},
  eprint = {2510.06662},
  archivePrefix = {arXiv},
  url = {https://arxiv.org/abs/2510.06662}
}

@misc{hsu2026multitask,
  title = {Attention-based Representations for Multi-task Computation},
  author = {Daniel Hsu and Mingyue Xu},
  year = {2026},
  eprint = {2608.04243},
  archivePrefix = {arXiv},
  url = {https://arxiv.org/abs/2608.04243}
}

@misc{cui2024superiority,
  title = {Superiority of Multi-Head Attention in In-Context Linear Regression},
  author = {Yingqian Cui and Jie Ren and Pengfei He and Jiliang Tang and Yue Xing},
  year = {2024},
  eprint = {2401.17426},
  archivePrefix = {arXiv},
  url = {https://arxiv.org/abs/2401.17426}
}

@inproceedings{mahdavi2024memorization,
  title = {Memorization Capacity of Multi-Head Attention in Transformers},
  author = {Sadegh Mahdavi and Renjie Liao and Christos Thrampoulidis},
  year = {2024},
  booktitle = {International Conference on Learning Representations},
  eprint = {2306.02010},
  archivePrefix = {arXiv},
  url = {https://arxiv.org/abs/2306.02010}
}

@misc{fu2024dataengineering,
  title         = {Data Engineering for Scaling Language Models to {128K} Context},
  author        = {Fu, Yao and Panda, Rameswar and Niu, Xinyao and Yue, Xiang and Hajishirzi, Hannaneh and Kim, Yoon and Peng, Hao},
  year          = {2024},
  howpublished  = {arXiv preprint arXiv:2402.10171},
  eprint        = {2402.10171},
  archivePrefix = {arXiv},
  primaryClass  = {cs.CL},
  doi           = {10.48550/arXiv.2402.10171},
  url           = {https://arxiv.org/abs/2402.10171}
}

@inproceedings{chen2024longdependency,
  title     = {Long Context is Not Long at All: A Prospector of Long-Dependency Data for Large Language Models},
  author    = {Chen, Longze and Liu, Ziqiang and He, Wanwei and Zheng, Yinhe and Sun, Hao and Li, Yunshui and Luo, Run and Yang, Min},
  booktitle = {Proceedings of the 62nd Annual Meeting of the Association for Computational Linguistics (Volume 1: Long Papers)},
  editor    = {Ku, Lun-Wei and Martins, Andre and Srikumar, Vivek},
  year      = {2024},
  month     = aug,
  address   = {Bangkok, Thailand},
  publisher = {Association for Computational Linguistics},
  pages     = {8222--8234},
  doi       = {10.18653/v1/2024.acl-long.447},
  url       = {https://aclanthology.org/2024.acl-long.447/}
}

@misc{lu2024controlled,
  title         = {A Controlled Study on Long Context Extension and Generalization in {LLMs}},
  author        = {Lu, Yi and Yan, Jing Nathan and Yang, Songlin and Chiu, Justin T. and Ren, Siyu and Yuan, Fei and Zhao, Wenting and Wu, Zhiyong and Rush, Alexander M.},
  year          = {2024},
  howpublished  = {arXiv preprint arXiv:2409.12181},
  eprint        = {2409.12181},
  archivePrefix = {arXiv},
  primaryClass  = {cs.CL},
  doi           = {10.48550/arXiv.2409.12181},
  url           = {https://arxiv.org/abs/2409.12181}
}

@inproceedings{gao2025prolong,
  title     = {How to Train Long-Context Language Models (Effectively)},
  author    = {Gao, Tianyu and Wettig, Alexander and Yen, Howard and Chen, Danqi},
  booktitle = {Proceedings of the 63rd Annual Meeting of the Association for Computational Linguistics (Volume 1: Long Papers)},
  editor    = {Che, Wanxiang and Nabende, Joyce and Shutova, Ekaterina and Pilehvar, Mohammad Taher},
  year      = {2025},
  month     = jul,
  address   = {Vienna, Austria},
  publisher = {Association for Computational Linguistics},
  pages     = {7376--7399},
  doi       = {10.18653/v1/2025.acl-long.366},
  url       = {https://aclanthology.org/2025.acl-long.366/}
}

@misc{xu2025ultralong,
  title         = {From {128K} to {4M}: Efficient Training of Ultra-Long Context Large Language Models},
  author        = {Xu, Chejian and Ping, Wei and Xu, Peng and Liu, Zihan and Wang, Boxin and Shoeybi, Mohammad and Li, Bo and Catanzaro, Bryan},
  year          = {2025},
  howpublished  = {arXiv preprint arXiv:2504.06214},
  eprint        = {2504.06214},
  archivePrefix = {arXiv},
  primaryClass  = {cs.CL},
  doi           = {10.48550/arXiv.2504.06214},
  url           = {https://arxiv.org/abs/2504.06214}
}

@misc{liang2026learningdynamics,
  title         = {Revealing the Learning Dynamics of Long-Context Continual Pre-training},
  author        = {Liang, Yupu and Chen, Shuang and Zhang, Guanwei and Wang, Shaolei and Zheng, Suncong},
  year          = {2026},
  howpublished  = {arXiv preprint arXiv:2604.02650},
  eprint        = {2604.02650},
  archivePrefix = {arXiv},
  primaryClass  = {cs.CL},
  doi           = {10.48550/arXiv.2604.02650},
  url           = {https://arxiv.org/abs/2604.02650}
}

@misc{bertsch2026olmpool,
  title         = {Cracks in the Foundation: Seemingly Minor Architectural Choices Impact Long Context Extension},
  author        = {Bertsch, Amanda and Soldaini, Luca and Gormley, Matthew R. and Neubig, Graham and Hajishirzi, Hannaneh and Lo, Kyle and Groeneveld, Dirk},
  year          = {2026},
  howpublished  = {arXiv preprint arXiv:2608.10296},
  eprint        = {2608.10296},
  archivePrefix = {arXiv},
  primaryClass  = {cs.CL},
  note          = {Accepted to COLM 2026},
  doi           = {10.48550/arXiv.2608.10296},
  url           = {https://arxiv.org/abs/2608.10296}
}

@article{hu2024canperplexity,
  title         = {Can Perplexity Reflect Large Language Model's Ability in Long Text Understanding?},
  author        = {Hu, Yutong and Huang, Quzhe and Tao, Mingxu and
                   Zhang, Chen and Feng, Yansong},
  journal       = {arXiv preprint arXiv:2405.06105},
  year          = {2024},
  eprint        = {2405.06105},
  archivePrefix = {arXiv},
  primaryClass  = {cs.CL},
  url           = {https://arxiv.org/abs/2405.06105}
}

@inproceedings{fang2025wrongperplexity,
  title     = {What is Wrong with Perplexity for Long-context Language Modeling?},
  author    = {Fang, Lizhe and Wang, Yifei and Liu, Zhaoyang and
               Zhang, Chenheng and Jegelka, Stefanie and Gao, Jinyang and
               Ding, Bolin and Wang, Yisen},
  booktitle = {The Thirteenth International Conference on Learning Representations},
  year      = {2025},
  url       = {https://openreview.net/forum?id=fL4qWkSmtM}
}

@article{puvvada2025swangpt,
  title         = {{SWAN-GPT}: An Efficient and Scalable Approach for Long-Context Language Modeling},
  author        = {Puvvada, Krishna C. and Ladhak, Faisal and
                   Akle Serrano, Santiago and Hsieh, Cheng-Ping and
                   Acharya, Shantanu and Majumdar, Somshubra and
                   Jia, Fei and Kriman, Samuel and Sun, Simeng and
                   Rekesh, Dima and Ginsburg, Boris},
  journal       = {arXiv preprint arXiv:2504.08719},
  year          = {2025},
  eprint        = {2504.08719},
  archivePrefix = {arXiv},
  primaryClass  = {cs.CL},
  url           = {https://arxiv.org/abs/2504.08719v1}
}

@inproceedings{roy2007effectiverank,
  title     = {The Effective Rank: A Measure of Effective Dimensionality},
  author    = {Roy, Olivier and Vetterli, Martin},
  booktitle = {15th European Signal Processing Conference},
  year      = {2007},
  url       = {https://eurasip.org/Proceedings/Eusipco/Eusipco2007/Papers/a5p-h05.pdf}
}
\bibliographystyle{iclr2027_conference}

\newpage

\appendix

\section{Experimental details} \label{sec:details}

\subsection{Architectures} \label{sec:arch}

Detailed architectural configurations are shown in Table~\ref{tbl:arch}.
Models in both scale families are decoder-only Transformers with interleaving sliding window attention (SWA) and full attention (FA) layers.
We adopt the standard pre-norm structure and use RMSNorm \citep{rmsnorm} for normalization.
In all the attention layers, we apply head-wise RMSNorm to each query and key.
We also apply RMSNorm to the final representation fed to the unembedding module.
Embedding and unembedding have distinct parameters.

We use partial RoPE in full-attention layers, motivated by the proportional RoPE (p-RoPE) design used in Gemma 4 \citep{gemma4}, which applies RoPE only to a subset of each attention head.
Gemma 4 constructs its rotary frequencies relative to the full head dimension and only uses 1/4 of them, effectively retaining the higher-frequency portion of the RoPE spectrum.
We use a simpler parameterization, applying RoPE to the first 64 dimensions with a small base $\theta=100$, which similarly avoids very low-frequency rotary components.
The use of high-frequency partial RoPE simplifies the context-extension setup as we do not need any tricks to generalize the model to unseen context length.
For SWA layers, we use standard RoPE with $\theta=10,000$.

\subsection{Datasets} \label{sec:data}

The data mix in each stage is summarized in Table~\ref{tbl:datamix}.
We construct data mix in pretraining and CPT stages with 60\% web data from Nemotron-CC-v2, 30\% code data from Swallow Code and Swallow Code v2, and 10\% SFT-style data from Nemotron-Pretraining-SFT-v1.
In our preliminary experiments, mixing code data contributed to the improved performance in retrieval tasks, despite RULER not requiring code comprehension.
The data mix used in the SFT stage consists of 50\% long sequence SFT data and 50\% short sequence chat conversations.
All the datasets are tokenized by Qwen3 tokenizer.

\paragraph{Preprocessing for Swallow Code v2:}
In the original data from Swallow Code v2, one sequence corresponds to one file, whose file path and the repository name are provided as a part of metadata.
To enhance training of distant token relations, we concatenate files from the same repository and use the resulting long text as a sequence.
We concatenate files in the following format:
\begin{verbatim}
<repository name="(repository name)">
<file path="(file path)">
(content)
</file>
...
<file path="(file path)">
(content)
</file>
</repository>
\end{verbatim}
We only use the resulting repository-wise sequences of length up to 48k for pretraining and those of length ranging between 48k and 384k for CPT.

\subsection{Training configuration} \label{sec:train-config}

\begin{table}[p]
    \caption{Architecture configurations.}
    \begin{center}
    \begin{tabular}{c|cc}
        \toprule
        Model size & 0.6B & 3B \\
        \midrule
        Vocab size & 151,699 & 151,699 \\
        Model dimension & 1,024 & 2,048 \\
        Number of layers & 24 & 40 \\
        SWA:FA ratio & 3:1 & 3:1 \\
        SWA number of heads & 4 & 8 \\
        SWA head size & 256 & 256 \\
        SWA window size & 512 & 512 \\
        FA partial RoPE dimension & 64 & 64 \\
        FA partial RoPE $\theta$ & 100 & 100 \\
        FA KV elements/layer-token (symmetric) & 1,024 & 2,048 \\
FA KV elements/layer-token (asymmetric) & 1,280 & -- \\
        MLP activation function & SwiGLU & SwiGLU \\
        MLP dimension & 3,072 & 6,144 \\
        \bottomrule
    \end{tabular}
    \end{center}
    \label{tbl:arch}
\end{table}

\begin{table}[p]
    \caption{Data mix used in each training stage.}
    \begin{center}
    \begin{tabular}{cl|c}
        \toprule
        Stage & Dataset & Tokens \\
        \midrule
        \multirow{4}{*}{Pretraining}
        & Nemotron-CC-v2 High Quality \citep{nemotronnano2} & 60B \\
        & Swallow Code \citep{swallowcode} & 15B\\
        & Swallow Code v2 (packed by repositories) & 15B \\
        & Nemotron-Pretraining-SFT-v1 General & 10B \\
        \midrule
        \multirow{3}{*}{CPT}
        & Nemotron-CC-v2 High Quality & 4.8B \\
        & Swallow Code v2 (packed by repositories) & 2.4B \\
        & Nemotron-Pretraining-SFT-v1 General & 0.8B \\
        \midrule
        \multirow{2}{*}{SFT}
        & ChatQA2-Long-SFT \citep{chatqa2} & 0.5B \\
        & Nemotron-Instruction-Following-Chat-v1 chat\_if & 0.5B \\
        \bottomrule
    \end{tabular}
    \end{center}
    \label{tbl:datamix}
\end{table}

\begin{table}[p]
    \caption{Training hyperparameters.}
    \begin{center}
    \begin{tabular}{cc|cc}
        \toprule
        & Model size & 0.6B & 3B \\
        \midrule
        \multirow{6}{*}{Pretraining} & Tokens & 100B & 100B \\
        & Learning rate & 2e-4 & 1e-4 \\
        & LR warm-up steps & 250 & 250 \\
        & Weight decay & 0.1 & 0.1 \\
        & Batch size & 1,024 & 1,024 \\
        & Sequence length & 4,096 & 4,096 \\
        \midrule
        \multirow{6}{*}{CPT} & Tokens & 8B & 8B \\
        & Learning rate & 1e-4 & 5e-5 \\
        & LR warm-up steps & 0 & 0 \\
        & Weight decay & 0.01 & 0.01 \\
        & Batch size & 16 & 16 \\
        & Sequence length & 262,144 & 262,144 \\
        \midrule
        \multirow{6}{*}{SFT} & Tokens after padding & - & 1B \\
        & Learning rate & - & 2e-5 \\
        & LR warm-up steps & - & 50 \\
        & Weight decay & - & 0.01 \\
        & Padded batch size (tokens) & - & 1,048,576 \\
        & Sequence length after packing & - & 32,768 \\
        \midrule
        \multirow{1}{*}{All stages} & LM-head z-loss multiplier & 1e-4 & 1e-4 \\
        \bottomrule
    \end{tabular}
    \end{center}
    \label{tbl:hp}
\end{table}

Table~\ref{tbl:hp} provides the training hyperparameters used in each stage.
We use AdamW optimizer in all the stages, and use warmup-stable learning rate scheduling.
In the CPT stage, we skip the warm-up phase.

\subsection{Inference-time attention scaling for PT models}
\label{app:pt_temperature_scaling}

We evaluate PT checkpoints trained for 100B tokens at a sequence length
of 4,096 using position-dependent attention-logit scaling at inference time,
without updating any model parameters.
The logarithmic scaling has the same functional form as the dynamic
attention scaling used for global NoPE layers in SWAN-GPT
\citep{puvvada2025swangpt}; here, we apply it to full-attention layers
with partial RoPE.

Let $c>1$ be a scaling constant and let $q_p$ denote the query vector used to compute attention logits at zero-indexed token position $p$.
For every query head in a full-attention layer, we use
\begin{equation}
    a_c(p)=\frac{\log(c+p)}{\log c},
    \qquad
    \widetilde{q}_p=a_c(p)q_p.
    \label{eq:pt_attention_scaling}
\end{equation}
For a causally visible key $k_j$ at position $j\le p$, this gives
\[
    \widetilde{z}_{p,j}
    =\frac{\widetilde{q}_p^\top k_j}{\sqrt{d_k}}
    =a_c(p)\frac{q_p^\top k_j}{\sqrt{d_k}},
\]
where $d_k$ is the query/key dimension.

We choose $c$ to be on the order of the pretraining context length (4,096 tokens) and use $c=4000$ for all reported scaled-PT evaluations.
The scaling is applied only to full-attention layers.
We refer to checkpoints evaluated with this procedure as \emph{scaled PT},
labeled ``PT (attention scaling)'' in the figures.

\section{Supplementary results on continual pretraining}
\label{sec:pt_vs_cpt}

Here we present detailed results for our long-context continual pretraining (CPT) experiments.
We first report next-token prediction (NTP) training losses for the 0.6B and 3B models during pretraining (PT) and Full CPT (Appendix~\ref{app:pt_cpt_loss}).
We then present RULER-100 results for both model scales, broken down by task and context length (Appendix~\ref{app:head_allocation_tasks}).
Next, we describe the experimental setup and parameter-update policy for Frozen CPT (Appendix~\ref{sec:frozen_cpt_setup}).
Finally, we report RULER-100 results by task and context length for the 0.6B models under Frozen CPT and provide detailed comparisons with Full CPT in terms of training loss and RULER-100 performance (Appendix~\ref{sec:frozen_cpt_results}).
Unless otherwise specified, CPT refers to Full CPT, in which all model parameters are updated.

\subsection{Training loss during PT and Full CPT}
\label{app:pt_cpt_loss}

This section provides detailed next-token prediction (NTP) training loss results for the three symmetric head allocations, \texttt{d128}, \texttt{d256}, and \texttt{d512}, at the 0.6B and 3B parameter scales (Table~\ref{tbl:head-size}).

\begin{figure}[t]
    \centering
    \begin{minipage}[t]{0.49\linewidth}
        \centering
        \includegraphics[width=\linewidth]{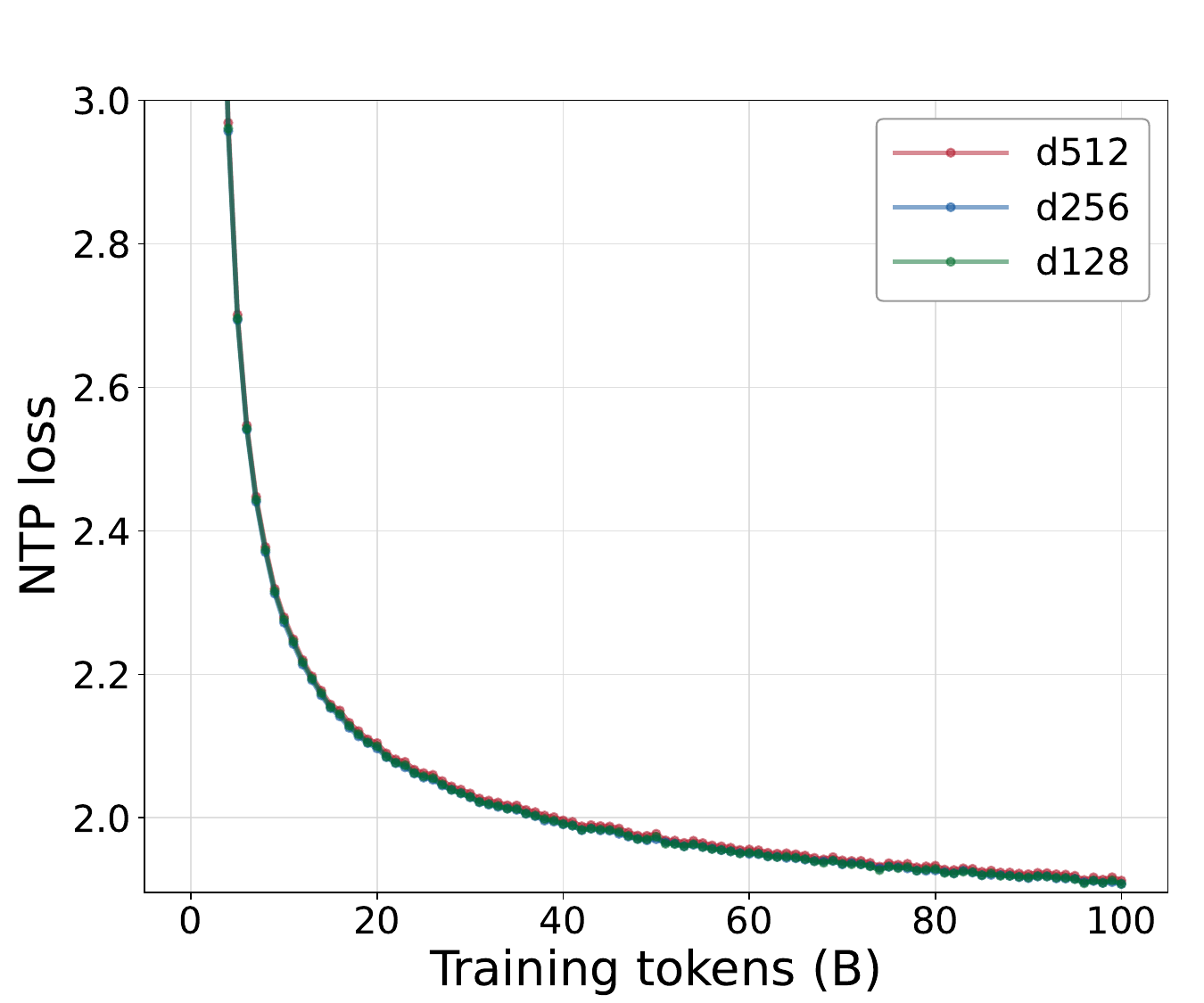}
        \par\small (a) 0.6B PT
    \end{minipage}
    \begin{minipage}[t]{0.49\linewidth}
        \centering
        \includegraphics[width=\linewidth]{figures0910/ntp_loss_0p6b_cpt.pdf}
        \par\small (b) 0.6B Full CPT
    \end{minipage}
    \par\medskip
    \begin{minipage}[t]{0.49\linewidth}
        \centering
        \includegraphics[width=\linewidth]{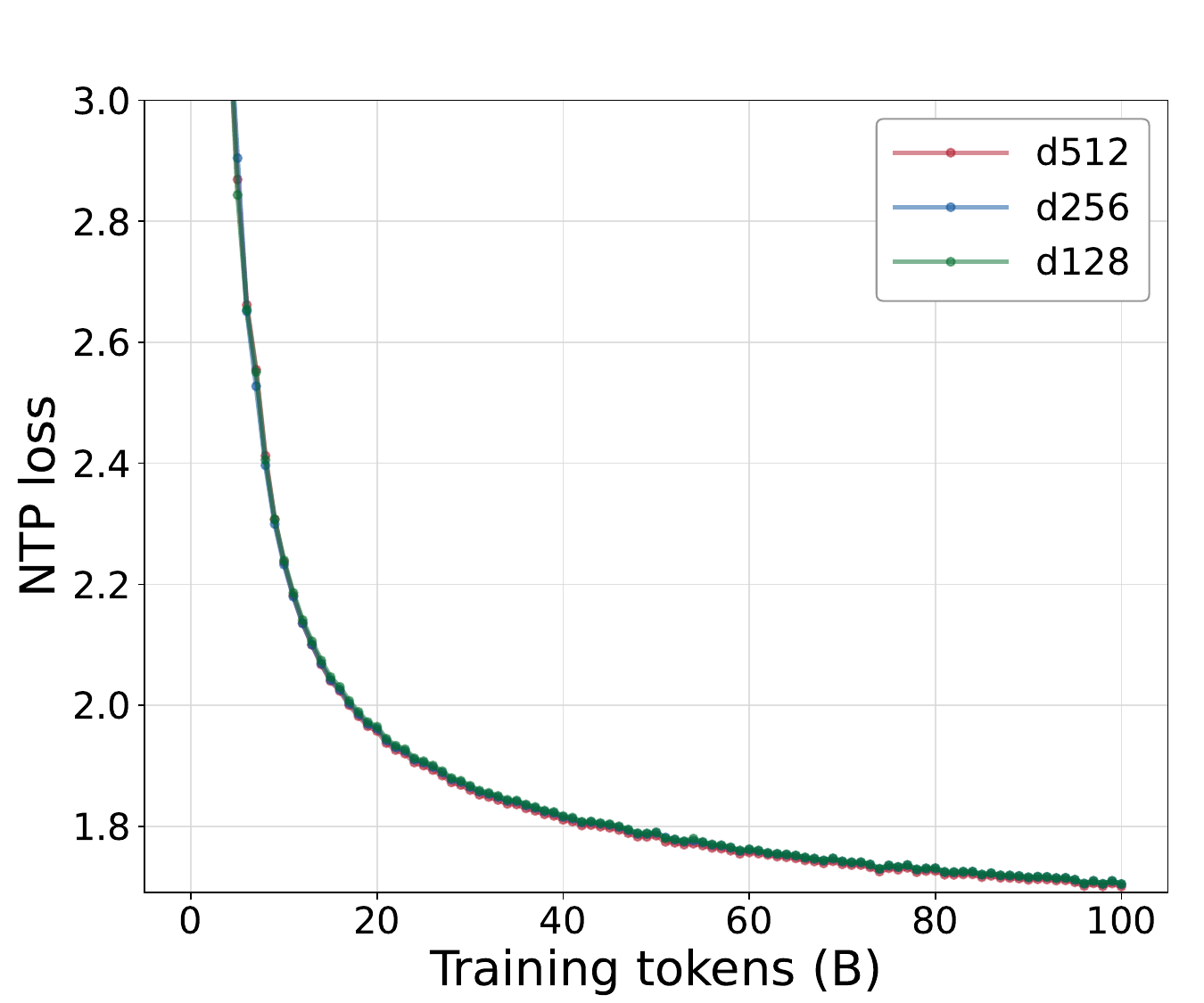}
        \par\small (c) 3B PT
    \end{minipage}
    \begin{minipage}[t]{0.49\linewidth}
        \centering
        \includegraphics[width=\linewidth]{figures0910/ntp_loss_3b_cpt.pdf}
        \par\small (d) 3B Full CPT
    \end{minipage}
    \caption{Training NTP loss during 100B-token PT at 4K and 8B-token Full CPT at 256K.
    The horizontal axis shows tokens in billions within the indicated stage; the vertical axis shows training cross-entropy.
    The 3B panels show one run per model.
    In the 0.6B panels, lines indicate means across three runs and shaded bands indicate $\pm1$ sample standard deviation.}
    \label{fig:pt_cpt_training_loss}
\end{figure}

Figure~\ref{fig:pt_cpt_training_loss} shows the training NTP loss as a function of the number of training tokens during PT and Full CPT.
PT uses 100B tokens at a context length of 4K, followed by an additional 8B tokens of Full CPT at 256K.
The horizontal axis counts training tokens in billions within each stage.
For the 0.6B models, lines show means across three runs and shaded bands indicate $\pm 1$ sample standard deviation; each 3B head allocation has one run.
Table~\ref{tab:pt_cpt_ntp_loss} reports the loss at the final logged interval of each stage, with means and sample standard deviations for the 0.6B models and single-run values for the 3B models.

At both model scales, the three head allocations exhibit similar loss trajectories during PT and Full CPT.
The final losses also vary little across head dimensions in both stages.
Thus, changing head allocation has only a small effect on NTP training loss under our training setup.

\begin{table}[t]
    \centering
    \caption{Training NTP loss at the final logged interval of PT and Full CPT.
    Entries for the 0.6B models are means $\pm$ sample standard deviations across three runs; entries for the 3B models are single-run values.}
    \label{tab:pt_cpt_ntp_loss}
    \begin{tabular}{lcc}
        \toprule
        Model & PT & Full CPT \\
        \midrule
        \texttt{0.6B-d128} & $1.908\pm0.001$ & $1.911\pm0.009$ \\
        \texttt{0.6B-d256} & $1.908\pm0.001$ & $1.888\pm0.042$ \\
        \texttt{0.6B-d512} & $1.913\pm0.004$ & $1.912\pm0.011$ \\
        \midrule
        \texttt{3B-d128} & $1.705$ & $1.677$ \\
        \texttt{3B-d256} & $1.704$ & $1.673$ \\
        \texttt{3B-d512} & $1.700$ & $1.665$ \\
        \bottomrule
    \end{tabular}
\end{table}

\subsection{RULER-100 results by task and context length}
\label{app:head_allocation_tasks}

\begin{figure}[p]
    \centering
    \includegraphics[width=0.89\linewidth]{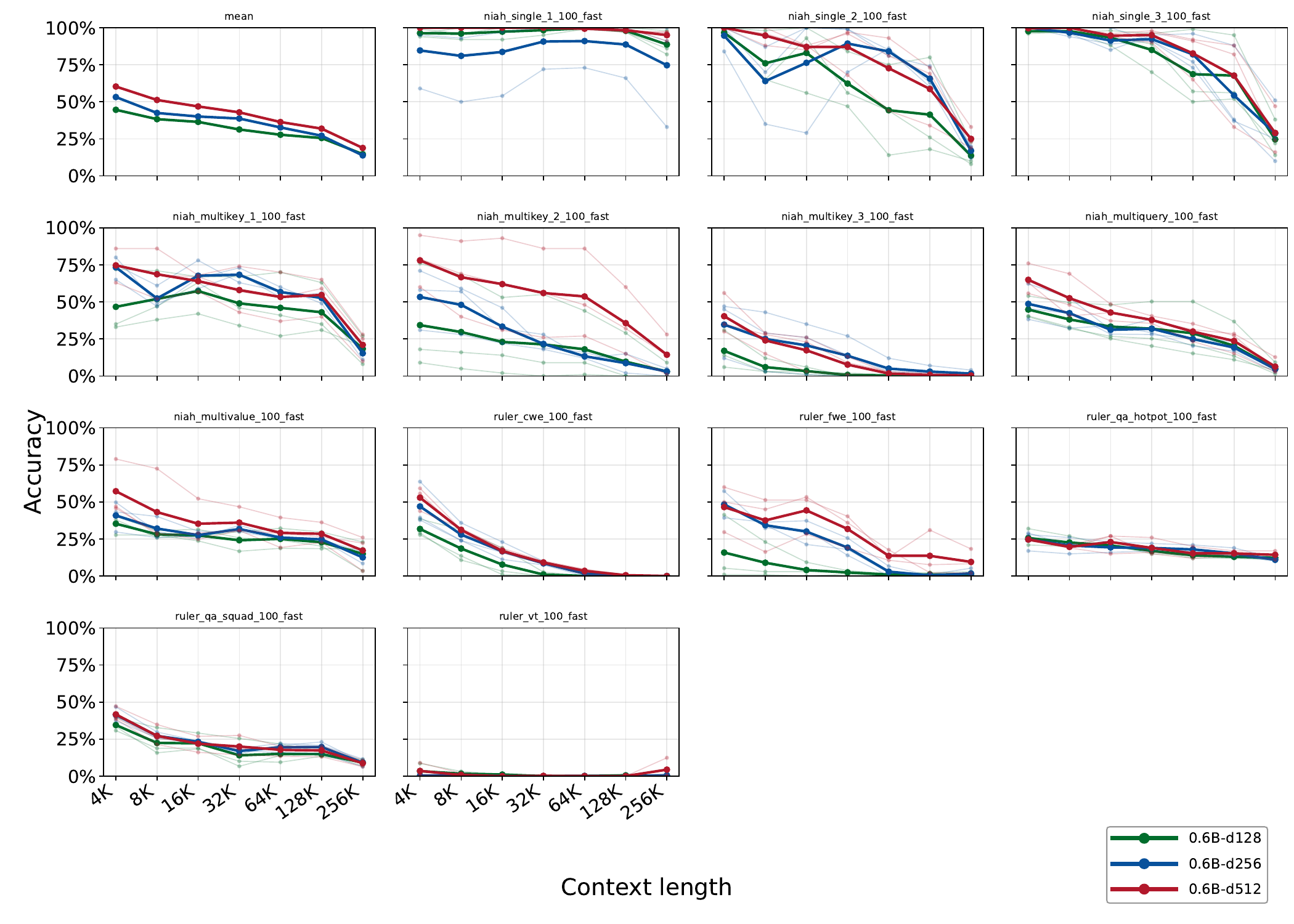}
    \caption{Task-level RULER-100 scores after 8B-token Full CPT for the 0.6B models.
    The first panel shows the 13-task mean; the remaining panels show individual tasks.
    Thick lines indicate means across three runs, and faint lines show individual runs.
    The horizontal axis shows context length, and the vertical axis shows task scores as percentages.}
    \label{fig:cpt_ruler_tasks_0p6b}
\end{figure}

\begin{figure}[p]
    \centering
    \includegraphics[width=0.89\linewidth]{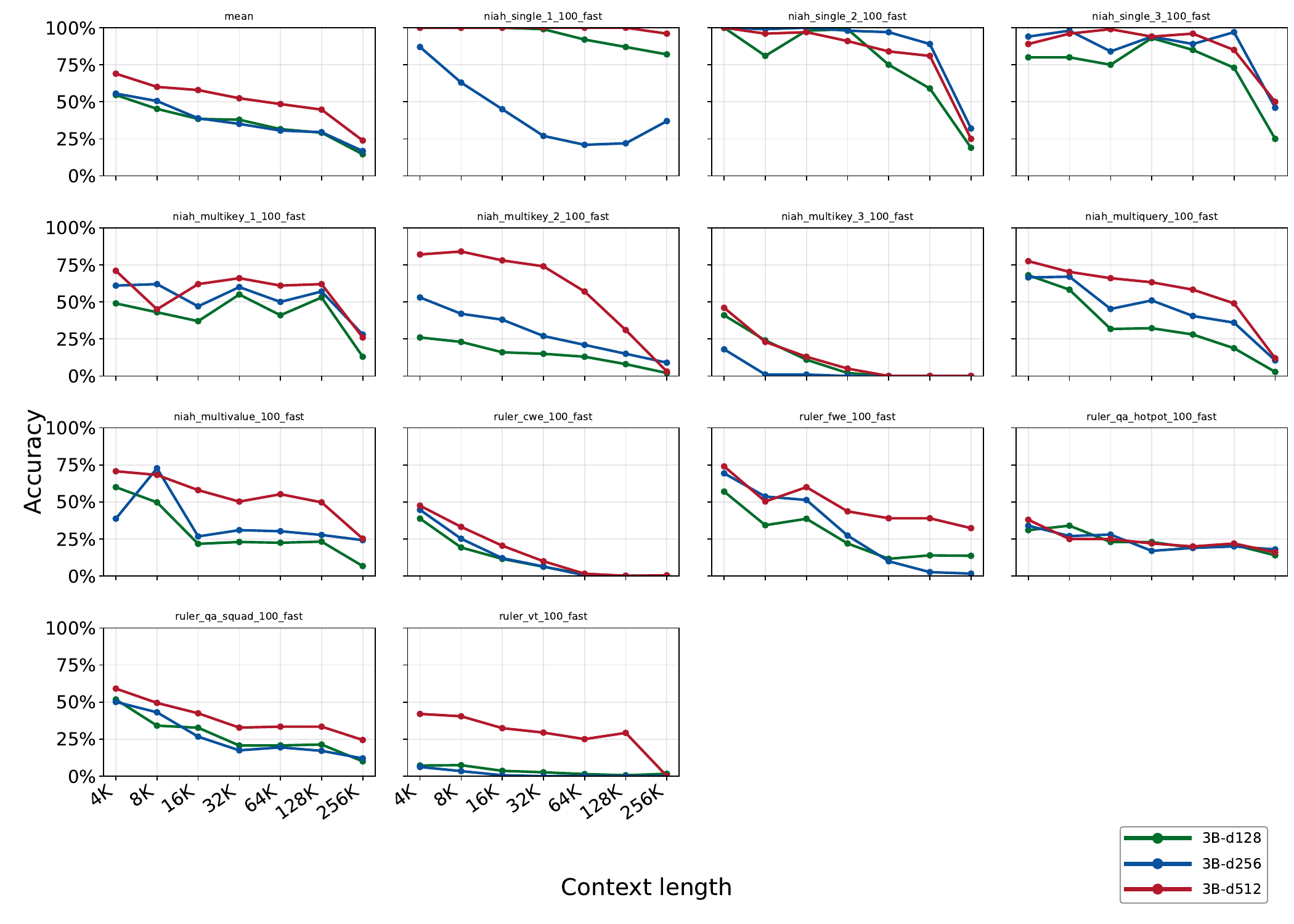}
    \caption{Task-level RULER-100 scores after 8B-token Full CPT for the 3B models.
    The first panel shows the 13-task mean; the remaining panels show individual tasks.
    The horizontal axis shows context length, and the vertical axis shows task scores as percentages.
    Each model has one run.
    \texttt{3B-d512} leads in aggregate but not on every task.}
    \label{fig:cpt_ruler_tasks_3b}
\end{figure}

\begin{figure}[t]
    \centering
    \includegraphics[width=0.89\linewidth]{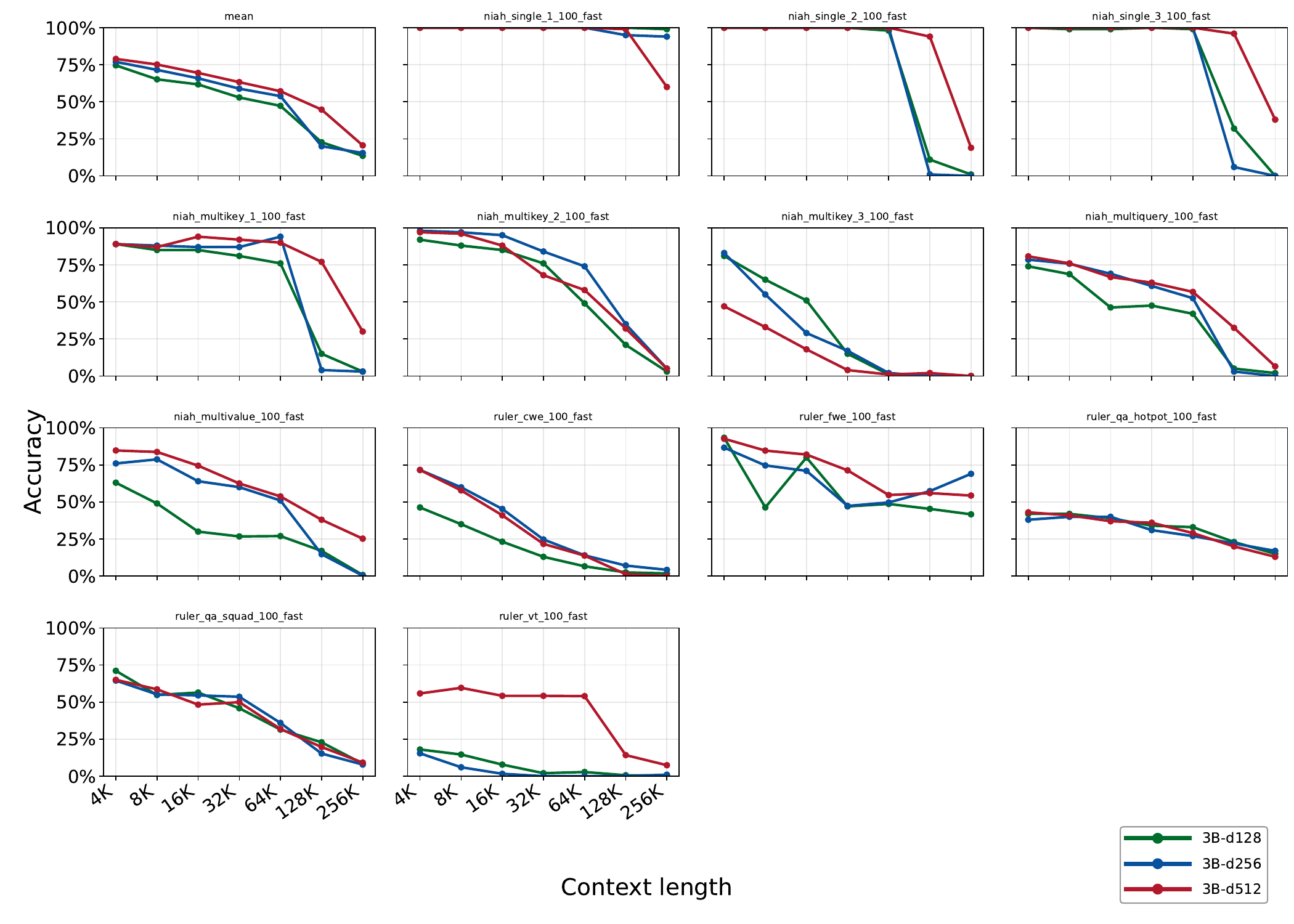}
    \caption{Task-level RULER-100 scores for the 3B models after 1B tokens of SFT following CPT.
    The first panel shows the 13-task mean; the remaining panels show individual tasks.
    The horizontal axis shows context length, and the vertical axis shows task scores as percentages.
    Each model has one run.
    Relative to the CPT results in Figure~\ref{fig:cpt_ruler_tasks_3b}, overall scores increase, whereas scores at 256K decrease for all three models.}
    \label{fig:sft_ruler_tasks_3b}
\end{figure}

Figures~\ref{fig:cpt_ruler_tasks_0p6b}, \ref{fig:cpt_ruler_tasks_3b}, and~\ref{fig:sft_ruler_tasks_3b} provide task- and length-level RULER-100 results for the 0.6B models after Full CPT, the 3B models after Full CPT, and the 3B models after SFT, respectively.
These correspond to panels (a), (b), and (c) of Figure~\ref{fig:cpt_ruler_by_length}.
In each figure, the first panel shows the 13-task mean, and the remaining panels show individual task scores across the seven evaluated context lengths from 4K to 256K.
For the 0.6B models, thick lines indicate means across three runs and faint lines show individual runs; each 3B model has one run.

The advantage of wider heads is not uniform across individual tasks: a larger head dimension does not always yield a higher score for a given task and context length.
These task-level differences may partly reflect run-to-run variability associated with random initialization and other sources of training randomness.
Nevertheless, averaging across the 13 RULER-100 tasks reveals a consistent advantage for wider heads: \texttt{d512} achieves the highest average score at every evaluated context length in all three settings (Figure~\ref{fig:cpt_ruler_by_length}).
Thus, wider heads improve long-context retrieval performance on average across tasks, even though the ordering of head allocations can differ on individual tasks.

Figures \ref{fig:cpt_ruler_tasks_0p6b} and \ref{fig:cpt_ruler_tasks_3b} provide the task-level results.
Figure~\ref{fig:sft_ruler_tasks_3b} reports task-level results for the 3B models after 1B tokens of SFT following CPT.
Overall RULER-100 scores increase from 35.93 to 48.28 for \texttt{3B-d128}, from 36.72 to 51.81 for \texttt{3B-d256}, and from 50.93 to 58.49 for \texttt{3B-d512}.
The widest-head model therefore retains the highest overall score after SFT.
However, scores at 256K decrease from 14.60 to 13.54, from 16.81 to 15.48, and from 23.88 to 20.60, respectively.

\subsection{Experimental setup for Frozen CPT}
\label{sec:frozen_cpt_setup}

This section describes the training setup and parameter-update policy for attention-frozen CPT (Frozen CPT).
For each of the three symmetric 0.6B configurations, \texttt{d128}, \texttt{d256}, and \texttt{d512}, we compare full-parameter CPT (Full CPT) and Frozen CPT using PT checkpoints from three independent runs.
Within each run, both conditions start from the same 100B-token PT checkpoint and undergo 8B tokens of CPT at a context length of 256K.

Full CPT updates all model parameters.
In contrast, Frozen CPT keeps the query, key, value, and attention-output projection matrices ($W_Q$, $W_K$, $W_V$, and $W_O$) in both full-attention and sliding-window attention layers fixed at their PT checkpoint values throughout CPT.
We follow the parameter-update policies summarized in Table~\ref{tab:attention_freeze_policy}.
Token embeddings, the language-model output head, MLP parameters, and all RMSNorm parameters remain trainable.

In particular, the head-wise query/key RMSNorm gain vectors within the attention modules are updated rather than frozen.
We keep these gain vectors trainable to allow the attention logit scale to adapt during long-context extension, even when the projection matrices are fixed.
Thus, fixed projections do not imply fixed attention maps: both the representations entering attention and the normalization gains can still adapt during Frozen CPT.
Neither CPT condition uses the additional inference-time attention scaling applied to the scaled-PT baseline.

\begin{table}[t]
    \centering
    \small
    \renewcommand{\arraystretch}{1.15}
    \setlength{\tabcolsep}{6pt}
    \caption{Parameter-update policies during CPT. Fixed parameters retain their values from the paired 100B-token PT checkpoints. Frozen CPT disables updates on weight matrices in both FA and SWA layers, while leaving the listed normalization and non-attention parameters trainable.}
    \label{tab:attention_freeze_policy}
    \begin{tabular}{@{}p{0.45\linewidth}cc@{}}
        \toprule
        Parameter group
        & \shortstack{Full-Parameter\\Update (Full)} & \shortstack{Attention-Frozen\\ Update (Frozen)} \\
        \midrule
        $W_Q,W_K,W_V,W_O$ in FA and SWA layers
        & Updated & Fixed \\
        Head-wise QK-RMSNorm gain vectors
        & Updated & Updated \\
        LM head, token embeddings, SwiGLU MLPs, and other RMSNorm parameters
        & Updated & Updated \\
        \midrule
        Trainable parameter ratio & $100\%$ & $83.4\%$ \\
        \bottomrule
    \end{tabular}
\end{table}

\subsection{Frozen CPT results and comparisons with Full CPT}
\label{sec:frozen_cpt_results}

This section presents detailed results for the three symmetric 0.6B configurations, \texttt{d128}, \texttt{d256}, and \texttt{d512}, under Frozen CPT and compares them with Full CPT and the corresponding PT checkpoints.
For RULER-100 results, we include both unmodified PT and scaled PT, which applies the inference-time attention logit scaling described in Appendix~\ref{app:pt_temperature_scaling} without updating any model parameters.
Neither CPT condition uses this additional scaling.

\begin{figure}[p]
    \centering
    \begin{minipage}[t]{0.32\linewidth}
        \centering
        \includegraphics[width=\linewidth]{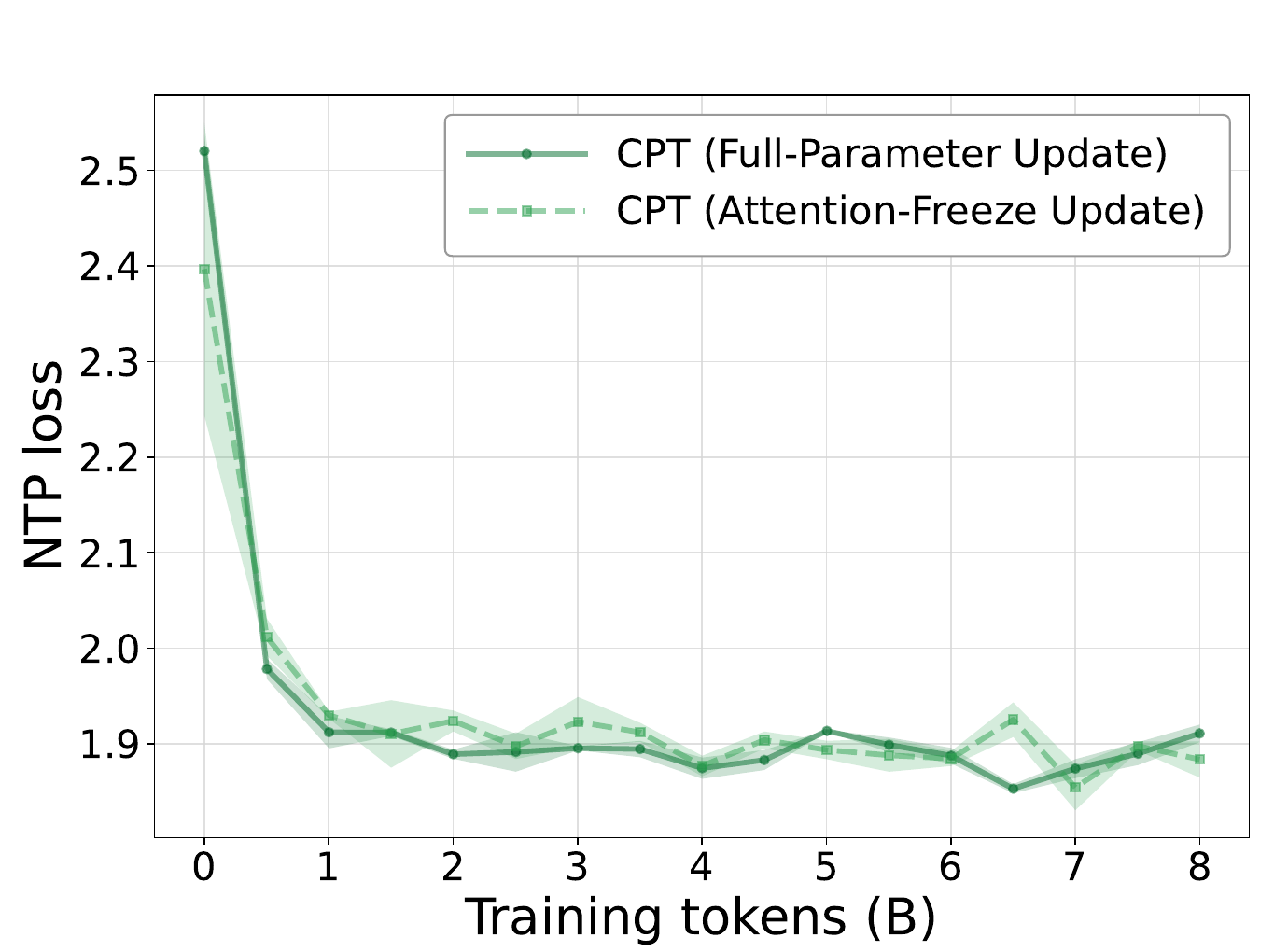}
        \vspace{-0.5em}
        \par\small (a) \texttt{0.6B-d128}
    \end{minipage}
    \begin{minipage}[t]{0.32\linewidth}
        \centering
        \includegraphics[width=\linewidth]{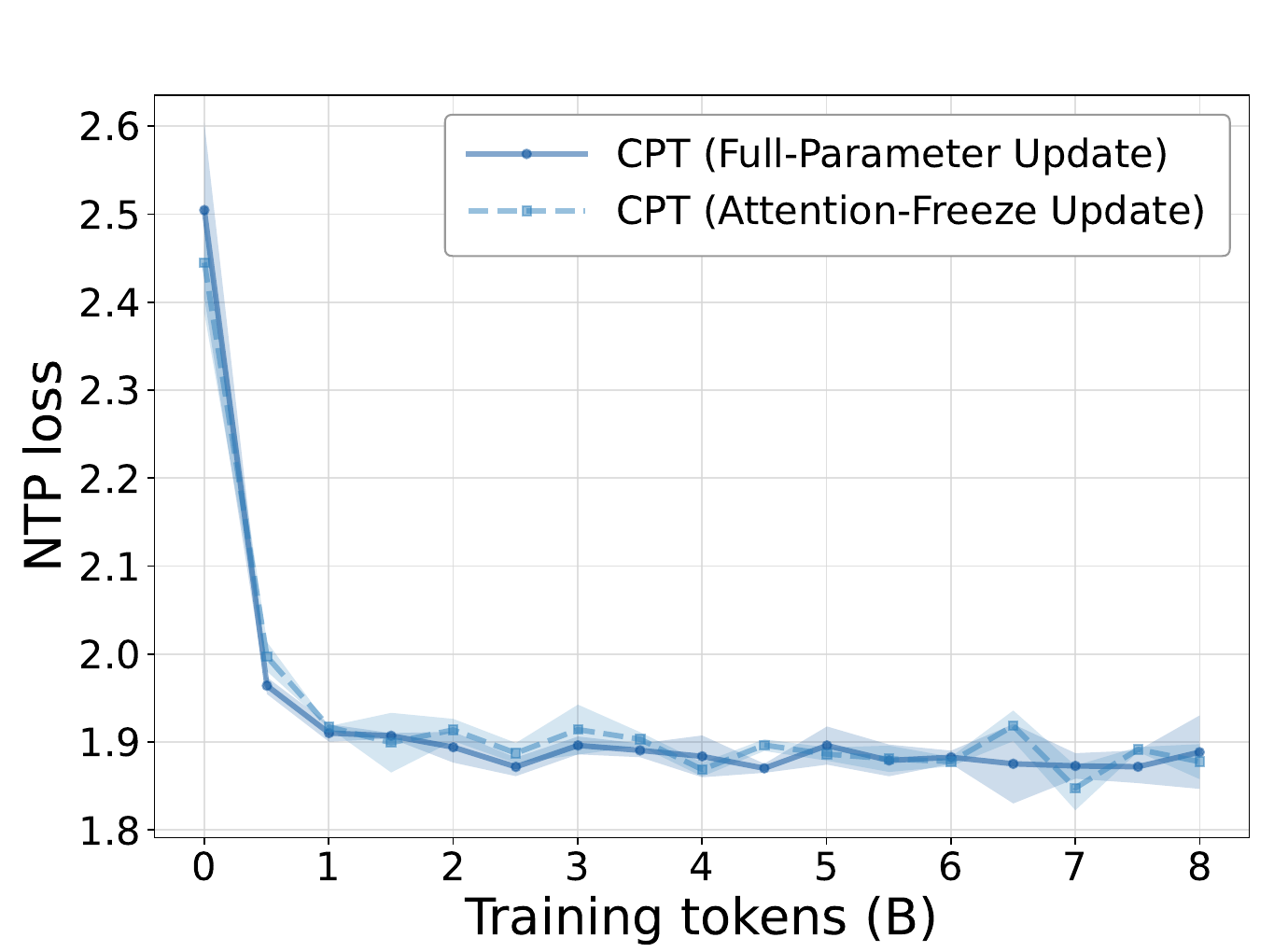}
        \vspace{-0.5em}
        \par\small (b) \texttt{0.6B-d256}
    \end{minipage}
    \begin{minipage}[t]{0.32\linewidth}
        \centering
        \includegraphics[width=\linewidth]{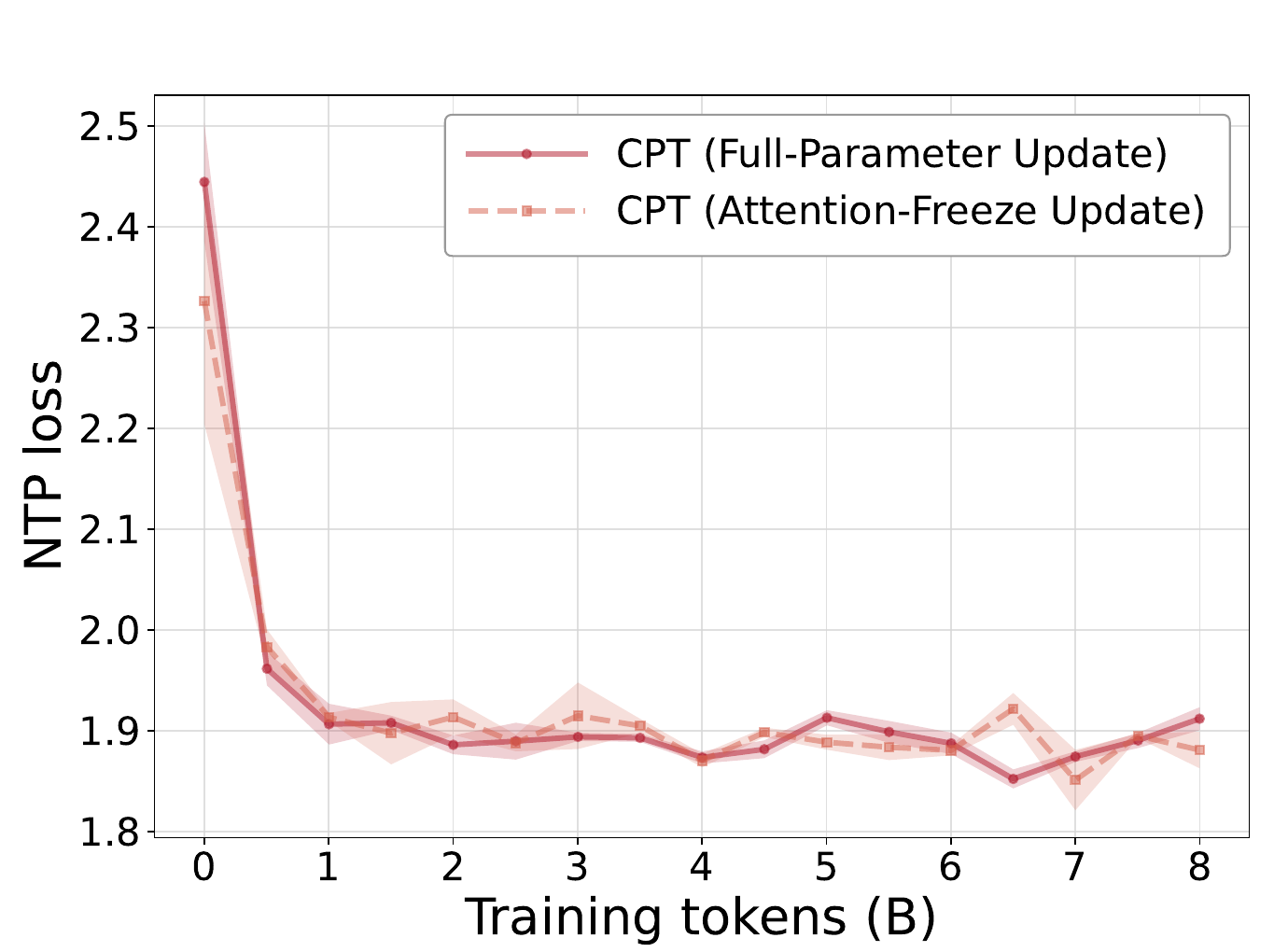}
        \vspace{-0.5em}
        \par\small (c) \texttt{0.6B-d512}
    \end{minipage}
    \caption{Training next-token prediction (NTP) loss during Full and Frozen CPT for the three 0.6B models.
    The horizontal axis shows CPT tokens in billions, and the vertical axis shows training cross-entropy.
    Lines indicate means across three runs, and shaded bands indicate $\pm$ sample standard deviation.
    Frozen CPT attains lower mean final training loss than Full CPT for all three models, despite achieving lower mean overall RULER-100 scores (Figure~\ref{fig:attention_freeze_ruler}).}
    \label{fig:attention_freeze_training_loss}
\end{figure}

We first compare the training next-token prediction (NTP) loss during Frozen CPT and Full CPT.
Figure~\ref{fig:attention_freeze_training_loss} shows the loss trajectories as a function of the number of CPT tokens for each head allocation.
Table~\ref{tab:attention_freeze_ntp_loss} reports the loss at the final logged interval of each condition.
The two CPT conditions exhibit similar loss trajectories and final losses, although Frozen CPT attains a slightly lower mean final loss for all three configurations.

\begin{figure}[p]
    \centering
    \begin{minipage}[t]{0.32\linewidth}
        \centering
        \includegraphics[width=\linewidth]{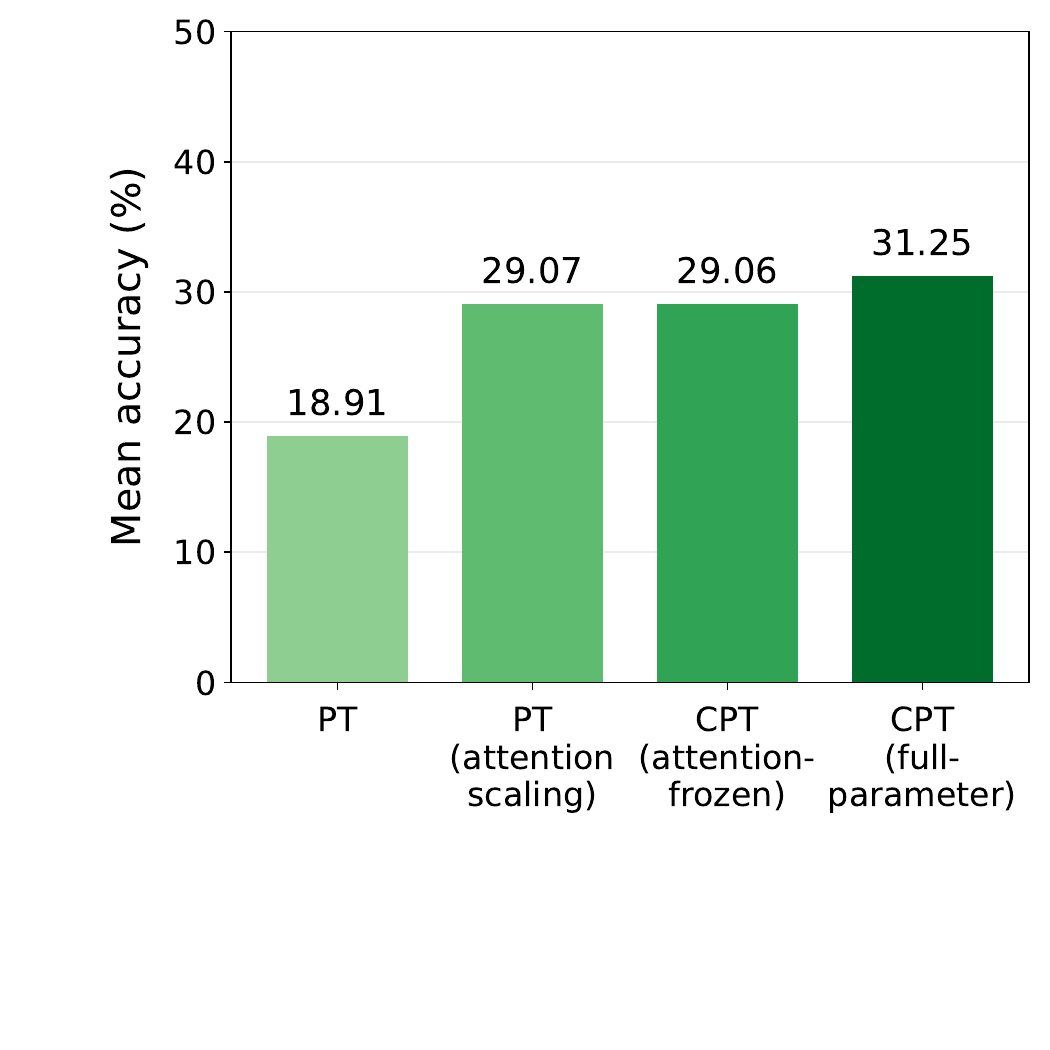}
        \vspace{-10mm}
        \par\small (a) \texttt{0.6B-d128}
    \end{minipage}
    \begin{minipage}[t]{0.32\linewidth}
        \centering
        \includegraphics[width=\linewidth]{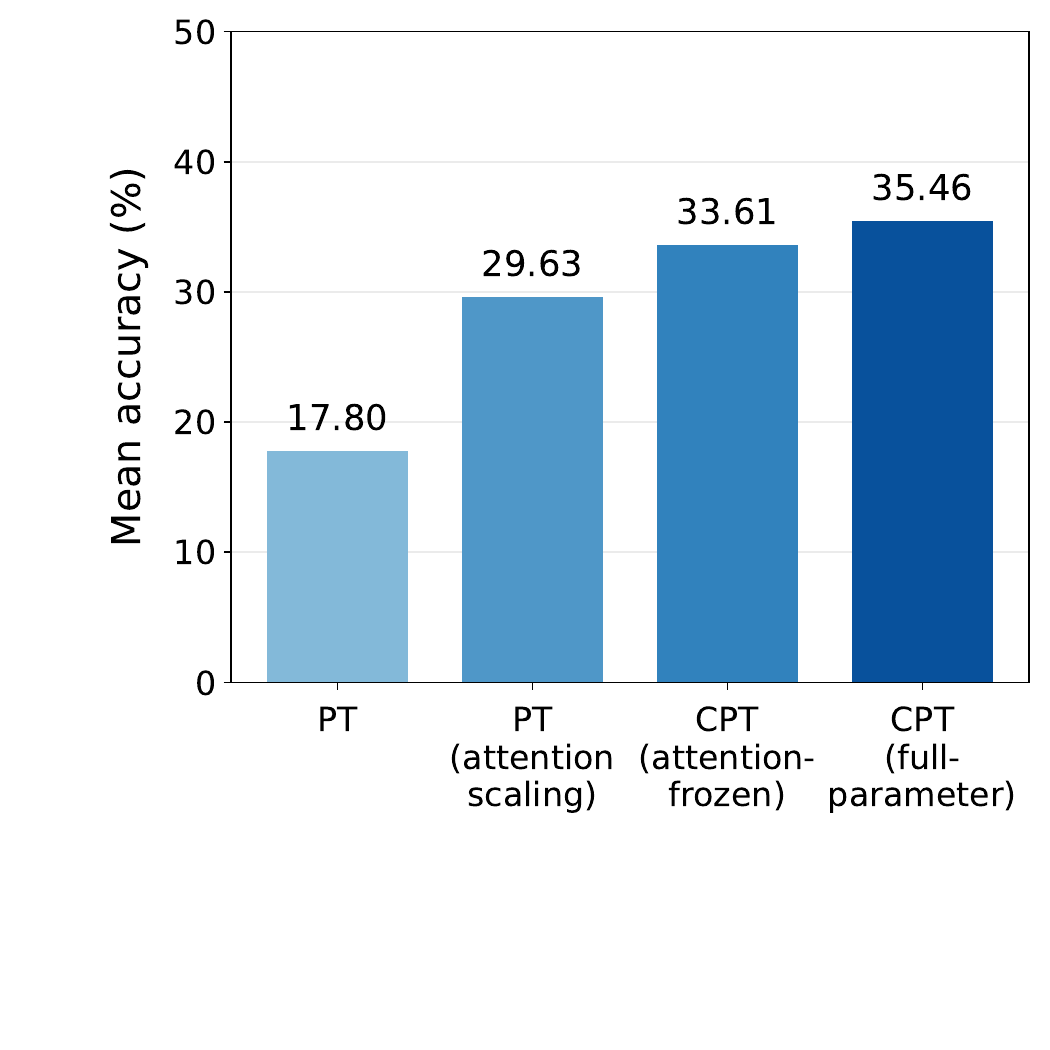}
        \vspace{-10mm}
        \par\small (b) \texttt{0.6B-d256}
    \end{minipage}
    \begin{minipage}[t]{0.32\linewidth}
        \centering
        \includegraphics[width=\linewidth]{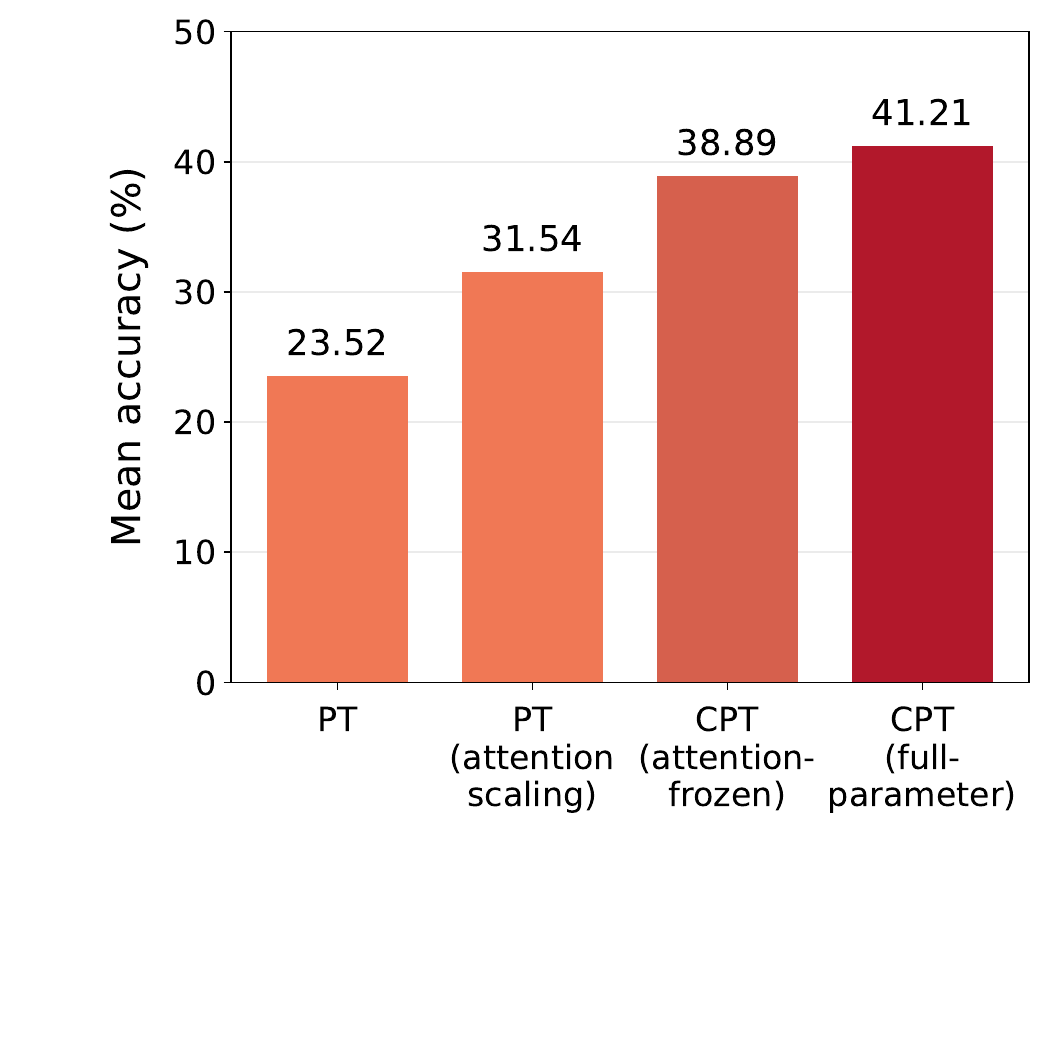}
        \vspace{-10mm}
        \par\small (c) \texttt{0.6B-d512}
    \end{minipage}
    \caption{Overall RULER-100 scores for unmodified PT, scaled PT,
    Frozen CPT, and Full CPT in the three 0.6B configurations.
    Bars show means across three runs, after averaging equally
    over 13 tasks and seven context lengths from 4K to 256K within
    each run. Scaled PT uses the inference-time procedure in
    Appendix~\ref{app:pt_temperature_scaling}; neither CPT condition
    uses additional scaling.}
    \label{fig:pt_cpt_ruler_overall}
\end{figure}

\begin{figure}[p]
    \centering
    \begin{minipage}[t]{0.32\linewidth}
        \centering
        \includegraphics[width=\linewidth]{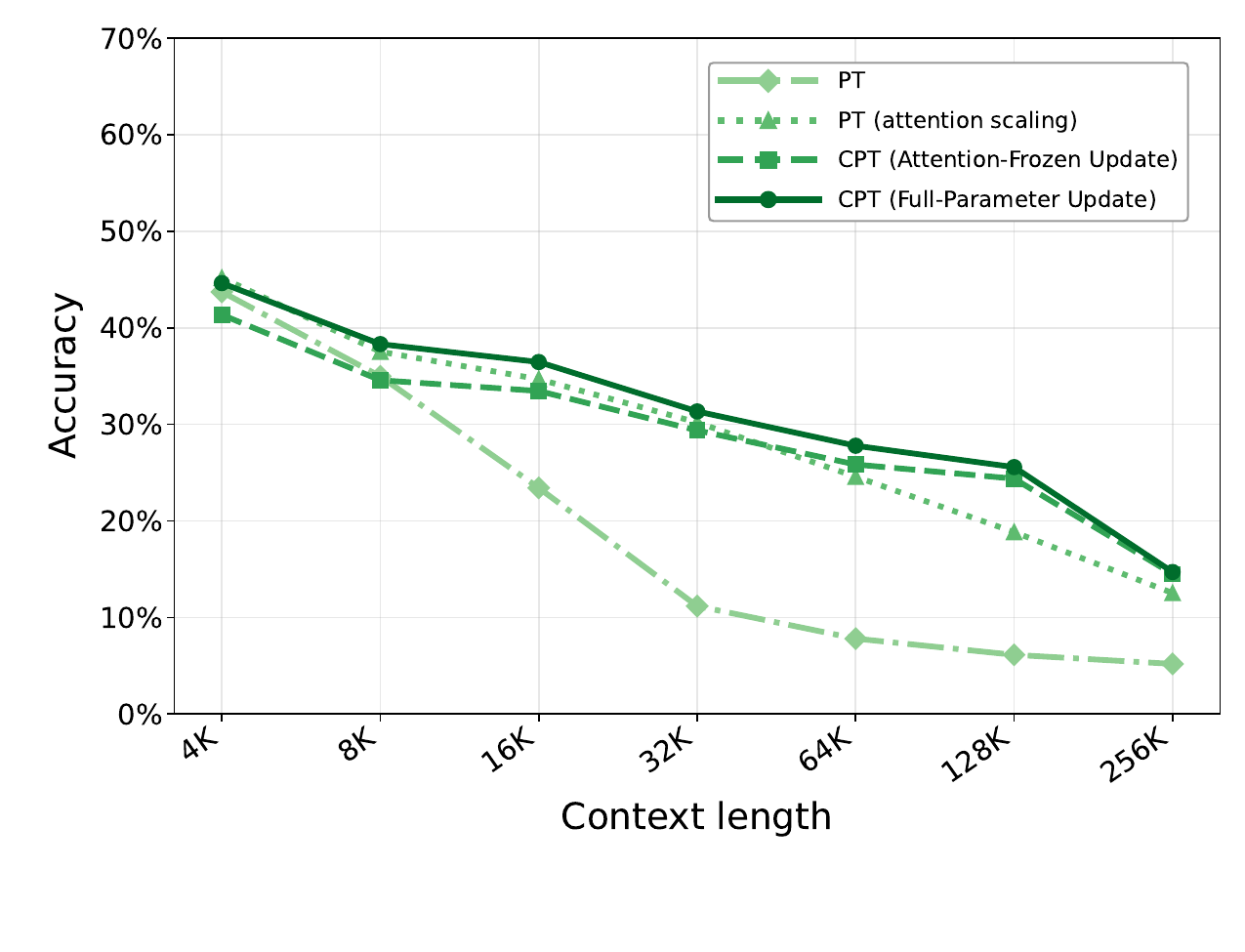}
        \vspace{-1.5em}
        \par\small (a) \texttt{0.6B-d128}
    \end{minipage}
    \begin{minipage}[t]{0.32\linewidth}
        \centering
        \includegraphics[width=\linewidth]{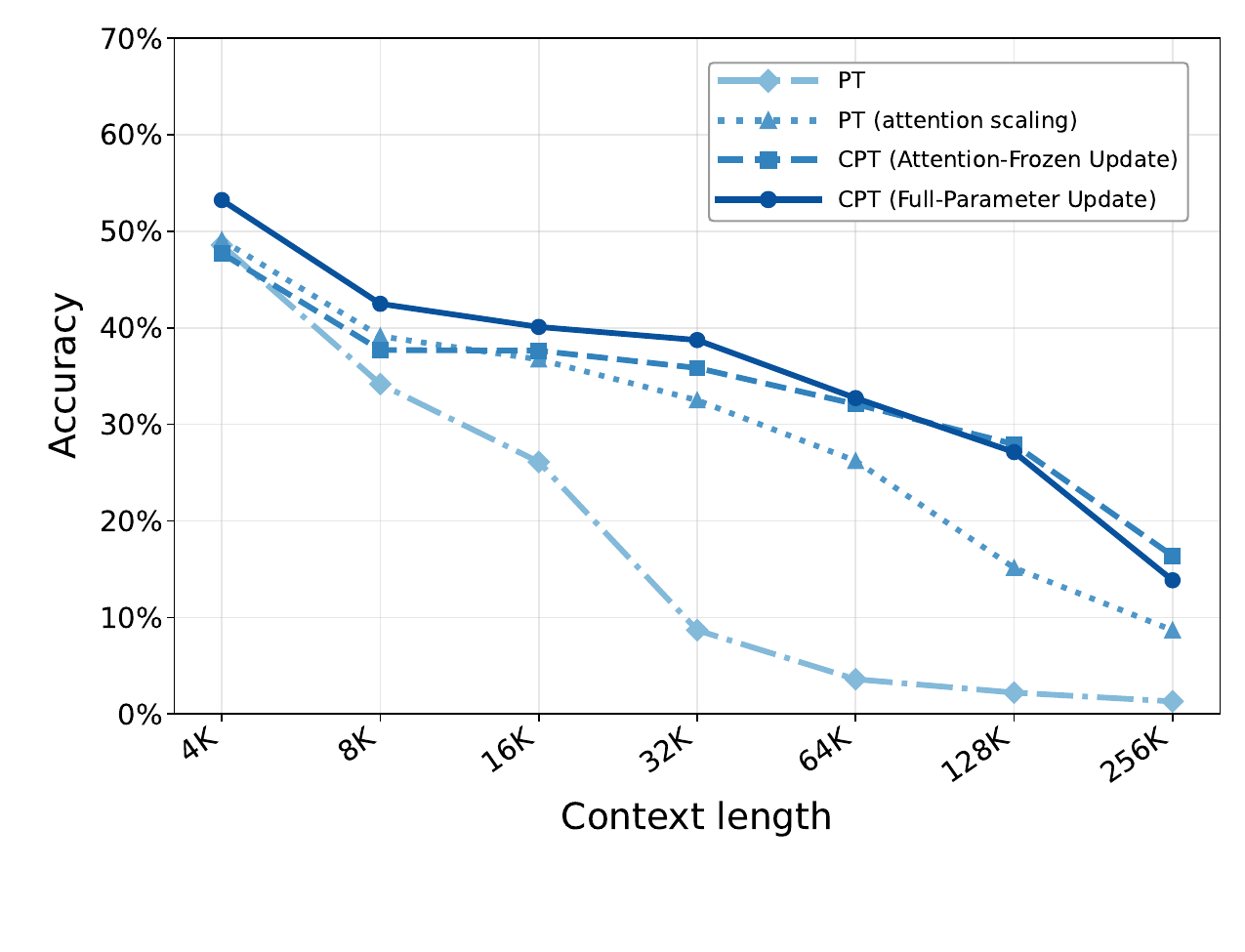}
        \vspace{-1.5em}
        \par\small (b) \texttt{0.6B-d256}
    \end{minipage}
    \begin{minipage}[t]{0.32\linewidth}
        \centering
        \includegraphics[width=\linewidth]{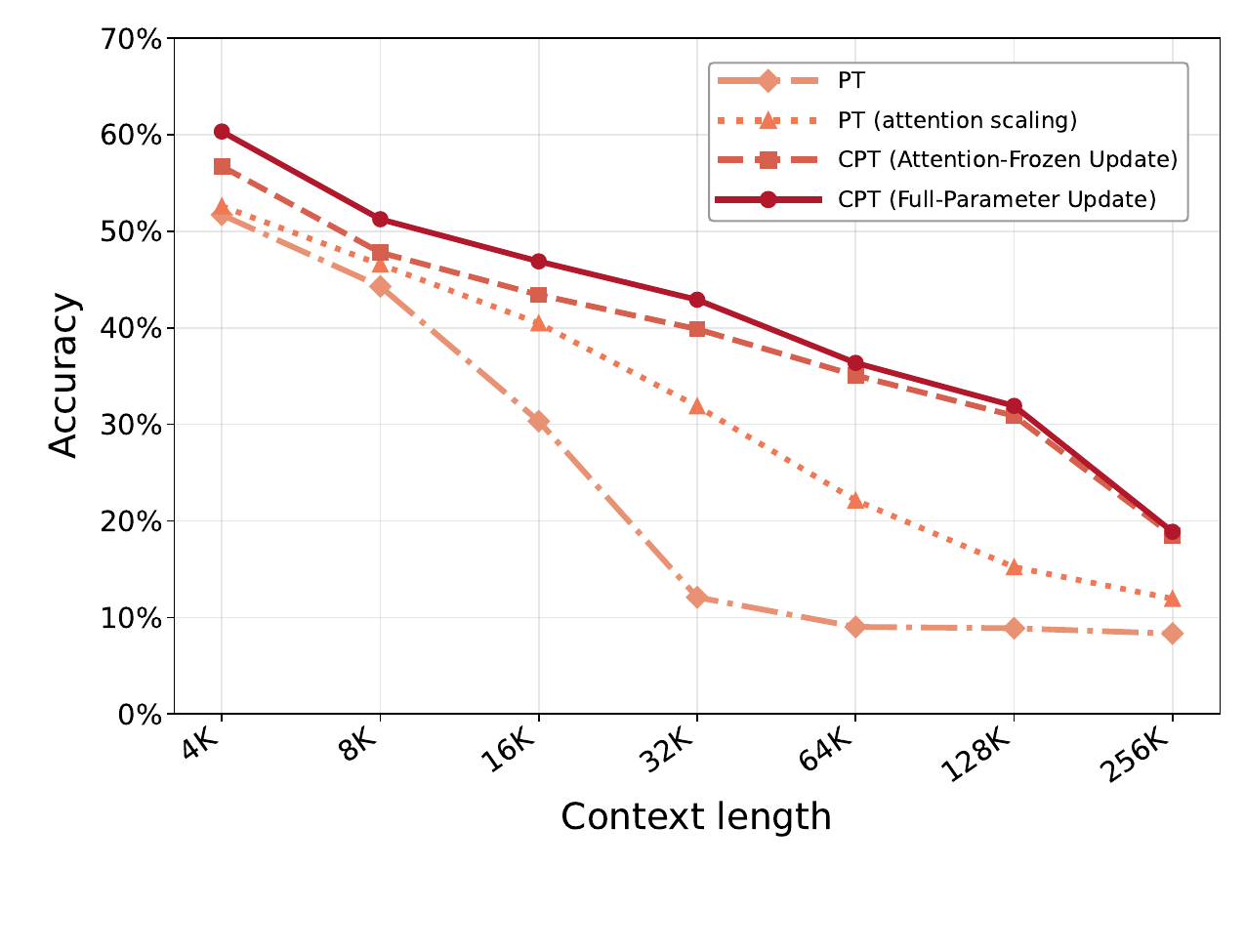}
        \vspace{-1.5em}
        \par\small (c) \texttt{0.6B-d512}
    \end{minipage}
    \caption{RULER-100 scores by context length for unmodified PT,
    scaled PT, Full CPT, and Frozen CPT. At each context length,
    scores are averaged over 13 tasks within each run and then
    across three runs. Scaled PT uses the inference-time procedure
    in Appendix~\ref{app:pt_temperature_scaling}.}
    \label{fig:attention_freeze_ruler_by_length}
\end{figure}

We then compare RULER-100 performance across the four conditions.
Figure~\ref{fig:pt_cpt_ruler_overall} reports mean overall scores, obtained by averaging equally over 13 tasks and seven context lengths from 4K to 256K within each run and then across three runs.
Figure~\ref{fig:attention_freeze_ruler_by_length} provides the corresponding length-wise results, averaging over the 13 tasks within each run and then across three runs.

Relative to unmodified PT, both CPT conditions substantially improve mean overall RULER-100 scores for all three head allocations, with particularly clear gains at longer context lengths.
Full CPT achieves higher mean overall scores than Frozen CPT for all three configurations, but the additional benefit is modest and similar across head dimensions: 2.19, 1.85, and 2.32 points for \texttt{d128}, \texttt{d256}, and \texttt{d512}, respectively.
Thus, the higher mean retrieval scores under Full CPT are not accompanied by lower mean final training loss.

In contrast, the mean improvement of Frozen CPT over scaled PT depends strongly on head dimension: $-0.01$, $+3.97$, and $+7.35$ points for \texttt{d128}, \texttt{d256}, and \texttt{d512}, respectively.
Frozen CPT therefore nearly matches scaled PT in mean overall performance for \texttt{d128}, while achieving progressively larger gains for the wider-head models.
The head-dimension dependence of retrieval gains thus persists even without updates to the projection weights in attention modules.

\begin{figure}[p]
    \centering
    \includegraphics[width=0.9\linewidth]{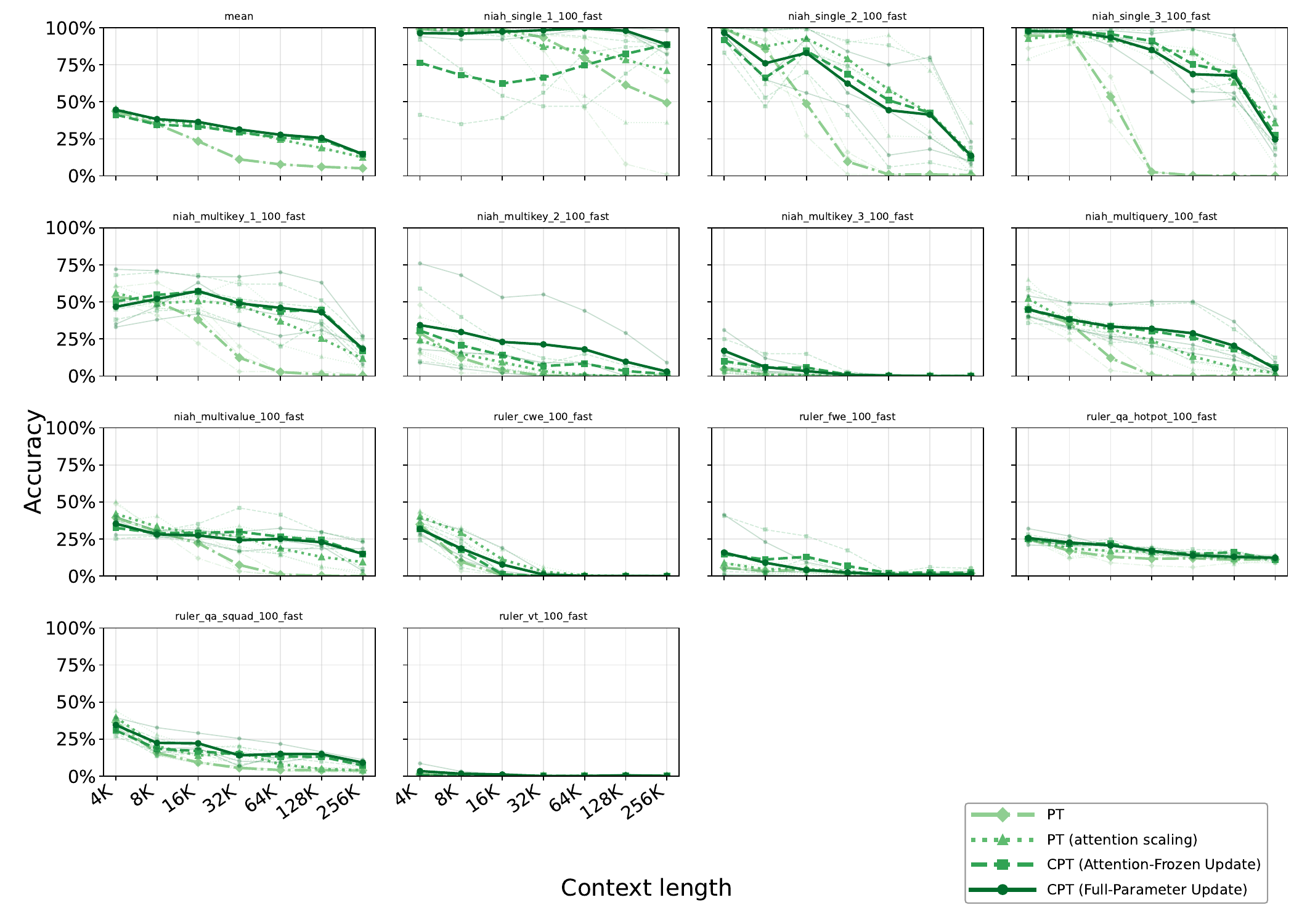}
    \caption{Task-level RULER-100 scores for unmodified PT, scaled PT, Full CPT, and Frozen CPT of \texttt{0.6B-d128}.
    The first panel shows the 13-task mean; the remaining panels show individual tasks.
    Thick lines indicate means across three runs, and faint lines show individual runs.
    The horizontal axis shows context length, and the vertical axis shows task scores as percentages.
    Scaled PT is labeled ``PT (attention scaling)'' and follows Equation~\ref{eq:pt_attention_scaling}.}
    \label{fig:freeze_tasks_d128}
\end{figure}

\begin{figure}[p]
    \centering
    \includegraphics[width=0.9\linewidth]{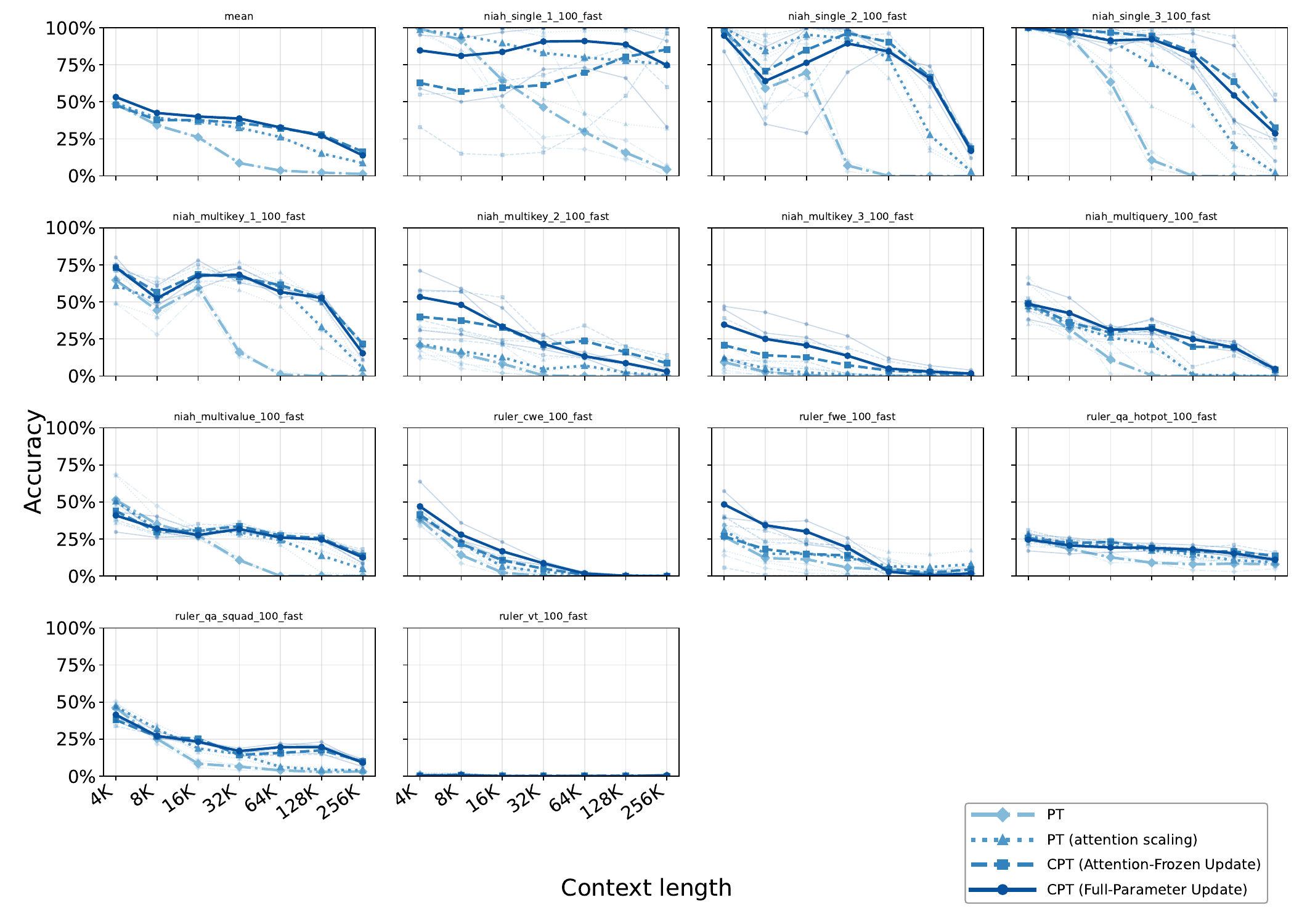}
    \caption{Task-level RULER-100 scores for unmodified PT, scaled PT, Full CPT, and Frozen CPT of \texttt{0.6B-d256}.
    The first panel shows the 13-task mean; the remaining panels show individual tasks.
    Thick lines indicate means across three runs, and faint lines show individual runs.
    Axes and scaling follow Figure~\ref{fig:freeze_tasks_d128}.}
    \label{fig:freeze_tasks_d256}
\end{figure}

\begin{figure}[p]
    \centering
    \includegraphics[width=0.9\linewidth]{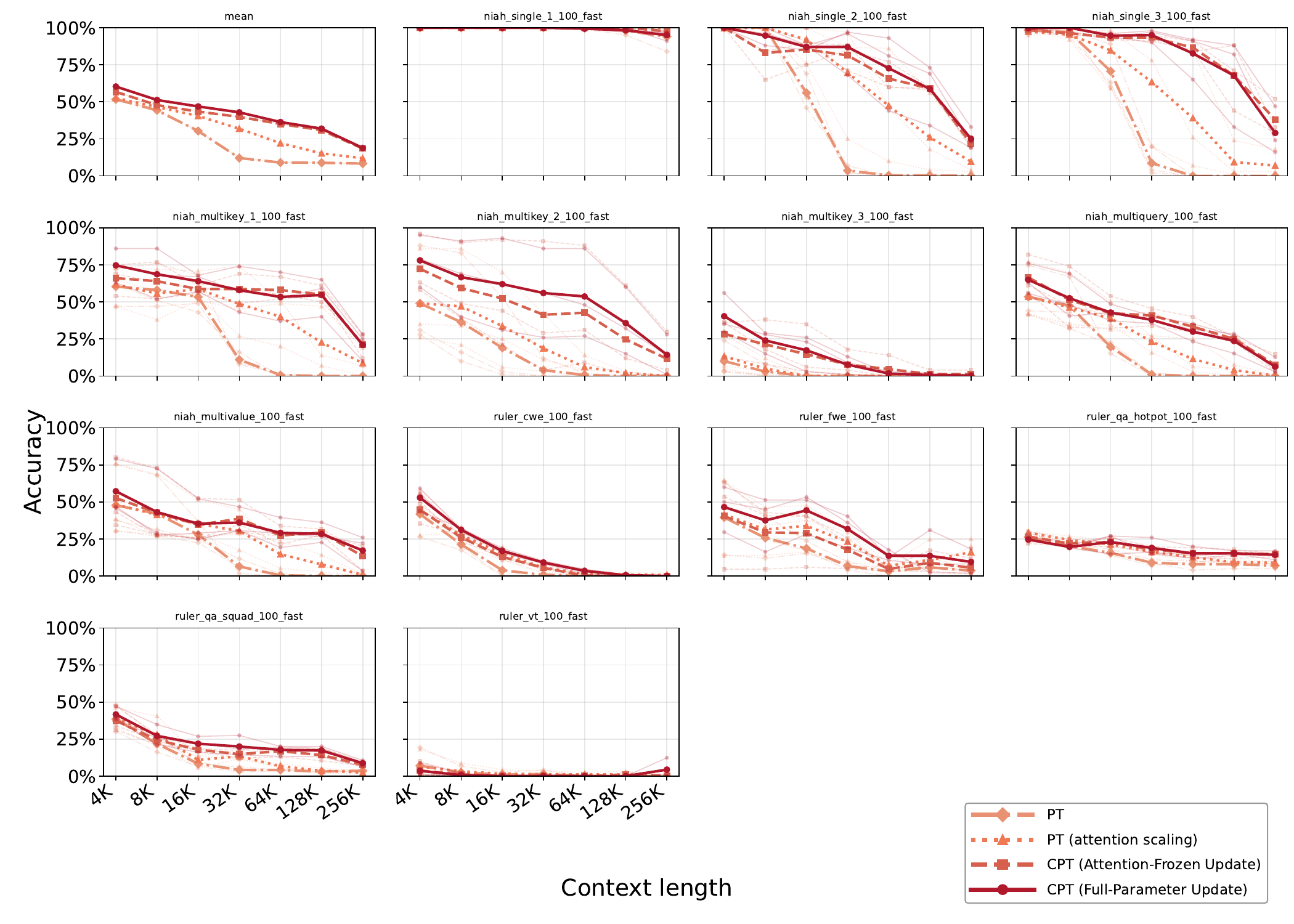}
    \caption{Task-level RULER-100 scores for unmodified PT, scaled PT, Full CPT, and Frozen CPT of \texttt{0.6B-d512}.
    The first panel shows the 13-task mean; the remaining panels show individual tasks.
    Thick lines indicate means across three runs, and faint lines show individual runs.
    Axes and scaling follow Figure~\ref{fig:freeze_tasks_d128}.
    Full achieves a higher overall mean score than Frozen, with gains and losses depending on the task and context length.}
    \label{fig:freeze_tasks_d512}
\end{figure}

Table~\ref{tab:ruler_seed_condition_scores} reports overall RULER-100 scores for individual training runs.
Figures~\ref{fig:freeze_tasks_d128}, \ref{fig:freeze_tasks_d256}, and~\ref{fig:freeze_tasks_d512} provide task- and length-level comparisons for the three head allocations.
These detailed results show that the relative performance of Full CPT and Frozen CPT can vary across runs, tasks, and context lengths, despite the higher mean overall scores under Full CPT.

\begin{table}[p]
    \centering
    \caption{Training NTP loss at the final logged interval of Full and Frozen CPT.
    Entries are means $\pm$ sample standard deviations across three runs.
    $\Delta_{\mathrm{loss}}$ is the Full-minus-Frozen loss difference, computed within each paired run before aggregation; positive values indicate lower loss under Frozen CPT.
    Differences are computed before rounding.}
    \label{tab:attention_freeze_ntp_loss}
    \begin{tabular}{c|ccc}
        \toprule
        Model & Full CPT & Frozen CPT & $\Delta_{\mathrm{loss}}$ \\
        \midrule
        \texttt{0.6B-d128} & $1.911\pm0.009$ & $1.884\pm0.019$ & $0.027\pm0.028$ \\
        \texttt{0.6B-d256} & $1.888\pm0.042$ & $1.878\pm0.020$ & $0.011\pm0.062$ \\
        \texttt{0.6B-d512} & $1.912\pm0.011$ & $1.881\pm0.018$ & $0.031\pm0.028$ \\
        \bottomrule
    \end{tabular}
\end{table}

\begin{table}[p]
    \centering
    \small
    \caption{Overall RULER-100 scores for individual 0.6B training runs, averaged equally over 13 tasks and seven context lengths from 4K to 256K. PT denotes evaluation without additional attention scaling; PT (attention scaling) denotes scaled PT. Full CPT and Frozen CPT follow the parameter-update policies in Table 5. Only Full CPT results are reported for the asymmetric configurations \(\texttt{k512-v128}\) and \(\texttt{k128-v512}\); dashes indicate unreported conditions.}
    \label{tab:ruler_seed_condition_scores}
    \begin{tabular}{llrrrr}
        \toprule
        Configuration & Run & PT & \shortstack{PT \\ (attention scaling)} & \shortstack{Frozen \\ CPT} & \shortstack{Full \\ CPT} \\ \midrule
        d128 & 1 & 18.57 & 28.36 & 23.23 & 25.50 \\
        d128 & 2 & 19.54 & 31.04 & 38.13 & 40.07 \\
        d128 & 3 & 18.63 & 27.81 & 25.83 & 28.19 \\
        \midrule
        d256 & 1 & 19.49 & 30.53 & 37.24 & 37.98 \\
        d256 & 2 & 17.57 & 29.21 & 33.81 & 37.05 \\
        d256 & 3 & 16.35 & 29.17 & 29.77 & 31.34 \\
        \midrule
        d512 & 1 & 21.52 & 24.41 & 35.43 & 40.77 \\
        d512 & 2 & 26.61 & 36.15 & 45.58 & 47.28 \\
        d512 & 3 & 22.42 & 34.06 & 35.65 & 35.58 \\
        \midrule
        k512-v128 & 1 & -- & -- & -- & 33.97 \\
        k512-v128 & 2 & -- & -- & -- & 42.19 \\
        k512-v128 & 3 & -- & -- & -- & 37.17 \\
        \midrule
        k128-v512 & 1 & -- & -- & -- & 29.29 \\
        k128-v512 & 2 & -- & -- & -- & 34.24 \\
        k128-v512 & 3 & -- & -- & -- & 30.32 \\
        \bottomrule
    \end{tabular}
\end{table}

\section{Supplementary results of retrieval-head analysis}
\label{app:retrieval_head_freeze}

This section provides the run-level breakdown of the retrieval-head counts and shared membership summarized in Table~\ref{tab:retrieval_head_pt_cpt_overlap} and extends the comparison to Frozen CPT.
Table~\ref{tab:retrieval_head_freeze_seeds} reports results for three training runs of each symmetric 0.6B configuration, comparing scaled PT with the corresponding Full CPT and Frozen CPT checkpoints.
We use the same detection protocol and retrieval-score threshold across all conditions.

For \texttt{d128}, the mean retrieval-head count across three runs increases from 4.00 under scaled PT to 7.00 under both Full CPT and Frozen CPT.
Both CPT conditions increase the detected count in all three runs, although the magnitude of the increase varies across runs.
Thus, the increase in detected retrieval-head count for \texttt{d128} is also observed when the projection weights in attention modules remain fixed during CPT.

Changes relative to scaled PT are smaller for the wider-head models.
For \texttt{d256}, the mean count is 8.33 under scaled PT, 8.33 under Full CPT, and 8.00 under Frozen CPT.
For \texttt{d512}, the corresponding mean counts are 5.00, 6.33, and 5.67.
These smaller changes are consistent with much of the wider-head models' detected retrieval-head allocation already being visible before CPT under attention scaling.

Full CPT and Frozen CPT yield similar retrieval-head counts across all three configurations.
Moreover, most heads detected under Full CPT are also detected under Frozen CPT in each run.
These results indicate that most of these heads can meet the retrieval criterion after CPT without updates to the projection weights in attention modules.

\begin{table}[t]
    \centering
    \small
    \setlength{\tabcolsep}{4pt}
    \caption{Retrieval-head counts and shared membership for each
    training run. PT denotes scaled PT; Full and Frozen denote the
    corresponding 8B-token CPT checkpoints without additional scaling.
    Shared counts are intersections of the detected $(\mathrm{layer},
    \mathrm{query\ head})$ sets. $\Delta N$ is the Full-CPT count minus
    the scaled-PT count; it measures the net count change, not the number
    of heads newly detected under Full CPT.}
    \label{tab:retrieval_head_freeze_seeds}
    \begin{tabular}{llrrrrrr}
        \toprule
        Model & Run & \shortstack{Scaled\\PT}
        & \shortstack{Full\\CPT} & \shortstack{Frozen\\CPT}
        & \shortstack{Shared\\PT/Full}
        & $\Delta N$ & \shortstack{Shared\\Full/Frozen} \\
        \midrule
        \texttt{0.6B-d128} & 1 & 3 & 9 & 8 & 3 & +6 & 8 \\
         & 2 & 5 & 6 & 7 & 5 & +1 & 6 \\
         & 3 & 4 & 6 & 6 & 4 & +2 & 6 \\
        \midrule
        \texttt{0.6B-d256} & 1 & 7 & 7 & 8 & 7 & 0 & 7 \\
         & 2 & 9 & 9 & 9 & 9 & 0 & 9 \\
         & 3 & 9 & 9 & 7 & 8 & 0 & 7 \\
        \midrule
        \texttt{0.6B-d512} & 1 & 6 & 9 & 8 & 6 & +3 & 8 \\
         & 2 & 3 & 4 & 3 & 3 & +1 & 3 \\
         & 3 & 6 & 6 & 6 & 6 & 0 & 6 \\
        \bottomrule
    \end{tabular}
\end{table}

The scaled-PT results show that much of the retrieval heads detected in the models with wider heads are already visible before CPT under attention scaling.
Most heads detected under Full CPT are also detected under Frozen CPT, indicating that updates on projection weights in attention modules during CPT are not necessary for most of these heads to meet the retrieval criterion after training.

\section{Experimental details for attention-map and representation analyses}
\label{app:attention_representation_details}

This appendix describes the experimental setup and measurement procedures for the analyses in Section~\ref{sec:token_rank_reorganization}.
We first detail the attention entropy and capped top-10k rank-correlation diagnostics.
We then describe the residual-stream effective-rank measurements at a context length of 32K.

\subsection{Attention entropy and capped rank correlation}
\label{app:attention_map_setup}

This section describes the experimental setup and metrics for the attention-map diagnostics in Figure~\ref{fig:attention_map}.
We analyze all six full-attention (FA) layers of \texttt{0.6B-d128}, \texttt{0.6B-d256}, and \texttt{0.6B-d512}, using three training runs per configuration.

We use six needle-retrieval tasks from RULER---three single-needle variants and three multi-key variants---whose reference answers have a uniquely specified token order.
At each context length of 32K and 64K, we use 300 shared examples per length.
We teacher-force the reference answers and analyze attention maps from queries predicting the first three reference-answer tokens.
These experiments focus on answer initiation, when retrieval is expected to be more demanding because little reference-answer content is available to guide the continuation.

For each combination of an example, an FA layer, a query head, and an answer query, let $p_j$ denote the attention probability at causally visible token position $j$.
We compute attention entropy as
\begin{equation}
    H = -\sum_{j\ \mathrm{visible}} p_j \log p_j,
\end{equation}
using natural logarithms.
Lower entropy indicates more concentrated attention, but does not establish whether the attended positions are relevant.
Comparing the entropy between PT and each CPT condition therefore measures changes in attention probability concentration during CPT.

To measure changes in token prioritization, we compute capped top-10k rank correlation between each CPT checkpoint and the corresponding PT checkpoint.
The comparison uses the same example, layer, query-head index, and answer-query position.
Let $r_j^M$ denote the descending, one-based rank of the attention probability at position $j$ in checkpoint $M$, using average ranks for ties.
For corresponding checkpoints $A$ and $B$, we define
\begin{align}
    \widetilde{r}_j^M(K) =
    \begin{cases}
        r_j^M, & r_j^M \le K, \\
        K+1,   & r_j^M > K,
    \end{cases}
    \qquad
    \rho_K(A,B) =
    \operatorname{corr}_{\mathrm{Pearson}}
    \left(
        \widetilde{\mathbf{r}}^A(K),
        \widetilde{\mathbf{r}}^B(K)
    \right)
\end{align}
with $K=10{,}000$ and rank vectors aligned by token position.
Capping removes rank distinctions within the low-attention tail while retaining those positions in the correlation.
Lower correlation indicates greater changes in which tokens receive high attention and how they are ordered, rather than changes in probability magnitudes.

\begin{figure}[t]
    \centering
    \includegraphics[width=\linewidth]{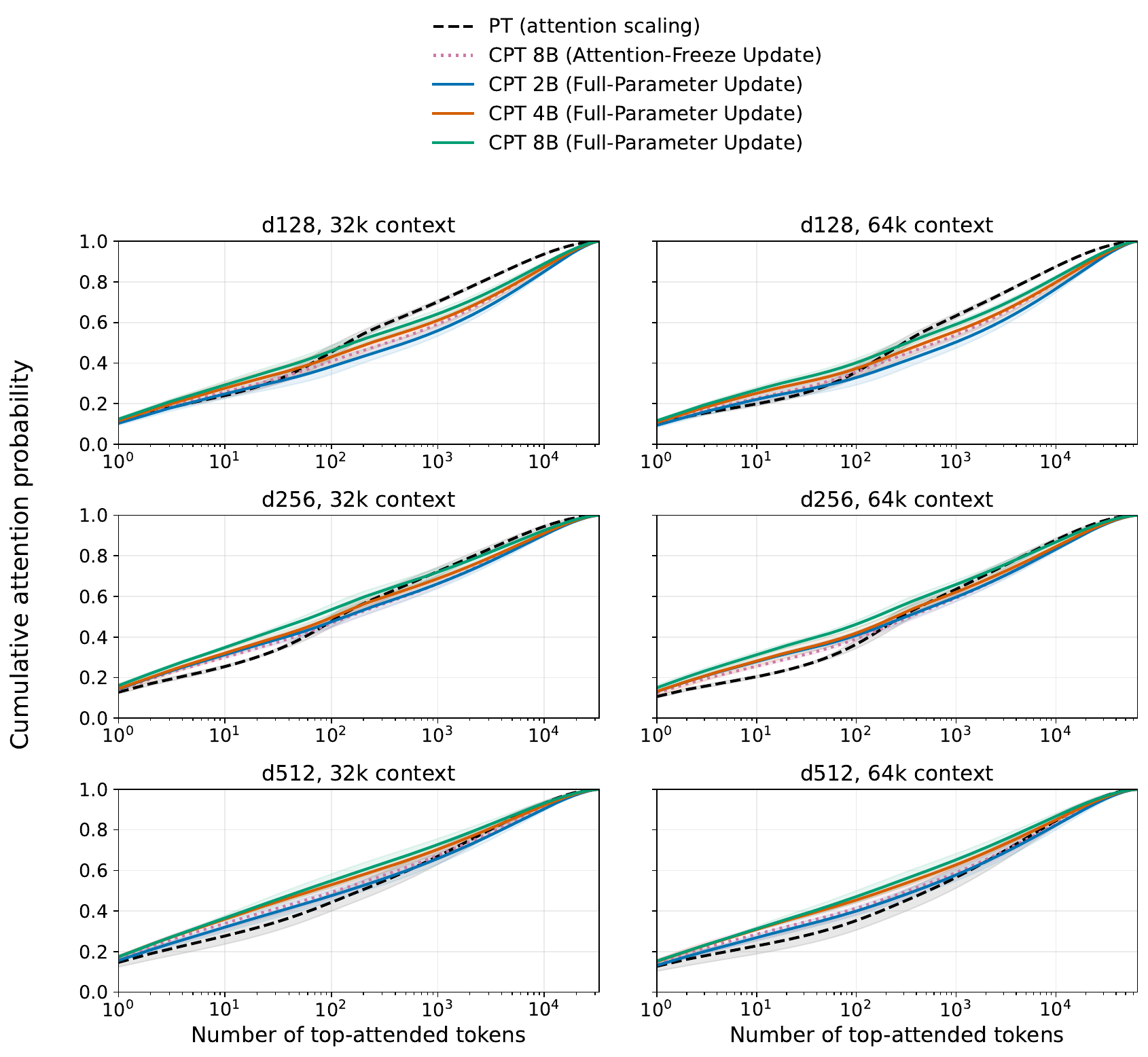}
    \caption{Cumulative attention probability over positions ranked by decreasing attention probability. Rows show \texttt{0.6B-d128}, \texttt{0.6B-d256}, and \texttt{0.6B-d512}; columns show 32K and 64K context lengths. Each panel compares scaled PT, Full CPT at 2B, 4B, and 8B tokens, and Frozen CPT at 8B tokens. The horizontal axis gives the number of ranked positions on a logarithmic scale. }
    \label{fig:attention_probability_concentration}
\end{figure}

To guide the choice of the rank cutoff, we examine the cumulative attention probability curves in Figure~\ref{fig:attention_probability_concentration}.
The figure compares scaled PT, Full CPT at 2B, 4B, and 8B tokens, and Frozen CPT at 8B tokens for the three head allocations at context lengths of 32K and 64K.
For each attention distribution, we sort the probabilities in descending order, $p_{(1)} \ge p_{(2)} \ge \cdots$, and define
\begin{equation}
    C(n) = \sum_{j=1}^{\min(n,N)} p_{(j)}
\end{equation}
where $N$ is the number of causally visible positions.
Thus, $C(n)$ measures the probability mass assigned to the $n$ most-attended token positions.
Sorting and accumulation are performed separately for each attention distribution before aggregation.

As shown in Figure~\ref{fig:attention_probability_concentration}, the top 10k positions account for most attention probability mass on average.
Guided by these curves, we choose $K=10{,}000$ as a heuristic cutoff, retaining rank distinctions among higher-attention positions while suppressing those in the low-attention tail.

For Figure~\ref{fig:attention_map}, both metrics are computed before aggregation.
Within each training run and context length, we take an average over examples, FA layers, query heads, and the three selected answer queries.
We then report the mean and sample standard deviation across three runs.

\subsection{Residual-stream effective rank}
\label{app:effective_rank_setup}

This section describes the effective-rank measurements in Figure~\ref{fig:residual_effective_rank}.
We analyze the full-attention (FA) blocks of \texttt{0.6B-d128}, \texttt{0.6B-d256}, and \texttt{0.6B-d512} under scaled PT, Frozen CPT, and Full CPT, using the same evaluation examples of 32K context length as in Appendix~\ref{app:attention_map_setup}.
Effective rank summarizes the concentration of the singular-value spectrum, with higher values indicating a more evenly distributed spectrum.

We measure the residual stream at three sites within each FA block: immediately before the attention block (Pre-FA), after adding the attention output to the residual stream (Post-FA), and after adding the output of the subsequent MLP block (Post-MLP).
For each example and measurement site, we stack the prompt-token representations as rows of a matrix and perform singular value decomposition (SVD).
We then average the ordered singular-value spectra across examples and three training runs, separately for each configuration, checkpoint condition, FA block, and measurement site.

Let $\bar{\sigma}_i$ denote the $i$-th entry of the mean spectrum.
We compute effective rank as
\begin{equation}
    \mathrm{ER} = \exp\left(-\sum_i p_i \log p_i\right),
    \qquad
    p_i = \frac{\bar{\sigma}_i}{\sum_j \bar{\sigma}_j}.
\end{equation}
Each reported value is therefore the effective rank of the mean singular-value spectrum, rather than the mean of effective ranks computed for individual examples or runs.

\end{document}